\documentclass[11pt]{article}

\usepackage{times}
\usepackage{parskip}
\usepackage{microtype}
\usepackage{graphicx}
\usepackage{subfigure}
\usepackage{tabularx,booktabs} 

\usepackage{algorithm,algorithmic}

\usepackage{epsfig,amsmath,amssymb,amsfonts,amstext,amsthm,mathrsfs}
\usepackage{latexsym,graphics,epsf,epsfig,psfrag}
\usepackage{dsfont,color,epstopdf,fixmath}
\usepackage{enumitem}

\usepackage[utf8]{inputenc} 
\usepackage[T1]{fontenc}    

\usepackage{booktabs}       

\usepackage{hhline}

\usepackage{threeparttable}
\usepackage{booktabs, caption, makecell}

\usepackage{amsfonts}       
\usepackage{nicefrac}       
\usepackage{microtype}      
\usepackage{subfigure}
\usepackage{mathtools}
\usepackage{amssymb}
\usepackage{amsthm}

\usepackage{natbib,har2nat}
\usepackage[colorlinks=true,citecolor=blue]{hyperref}

\usepackage{hyperref}
\usepackage{cleveref}

\usepackage{multirow}

\usepackage[table]{xcolor}

\newcommand{\mP}{\mathbb P}
\newcommand{\msP}{\mathsf{P}}
\newcommand{\msbP}{\breve{\mathsf{P}}}
\newcommand{\bs}{\breve{s}}
\newcommand{\ba}{\breve{a}}
\newcommand{\mE}{\mathbb E}
\newcommand{\mcw}{\mathcal W}
\newcommand{\mcs}{\mathcal S}
\newcommand{\mca}{\mathcal A}
\newcommand{\mf}{\mathcal F}

\newcommand{\mcpi}{{\rm \Pi}}
\newcommand{\ltwo}[1]{\left\|#1\right\|_2}
\newcommand{\lone}[1]{\left|#1\right|}

\newcommand{\lF}[1]{\left\|#1\right\|_F}
\newcommand{\lTV}[1]{\left\|#1\right\|_{TV}}
\newcommand{\linf}[1]{\left\|#1\right\|_\infty}

\newcommand{\mR}{\mathbb{R}}

\newtheorem{theorem}{Theorem}

\newtheorem{lemma}{Lemma}

\newtheorem{assumption}{Assumption}

\allowdisplaybreaks[4]

\begin{document}

\begin{center}
	\baselineskip 1.3ex {\Large \bf Non-asymptotic Convergence Analysis of Two Time-scale \\ \vspace{2mm} (Natural) Actor-Critic Algorithms
		\footnote{The results of this paper were initially submitted for publication in February 2020.}
		\footnote{The work was supported partially by the U.S. National Science Foundation under the grants CCF-1801855, CCF-1761506 and CCF-1900145.}
	}
	\\
	\vspace{1cm} 
	
	{\large Tengyu Xu, Zhe Wang, Yingbin Liang}\\
	\vspace{4mm}
	\begin{small}
		Department of Electrical and Computer Engineering \\
		The Ohio State University \\
		Columbus, OH 43220 USA \\ \vspace{1mm}
		Email: \{xu.3260,wang.10982,liang.889\}@osu.edu
	\end{small}
\end{center}

\vspace{0.15in}

\begin{abstract}
As an important type of reinforcement learning algorithms, actor-critic (AC) and natural actor-critic (NAC) algorithms are often executed in two ways for finding optimal policies. In the first nested-loop design, actor's one update of policy is followed by an entire loop of critic's updates of the value function, and the finite-sample analysis of such AC and NAC algorithms have been recently well established. The second two time-scale design, in which actor and critic update simultaneously but with different learning rates, has much fewer tuning parameters than the nested-loop design and is hence substantially easier to implement. Although two time-scale AC and NAC have been shown to converge in the literature, the finite-sample convergence rate has not been established. In this paper, we provide the first such non-asymptotic convergence rate for two time-scale AC and NAC under Markovian sampling and with actor having general policy class approximation. We show that two time-scale AC requires the overall sample complexity at the order of $\mathcal{O}(\epsilon^{-2.5}\log^3(\epsilon^{-1}))$ to attain an $\epsilon$-accurate stationary point, and two time-scale NAC requires the overall sample complexity at the order of $\mathcal{O}(\epsilon^{-4}\log^2(\epsilon^{-1}))$ to attain an $\epsilon$-accurate global optimal point. We develop novel techniques for bounding the bias error of the actor due to dynamically changing Markovian sampling and for analyzing the convergence rate of the linear critic with dynamically changing base functions and transition kernel.\end{abstract}

\section{Introduction}

Policy gradient (PG) \cite{sutton2000policy,williams1992simple} is one of the most popular algorithms used in reinforcement learning (RL) \cite{sutton2018reinforcement} for searching a policy that maximizes the expected total reward over a period of time. The idea is to parameterize the policy and then apply the gradient-based method to iteratively update the parameter in order to obtain a desirable solution. The performance of PG algorithms highly depends on how we estimate the value function in the policy gradient based on collected samples in practice. Since PG directly utilizes Monte Carlo rollout to estimate the value function, it typically have large variance and are not stable in general. 
Actor-critic (AC) algorithms were proposed in \cite{konda1999actor,konda2000actor}, in which the estimation of the value function is improved by separately running critic's update in an alternating manner jointly with actor's update of the policy, and hence the stability and the overall performance are substantially improved. The natural actor-critic (NAC) algorithm was further proposed in \cite{bhatnagar2009natural} using the natural policy gradient (NPG) \cite{kakade2002natural,amari1998natural} so that the policy update is invariant to the parameterization of the policy.


AC algorithms are typically implemented in two ways: nested loop and two time-scale. First, in the {\em nested-loop} AC and NAC algorithms, actor's one update in the outer loop is followed by critic's numerous updates in the inner loop to obtain an accurate value function. 
The convergence rate (or sample complexity) of nested-loop AC and NAC has been extensively studied recently \cite{wang2019neural,yang2019provably,kumar2019sample,qiu2019finite,xu2020improving} (see \Cref{sec:relatedwork} for more details). Second, in {\em two time-scale} AC and NAC algorithms \cite{konda2000actor,bhatnagar2009natural}, actor and critic update {\em simultaneously} but with the stepsize diminishing at different rates. Typically, actor updates at a slower time-scale, and critic updates at a faster time-scale. The asymptotic convergence of two time-scale AC and NAC has been well established under both i.i.d.\ sampling \cite{bhatnagar2009natural} and Markovian sampling \cite{konda2002actor}. However, the finite-time analysis, i.e., the sample complexity, of two time-scale AC and NAC has not been characterized yet. 

{\em Thus, our goal here is to provide the first sample complexity (i.e., non-asymptotic convergence) analysis for two time-scale AC and NAC under the dynamic Markovian sampling.} Such a study is important, because two time-scale AC and NAC algorithms have much fewer tuning parameters than the nested-loop design (which needs to additionally tune the running length for each inner loop) and is hence substantially easier to implement.

\subsection{Our Contributions}

In this paper, we provide the first non-asymptotic convergence and sample complexity analysis for two time-scale AC and NAC under Markovian sampling and with actor having general policy class approximation, in which both actor and critic's iterations take diminishing stepsizes but at different rates. We show that two time-scale AC requires the overall sample complexity at the order of $\mathcal{O}(\epsilon^{-2.5}\log^3(\epsilon^{-1}))$ to attain an $\epsilon$-accurate stationary point, and two time-scale NAC requires the overall sample complexity at the order of $\mathcal{O}(\epsilon^{-4}\log^2(\epsilon^{-1}))$ to attain an $\epsilon$-accurate {\em globally} optimal point.



Two time-scale AC and NAC generally falls into the type of two time-scale {\em nonlinear} SA algorithms, due to the nonlinear parameterization of the policy. Thus, this paper develops the first finite-sample analysis for such two time-scale nonlinear SA, which is very different from the existing finite-sample analysis of two time-scale linear SA (see more discussions in \Cref{sec:relatedwork}).
More specifically, we have the following new technical developments. 
\begin{itemize}
\item[(a)] The iteration of critic corresponds to a linear SA with {\em dynamically changing} base functions and a transition kernel due to its simultaneous update with actor, which is significantly different from the typical policy evaluation algorithms (or more generally stochastic approximation (SA) algorithms) that are associated with fixed base functions and a transition kernel. Thus, we develop several new techniques to analyze how critic tracks the dynamically changing fixed point for bounding the bias error and a fixed-point drift error, which add new contributions to the literature of linear SA. 

\item[(b)] The iteration of actor corresponds to a nonlinear SA update due to the nonlinear parameterization of the policy, and hence requires to bound the bias error due to dynamically changing Markovian sampling for nonlinear SA, which has not been studied before. We develop new techniques to provide such a bias error bound and show that the bias error converges to zero under a diminishing stepsize, which is new in the literature.
\end{itemize}


\subsection{Related Work}\label{sec:relatedwork}

Due to the extensive studies on the general topic of policy gradient, we include here only theoretical studies of AC and NAC as well as the finite-sample analysis of two time-scale RL algorithms, which are highly relevant to our work. 

{\bf Two time-scale AC and NAC.} The first AC algorithm was proposed by \cite{konda2000actor} and was later extended to NAC in \cite{peters2008natural} using the natural policy gradient (NPG) \cite{kakade2002natural}. The asymptotic convergence of two time-scale (or multi-time-scale) AC and NAC algorithms under both i.i.d.\ sampling and Markovian sampling have been established in \cite{kakade2002natural,konda2002actor,bhatnagar2010actor,bhatnagar2009natural,bhatnagar2008incremental}, but the non-asymptotic convergence and sample complexity were not established for two time-scale AC and NAC, which is the focus of this paper.

{\bf Nested-loop AC and NAC.} The convergence rate (or sample complexity) of nested-loop AC and NAC has been studied recently. More specifically,
\cite{yang2019provably} studied the sample complexity of AC with linear function approximation in the LQR problem. \cite{wang2019neural} studied AC and NAC in the regularized MDP setting, in which both actor and critic utilize overparameterized neural networks as approximation functions. \cite{agarwal2019optimality} studied nested-loop natural policy gradient (NPG) for general policy class (which can be equivalently viewed as NAC, although is not explicitly formulated that way). \cite{kumar2019sample} studied AC with general policy class and linear function approximation for critic, but with the requirement that the true value function is in the linear function class of critic. \cite{qiu2019finite} studied a similar problem as in \cite{kumar2019sample} with weaker assumptions. More recently, \cite{xu2020improving} studied AC and NAC under Markovian sampling and with actor having general policy class, and showed that the mini-batch sampling improves the sample complexity of previous studies orderwisely.



\textbf{Policy gradient.} PG and NPG algorithms \cite{williams1992simple,baxter2001infinite,sutton2000policy,kakade2002natural} have been extensively studied in the past for various scenarios. More specifically, \cite{fazel2018global,malik2018derivative,tu2018gap} established the global convergence of PG/NPG in LQR problem, and \cite{bhandari2019global} studied the the global property of landscape in tabular cases. Furthermore, \cite{shen2019hessian,papini2018stochastic,papini2017adaptive,xu2019improved,xu2019sample} studied variance reduced PG with general nonconcave/nonconvex function approximation for finite-horizon scenarios, and showed that variance reduction can effectively reduce the sample complexity both theoretically and experimentally. \cite{karimi2019non,zhang2019global,xiong2020amsgradRL} studied the convergence of PG for the infinite-horizon scenario and under Markovian sampling. Moreover, \cite{shani2019adaptive,liu2019neural} studied TRPO/PPO for the tabular case and with the neural network function approximation, respectively. This paper focuses on a different variant of PG, i.e., AC and NAC algorithms, in which the estimation of the value function by critic is separate from the PG update by actor to reduce the variance. The analysis of these algorithms thus involves very different techniques.


\textbf{Two time-scale SA.} The finite-sample analysis of critic in two time-scale AC and NAC in this paper is related to but different from the existing studies in two time-scale SA, which we briefly summarize as follows. The asymptotic convergence of two time-scale linear SA with martingale noise has been established in \cite{borkar2009stochastic}, and the non-asymptotic analysis has been provided in \cite{dalal2018,dalal2019tale}. Under Markovian setting, the asymptotic convergence of two time-scale linear SA has been studied in \cite{karmakar2017two,tadic2004almost,yaji2016stochastic}, and the non-asymptotic analysis of two time-scale linear SA was established recently in \cite{xu2019two,kaledin2020finite} under diminishing stepsize and in \cite{gupta2019finite} under constant stepsize.

For two time-scale nonlinear SA, most of the convergence results are developed under global (local) asymptotic stability assumptions or local linearization assumptions. The asymptotic convergence of two time-scale nonlinear SA with martingale noise has been established in \cite{borkar1997stochastic,borkar2009stochastic,tadic2004almost}, and the non-asymptotic convergence of two time-scale nonlinear SA with martingale noise has been studied in \cite{borkar2018concentration,mokkadem2006convergence}. Under the Markovian setting, the asymptotic convergence of two time-scale nonlinear SA was studied in \cite{karmakar2016dynamics,yaji2016stochastic,karmakar2017two}. In this paper, AC and NAC can be modeled as a two time-scale nonlinear SA, in which the fast time-scale iteration corresponds to a linear SA with dynamically changing base functions and transition kernel, and the slow time-scale iteration corresponds to a general nonlinear SA. Without the stability and linearization assumptions, the asymptotic convergence of this special two time-scale nonlinear SA with Markovian noise was studied in \cite{konda2002actor}, but the non-asymptotic convergence rate has not been studied before, which is the focus of this paper.

\section{Preliminaries}
In this section, we introduce the AC and NAC algorithms under the general framework of Markov decision process (MDP) and discuss the technical assumptions in our analysis.

\subsection{Problem Formulation}\label{sec:prob}

We consider a dynamics system modeled by a Markov decision process (MDP). Here, at each time $t$, the state of the system is represented by $s_t$ that belongs to a state space $\mcs$, and an agent can take an action $a_t$ chosen from an action space $\mca$. Then the system transits into the next state $s_{t+1}\in\mcs$ with the probability governed by a transition kernel $\mathsf{P}(s_{t+1} |s_t,a_t)$, and receives a reward $r(s_t,a_t,s_{t+1})$. The agent's strategy of taking actions is captured by a policy $\pi$, which corresponds to a conditional probability distribution $\pi(\cdot|s)$, indicating the probability that the agent takes an action $a\in\mca$ given the present state $s$. 

Given an initial state $s_0$, the performance of a policy $\pi$ is measured by the state value function defined as
 $V_\pi(s)=\mE[\sum_{t=0}^{\infty}\gamma^t r(s_t,a_t, s_{t+1})|s_0=s,\pi],$
which is the accumulated reward over the entire time horizon, and where $\gamma\in(0,1)$ denotes the discount factor, and $a_t\sim\pi(\cdot|s_t)$ for all $t\geq 0$. If given the initial state $s_0=s$, an action $a_0=a$ is taken under the policy $\pi$, we further define the state-action value function as $Q_\pi(s,a)=\mE[\sum_{t=0}^{\infty}\gamma^t r(s_t,a_t, s_{t+1})|s_0=s,a_0=a,\pi]$. 



In this paper, we study the problem of finding an optimal policy $\pi^*$ that maximizes the expected total reward function given by 
\begin{align}\label{eq:prob}
\max_{\pi}J(\pi):=(1-\gamma)\mE\left[\sum_{t=0}^{\infty}\gamma^t r(s_t,a_t,s_{t+1})\right]=\mE_\zeta[V_\pi(s)],
\end{align}
where $\zeta$ denotes the distribution of the initial state $s_0\in\mcs$.

\subsection{Two Time-scale AC and NAC Algorithms}

In order to solve \cref{eq:prob}, we first parameterize the policy $\pi$ by $w\in \mcw \subset \mR^d$, which in general corresponds to a {\em nonlinear} function class. In this way, the problem in \cref{eq:prob} can be efficiently solved by searching over the parameter space $\mcw$. We further parameterize the value function for a given policy $\pi$ via the advantage function $A_\pi(s,a):=Q_\pi(s,a)-V_\pi(s)$ by a linear function class with base function $\phi(s,a)$, i.e., $A_{\theta}(s,a)=\phi(s,a)^\top \theta$. Such linear function parameterization does not lose the generality/optimality for finding the optimal policy as long as the compatibility condition is satisfied \cite{sutton2000policy,konda1999actor,konda2000actor}. The algorithms we consider below guarantee this by allowing the feature vector function $\phi(s,a)$ to vary in each actor's iteration.

We next describe the two time-scale AC and NAC algorithms for solving \cref{eq:prob}, which takes the form $\max_{w\in \mcw} J(\pi_w):=J(w)$ due to the parameterization of the policy. Both algorithms have actor and critic simultaneously update their corresponding variables with different stepsizes, i.e., actor updates at a slow time scale to optimize the policy $\pi_w$, and critic updates at a fast time scale to estimate the advantage function $A_\theta(s,a)$. 

For the two time-scale AC algorithm (see Algorithm \ref{algorithm_ttsac}), at step $t$, actor updates the parameter $w_t$ of policy $\pi_{w_t}$ via the first-order stochastic policy gradient step as
\begin{flalign*}
w_{t+1}=w_t+\alpha_t\widetilde{\nabla}J(w_t,\theta_t).
\end{flalign*}
where $\alpha_t>0$ is the stepsize, and $\widetilde{\nabla}J(w_t,\theta_t) = A_{\theta_t}(s_t,a_t)\phi_{w_t}(s_t,a_t)$. Here, $\widetilde{\nabla}J(w_t,\theta_t)$ serves as a stochastic approximation of the true gradient $\nabla J(w) = \mE_{\nu_{\pi_{w}}}\big[ A_{\pi_{w}}(s,a)\phi_{w}(s,a) \big]$ at time $t$, where $\nu_{\pi_w}(s,a)=(1-\gamma)\sum_{t=0}^{\infty}\gamma^t \mP(s_t=s)\pi_w(s|a)$ is the state-action visitation measure.





Critic's update of the parameter $\theta$ is to find the solution of the following problem
\begin{flalign}
 \min_{\theta\in\mR^d} L_w(\theta) := \mE_{\nu_{\pi_{w}}}[A_{\pi_{w}}(s,a) - \phi(s,a)^\top \theta ]^2 +\frac{1}{2}\lambda \|\theta\|_2^2.\label{eq: solu1}
\end{flalign}
where the regularization term $\lambda \|\theta\|_2^2$ is added here to guarantee the objective function to be strongly convex such that the corresponding linear SA has a unique fixed point. Note that $\lambda$ can be arbitrarily small positive constant. Such a stability condition is typically required for the analysis of linear SA and AC algorithms in the literature \cite{dalal2018finite,bhandari2018finite,konda1999actor,bhatnagar2009natural,konda2000actor}.

Hence, the update of $\theta_t$ is given by
\begin{align}
\theta_{t+1}=\mcpi_{R_\theta}(\theta_t+\beta_t g_t(\theta_t))
\end{align}
where $\beta_t>0$ is the stepsize, and $g_t(\theta_t)=(-\phi_{w_t}(s_t,a_t)^\top\theta_t+\widehat{Q}(s_t,a_t))\phi_{w_t}(s_t,a_t)-\lambda\theta_t$ serves as the stochastic estimation of the true gradient of $-L_w(\theta)$ in \cref{eq: solu1}. In particular, $\widehat{Q}(s_t,a_t))$ is an unbiased estimator of the true state-action value function $Q_\pi(s,a)$, and is obtained by $\text{Q-Sampling}(s,a,\pi)$ (see \Cref{algorithm_qsample}) proposed by \cite{zhang2019global}.

\begin{algorithm}[tb]
	\caption{Two Time-scale AC and NAC}
	\label{algorithm_ttsac}
	\begin{algorithmic}
		\STATE {\bfseries Input:} Parameterized policy $\pi_w$, actor stepsize $\alpha_t$, critic stepsize $\beta_t$, regularization constant $\lambda$
		\STATE {\bfseries Initialize:} actor parameter $w_0$, critic parameter $\theta_0$
		\FOR{$t=0,\cdots,T-1$}
		\STATE $s_t\sim \widetilde{\mathsf{P}}(\cdot |s_{t-1},a_{t-1})$
		\STATE Sample $a_t$ and $a^\prime_t$ independently from $\pi_{w_t}(\cdot|s_t)$
		\STATE $\widehat{Q}(s_t,a_t)=\text{Q-Sampling}(s_t,a_t,\pi_{w_t})$
		\STATE $g_t(\theta_t)=(-\phi_{w_t}(s_t,a_t)^\top\theta_t+\widehat{Q}(s_t,a_t))\phi_{w_t}(s_t,a_t)$
		$-\widehat{Q}(s_t,a_t)\phi_{w_t}(s_t,a^\prime_t) - \lambda\theta_t$
		\STATE  
		\STATE  \textit{\textbf{Critic Update: }}$\theta_{t+1}=\mcpi_{R_\theta}(\theta_t+\beta_t g_t(\theta_t))$
		\STATE
		\STATE \textit{\textbf{Option I: Actor update in AC}}
		\STATE $w_{t+1}=w_t+\alpha_t [\phi_{w_t}(s_t,a_t)^\top\theta_{t}]\phi_{w_t}(s_t,a_t)$
		\STATE
		\STATE \textit{\textbf{Option II: Actor update in NAC}}
		\STATE $w_{t+1}=w_t+\alpha_t \theta_{t}$
		\ENDFOR
		\STATE {\bfseries Output:} $w_{\hat{T}}$ with $\hat{T}$ chosen from distribution $P_{T}$
	\end{algorithmic}
\end{algorithm}

\begin{algorithm}[tb]
	\caption{$\text{Q-Sampling}(s,a,\pi)$}
	\label{algorithm_qsample}
	\begin{algorithmic}
		\STATE {\bfseries Initialize:} $\widehat{Q}_\pi(s,a)=0$, $s_0=s$ and $a_0=a$
		\STATE $T\sim \text{Geom}(1-\gamma^{1/2})$
		\FOR {$t=0,\cdots,T-1$}
		\STATE $s_{t+1}\sim \mathsf{P}(\cdot|s_t,a_t)$
		\STATE $\widehat{Q}_\pi(s,a)\leftarrow\widehat{Q}_\pi(s,a)+\gamma^{t/2}r(s_t,a_t,s_{t+1})$
		\STATE $a_{t+1}\sim \pi(\cdot|s_{t+1})$
		\ENDFOR
		\STATE {\bfseries Output:} $\widehat{Q}_\pi(s,a)$
	\end{algorithmic}
\end{algorithm}

The two time-scale natural actor-critic (NAC) (see Algorithm \ref{algorithm_ttsac}) is based on the natural policy gradient algorithm developed in \cite{bhatnagar2009natural,agarwal2019optimality}, which utilizes natural gradient ascent  \cite{amari1998natural,kakade2002natural} and guarantees that the policy update is invariant to the parameterization of the policy. At each step $t$, critic's update is the same as that in AC, but actor's update should take the form $w_{t+1}=w_t+\alpha_t F(w_t)^\dagger\nabla J(w_t),\label{eq: npg}$ as given in \cite{kakade2002natural},
where $F(w_t)$ is the Fisher information matrix defined as $F(w_t)\coloneqq\mE_{\nu_{\pi_{w_t}}}\big[ \phi_{w_t}(s,a) \phi_{w_t}(s,a)^\top \big]$, and $F(w_t)^\dagger$ represents the pseudoinverse of $F(w_t)$. Since the visitation distribution $\nu_{\pi_{w_t}}$ is usually unknown, the above update cannot be implemented in practice. As a solution, since critic approximately solves \cref{eq: solu1} due to the two time-scale nature of the algorithm, the minimum-norm solution of which satisfies $\theta^*_{w_t}=F(w_t)^\dagger\nabla J(w_t)\approx (F(w_t)+\lambda I)^{-1}\nabla J(w_t)\approx \theta_t$, the actor's update can be implemented as follows as given in \cite{agarwal2019optimality}
\begin{flalign*}
w_{t+1}=w_t+\alpha_t \theta_t.
\end{flalign*}


We next provide a few further comments on the two time-scale AC and NAC algorithms in \Cref{algorithm_ttsac}.
\begin{itemize}
\item We set the actor and critic's update stepsizes as $\alpha_t=\Theta(1/(t+1)^\sigma)$ and $\beta_t=\Theta(1/(t+1)^\nu)$, with $0<\nu<\sigma\leq 1$. Since $\alpha_t/\beta_t\rightarrow0$ as $t\rightarrow\infty$, $w_t$ is almost static with respect to $\theta_t$ asymptotically. If we treat $w_t$ as a fixed vector, then the critic is expected to track the fix point $\theta^*_{w_t}$ of the corresponding ODE. Thus, if $t$ is sufficiently large, we expect $\theta_t$ to be close to $\theta^*_{w_t}$.

\item Algorithm \ref{algorithm_ttsac} applies the transition kernel $\widetilde{\mathsf{P}}(\cdot|s,a)=\gamma \mathsf{P}(\cdot|s,a) + (1-\gamma)\zeta(\cdot)$, the stationary distribution of which has been shown in \cite{konda2002actor} to be $\nu_\pi(s,a)$ if the Markov chain is ergodic.


\item Critic's update includes the projection operator $\mcpi_{R_\theta}$ onto a norm ball with the radius satisfying $\max_{w\in \mR^d}\{ \theta^*_w \}\leq R_\theta=\mathcal{O}((1-\gamma)^{-1}\lambda^{-1})$. Here we use the projection in critic's update to prevent actor from taking a large step in a ``wrong" direction, which has been commonly adopted in \cite{konda2000actor,konda2002actor,wang2019neural}. 

\item Algorithm \ref{algorithm_ttsac} does not require the accessibility of the visitation distribution, as all state-action pairs are sampled sequentially by policy $\pi_{w_t}$ that changes dynamically as $w_t$ is updated.

\end{itemize}

\subsection{Technical Assumptions}

Our convergence analysis in this paper takes a few standard assumptions. 
\begin{assumption}\label{ass1}
	For any $w,w^\prime\in R^d$ and any state-action pair $(s,a)\in \mcs\times\mca$, there exist positive constants $L_\phi$, $C_\phi$, and $C_\pi$ such that the following hold:
	\begin{enumerate}
		\item $\ltwo{\phi_w(s,a)-\phi_{w^\prime}(s,a)}\leq L_{\phi}\ltwo{w-w^\prime}$,
		\item $\ltwo{\phi_w(s,a)}\leq C_\phi$,
		\item $\lTV{\pi_w(\cdot|s)-\pi_{w^\prime}(\cdot|s)}\leq C_\pi\ltwo{w-w^\prime}$, where $\lTV{\cdot}$ denotes the total-variation norm.
	\end{enumerate}
\end{assumption}
The first two items require the score function $\phi_w$ to be smooth and bounded, which hold for many policy classes such as Boltzman policy \cite{konda1999actor} and Gaussian policy \cite{doya2000reinforcement}. Such assumptions have also been often taken by the finite-time analysis of RL algorithms in \cite{kumar2019sample,zhang2019global,agarwal2019optimality,konda2002actor,zou2019finite}. The third item requires that the policy is Lipschitz with respect to the parameter $w$, which holds for any smooth policy with bounded action space or Gaussian policy. This has also been further justified in \cite[Lemma 1]{xu2020improving}.


The following assumption on the ergodicity of the Markov chain has been commonly adopted to establish the finite-sample analysis for RL algorithms \cite{bhandari2018finite,xu2020reanalysis,zou2019finite}, which holds for any time-homogeneous Markov chain with finite state space or any uniformly ergodic Markov chain with general state space.
\begin{assumption}[Ergodicity]\label{ass2}
	For any $w\in\mR^d$, consider the MDP with policy $\pi_w$ and transition kernel $\mathsf{P}(\cdot|s,a)$ or $\widetilde{\mathsf{P}}(\cdot|s,a)=\gamma \mathsf{P}(\cdot|s,a) + (1-\gamma)\eta(\cdot)$, where $\eta(\cdot)$ can either be $\xi(\cdot)$ or $\mathsf{P}(\cdot|\hat{s},\hat{a})$ for any given $(\hat{s},\hat{a})\in \mathcal{S}\times\mathcal{A}$. There exist constants $\kappa>0$ and $\rho\in(0,1)$ such that
	\begin{flalign*}
	\sup_{s\in\mcs}\lTV{\mP(s_t\in\cdot|s_0=s)-\chi_{\pi_w}}\leq \kappa\rho^t,\quad \forall t\geq 0,
	\end{flalign*}
	where $\chi_{\pi_w}$ is the stationary distribution of the corresponding MDP with transition kernel $\mathsf{P}(\cdot|s,a)$ or $\widetilde{\mathsf{P}}(\cdot|s,a)$ under policy $\pi_w$.
\end{assumption}

\section{Main Results}\label{sc: mainresults}
In this section, we first analyze the convergence of critic’s update as a linear SA with dynamically changing base function and transition kernel. Based on such an analysis, we further provide the convergence rate for the two time-scale AC and NAC algorithms.

\subsection{Convergence Analysis of Tracking Error of Critic}
A major challenge to analyze the sample complexity of two time-scale AC/NAC lies in characterizing the convergence rate of the fast time-scale (critic's) update. 
The following theorem provides the convergence rate of the tracking error of critic, which is defined as $\mE[\ltwo{\theta_t-\theta^{\lambda*}_{w_t}}^2]$, where $\theta^{\lambda*}_{w_t}=(F(w_t)+\lambda I)^{-1}\nabla J(w_t)$.
\begin{theorem}\label{critic_rate}
	Suppose Assumptions \ref{ass1} and \ref{ass2} hold. Consider two time-scale AC and NAC in Algorithm \ref{algorithm_ttsac}. We have
	\begin{flalign*}
	\mE\left[\ltwo{\theta_t-\theta^{\lambda*}_{w_t}}^2\right]=\left\{
	\begin{array}{lr}
	\mathcal{O}\big(\frac{\log^2 t}{(1-\gamma)^2t^\nu}\big), & \sigma\geq 1.5\nu, \\
	\mathcal{O}\big(\frac{1}{(1-\gamma)^2t^{2(\sigma-\nu)}}\big), &  \nu<\sigma<1.5\nu.
	\end{array}
	\right.
	\end{flalign*}
\end{theorem}
Theorem \ref{critic_rate} characterizes how the convergence rate of critic's tracking error depends on the stepsize. This result shares the same nature as that of two time-scale linear SA given in \cite{xu2019two}, in which the optimal convergence rate of the tracking error is also obtained when $\sigma=1.5\nu$, with an extra factor $\mathcal{O}(\log t)$ caused by the dynamically changing policy. However, the analysis of Theorem \ref{algorithm_ttsac} is more challenging than that in \cite{xu2019two} and requires the development of new techniques as we discuss in the proof sketch of \Cref{critic_rate} below. We relegate the detailed proof to \Cref{proof7}.
%

\begin{proof}[Proof Sketch of \Cref{critic_rate}]
The proof of \Cref{critic_rate} consists of three steps as we briefly describe as follows.

\textbf{Step 1.} \emph{Decomposing tracking error.} We decompose the tracking error $\mE[\ltwo{\theta_t-\theta^{\lambda*}_t}^2]$ into an exponentially decaying term, a variance term, a bias error term, a fixed-point shift error term, and a slow drift error term. 

\textbf{Step 2.} \emph{Bounding three error terms.} We bound the three error terms identified in Step 1. 
\begin{itemize}
\item[(a)] The bias error is caused by the correlation between samples due to the time-varying Markovian sampling. We develop a novel technique to characterize the relationship between the bias error and the dynamically changing policy and base functions. The bias error of dynamic linear SA has also been studied in \cite{zou2019finite}, but with fixed base function and a strong contraction-like assumption to force the algorithm to converge to a static fixed point. Thus, the proof in \cite{zou2019finite} is not applicable here. 

\item[(b)] The fixed-point shift error, i.e., the difference between $\theta^{\lambda*}_{w_t}$ and $\theta^{\lambda*}_{w_{t+1}}$, is caused by the dynamically changing base functions $\phi_{w_t}(s,a)$ and the dynamically changing transition kernel as $w_t$ updates. Such a type of error does not appear in the previously studied two time-scale RL algorithms such as in \cite{xu2019two} (in which both quantities are fixed). Thus, we develop new techniques to bound such an error, by analyzing the difference between visitation distributions and state-action value functions.



\item[(c)] The slow-drift error term due to the two time-scale nature of the algorithm can be bounded by adapting the techniques in \cite{xu2019two}. It terms out that such an error term dominates the convergence rate of the tracking error 
at the order of $\mathcal{O}(1/t^{\sigma-\nu})$.
\end{itemize}

\textbf{Step 3.} \emph{Recursively refining tracking error bound.} We further show that the slow-drift error term diminishes as the tracking error diminishes. By recursively substituting the preliminary bound of $\mE[\ltwo{\theta_t-\theta^{\lambda*}_t}^2]$ into the slow-drift term, we obtain the refined decay rate of the tracking error.
\end{proof}

\subsection{Convergence Analysis of Two Time-scale AC}

In order to analyze the two time-scale AC algorithm, the following Lipschitz gradient condition for $J(w)$ is important, which captures the tightest dependence of Lipschitz constant on $\mathcal{O}((1-\gamma)^{-1})$ among other studies, e.g., \cite{zhang2019global}.
\begin{lemma}[Proposition 1 in \cite{xu2020improving}]\label{lemma: sum}
	Suppose Assumptions \ref{ass1} and \ref{ass2} hold. For any $w,w^\prime\in \mR^d$, we have 
	$$\ltwo{\nabla_wJ(w)-\nabla_wJ(w^\prime)}\leq L_J\ltwo{w-w^\prime}, \quad \text{for all } \; w,w^\prime\in \mR^d,$$ 
	where $L_J=\frac{r_{\max}}{1-\gamma}(4C_\nu C_\phi+L_\phi)$ and $C_\nu=\frac{1}{2}C_\pi\left( 1 +  \lceil \log_\rho \kappa^{-1} \rceil + \frac{1}{1-\rho} \right)$.
\end{lemma}
We note that \Cref{lemma: sum} has been taken as the Lipschitz assumption in the previous studies of policy gradient and AC \cite{kumar2019sample,qiu2019finite,wang2019neural}. In our analysis, we adopt \Cref{lemma: sum} so that our results on the convergence rate explicitly reflect the dependence of $\mathcal{O}((1-\gamma)^{-1})$ via the Lipschitz constant.


Since the objective function $J(w)$ in \cref{eq:prob} is nonconcave in general, the convergence analysis of AC is with respect to the standard metric of $\mE\ltwo{\nabla_wJ(w)}^2$. 
The following theorem provides the complexity guarantee of two time-scale AC.
\begin{theorem}\label{thm3}
	Consider two time-scale AC in Algorithm \ref{algorithm_ttsac}. Suppose Assumptions \ref{ass1} and \ref{ass2} hold. Let $\nu=\frac{2}{3}\sigma$. Then the convergence rate of $\mE\ltwo{\nabla_w J(w_{\tilde{T}})}^2$ is given by
	\begin{flalign*}
	&\mE\ltwo{\nabla_w J(w_{\tilde{T}})}^2=\frac{C^3_\phi C_r r_{\max}}{1-\gamma}\lambda +\left\{
	\begin{array}{lr}
	\mathcal{O}\big(\frac{\log^2 T}{(1-\gamma)^2T^{1-\sigma}}\big), & \frac{3}{5}<\sigma\leq 1 , \\
	\mathcal{O}\big(\frac{\log^3 T}{(1-\gamma)^2T^{\frac{2}{5}}}\big), &  \sigma= \frac{3}{5}, \\
	\mathcal{O}\big(\frac{\log^2 T}{(1-\gamma)^2T^{\frac{2}{3}\sigma}}\big), &  0<\sigma<\frac{3}{5}. \\
	\end{array}
	\right.
	\end{flalign*}
	where $C_r$ (with its specific form given in \Cref{lemma: regularization}) is a positive constant depending on the policy $\pi_w$. Moreover, let $\sigma=\frac{3}{5}$ and $\nu=\frac{2}{5}$. Then the expected overall sample complexity of Algorithm \ref{algorithm_ttsac} to obtain $\mE\ltwo{\nabla_w J(w_{\tilde{T}})}^2\leq \epsilon + \mathcal{O}(\lambda)$ is given by
\begin{flalign*}
	T(N_Q+1)=\mathcal{O}\left(\frac{1}{(1-\gamma)^5\epsilon^{2.5}}\log^3\left(\frac{1}{\epsilon}\right)\right).
\end{flalign*}
\end{theorem}
\Cref{thm3} provides the sample complexity for two time-scale AC (with single sample for each update), and it outperforms the best known sample complexity \cite{qiu2019finite} for single-sample nested-loop AC by a factor of $\mathcal{O}(\frac{1}{\epsilon^{0.5}})$, indicating that two time-scale implementation of AC can be more efficient than nested-loop under single-sample update for each iteration.

Note that here actor's update also suffers from the bias error, because $(s_t, a_t)$ is strongly correlated with samples used in previous steps. 
To prove the convergence, we show that the bias error at $t$-th step can be upper bounded by $\mathcal{O}(\frac{\log^2(t)}{t^\sigma})$, and the accumulated bias error converges to zero at a rate of $\mathcal{O}(\frac{\log^2(T)}{T^{\frac{2}{5}}})$ under the optimal stepsize. 
Note that a similar bias error caused by dynamic Markovian sampling in nonconcave optimization has also been studied in \cite{karimi2019non}, which shows that the accumulated bias error can be upper bounded by a constant that diminishes as $\gamma$ increases. Whereas in Theorem \ref{thm3}, we show that such an accumulated bias converges to zero no matter how large $\gamma$ is, which is tighter than the bound given in \cite{karimi2019non}.

We provide a sketch of the proof of \Cref{thm3} below and relegate the detailed proof to \Cref{proof5}.
\begin{proof}[Proof Sketch of \Cref{thm3}]
The proof of \Cref{thm3} consists of three steps as we briefly describe as follows.

\textbf{Step 1.} \emph{Decomposing convergence error.} We show that the gradient $\mE[\ltwo{\nabla_w J(w_t)}^2]$ can be bounded by the difference between the objective function values, the bias error of actor's update, the tracking error (which has been bounded in \Cref{critic_rate}), and the variance error (which is bounded by a constant).

\textbf{Step 2.} \emph{Bounding bias error of actor's update.} We bound the bias error of actor's update as identified in Step 1. Such a bias error is caused by the correlation between samples due to the dynamically changing Markovian sampling. We develop a new proof to bound such a bias error in a nonlinear SA update due to the nonlinear parameterization of the policy, which is different from the bias error of linear SA that we studied in \Cref{critic_rate}.

\textbf{Step 3.} \emph{Analyzing convergence rate under various stepsizes.} We analyze the error bounds on the convergence rate under various stepsize settings for fast and slow time scales. 
It turns out the relative scaling of the stepsizes of the two time scales determines which error term dominates the final convergence rate, and we identify the dominating error terms for each setting.
\end{proof}

\subsection{Convergence Analysis of Two Time-scale NAC}
Our analysis of NAC is inspired by the analysis of natural policy gradient (NPG) in \cite{agarwal2019optimality}, but we here provide a finite sample analysis in the two time-scale and Markovian sampling setting.

Differently from AC algorithms, due to the parameter invariant property of the NPG update, we can establish the global convergence of NAC algorithm in terms of the function value convergence. As shown in \cite{agarwal2019optimality}, NPG is guaranteed to converge to a policy $\pi_{w_{\tilde{T}}}$ in the neighborhood of the global optimal $\pi^*$, which satisfies $J(\pi^*)-\mE\big[J(\pi_{w_{\tilde{T}}})\big] \leq\epsilon+\mathcal{O}(\sqrt{ \zeta^\prime_{\text{approx}}})$, where $\zeta^\prime_{\text{approx}}$ represents the approximation error of the compatible function class given by
\begin{flalign*}
\zeta^\prime_{\text{approx}} = \max_{w\in \mR^d} \min_{\theta\in \mR^d} \mE_{\nu_{\pi_{w}}}\big[ \phi_w(s,a)^\top \theta -  A_{\pi_{w}}(s,a) \big]^2.
\end{flalign*}
It can be shown that $\zeta^\prime_{\text{approx}}$ is zero or small if the express power of the policy class $\pi_w$ is large, e.g., tabular policy \cite{agarwal2019optimality} and overparameterized neural policy \cite{wang2019neural}.

The following theorem characterizes the convergence of two time-scale NAC in Algorithm \ref{algorithm_ttsac}. 
\begin{theorem}\label{thm4}
	Consider two time-scale NAC update in Algorithm \ref{algorithm_ttsac}. Suppose Assumptions \ref{ass1} and \ref{ass2} hold. Let $\nu=\frac{2}{3}\sigma$. Then we have
	\begin{flalign}
	&J(\pi^*)-\mE\big[J(\pi_{w_{\tilde{T}}})\big]\leq \sqrt{\frac{1}{(1-\gamma)^3}  \linf{\frac{\nu_{\pi^*}}{\nu_{\pi_{w_0}}}} \zeta_{\text{approx}}^\prime} + \frac{C_\phi C_{r} }{1-\gamma}\lambda + \left\{
	\begin{array}{lr}\label{eq: thm2}
	\mathcal{O}\big(\frac{\log T}{(1-\gamma)^2T^{1-\sigma}}\big), & \sigma\geq \frac{3}{4}, \\
	\mathcal{O}\big(\frac{\log^2 T}{(1-\gamma)^2T^{\frac{1}{4}}}\big), &  \sigma= \frac{3}{4}, \\
	\mathcal{O}\big(\frac{\log T}{(1-\gamma)^2T^{\frac{1}{3}\sigma}}\big), &  \sigma<\frac{3}{4}. \\
	\end{array}
	\right.
	\end{flalign}
	where $C_r$ (with its specific form given in \Cref{lemma: regularization}) is a positive constant depending on the policy $\pi_w$. Moreover, let $\sigma=\frac{3}{4}$, $\nu=\frac{1}{2}$ and $\lambda=\mathcal{O}(\sqrt{\zeta_{\text{approx}}^\prime})$. Then the expected overall sample complexity of Algorithm \ref{algorithm_ttsac} to obtain $J(\pi^*)-\mE\big[J(\pi_{w_{\tilde{T}}})\big]\leq \epsilon + \mathcal{O}(\sqrt{\zeta_{\text{approx}}^\prime})$ is given by 
\begin{flalign*}
	TN_Q=\mathcal{O}\left(\frac{1}{(1-\gamma)^9\epsilon^4}\log^2\frac{1}{\epsilon}\right).
\end{flalign*}
\end{theorem}

Theorem \ref{thm4} provides the first non-asymptotic convergence rate for two time-scale NAC in terms of the function value. In previous studies, two time-scale NAC was only shown to converge to a first-order stationary point under i.i.d. sampling \cite{bhatnagar2009natural} without characterization of the convergence rate. Theorem \ref{thm4} considers the more general Markovian sampling and establishes the convergence to the global optimal point. Furthermore, our analysis is non-asymptotic and provides how the convergence rate of the function value $J(w)$ depends on the diminishing stepsizes of actor and critic.
The sample complexity for two time-scale NAC given in Theorem \ref{thm4} is almost the same as that for nested-loop NPG given in \cite{agarwal2019optimality} (Corollary 6.10). The extra $\log^2(\frac{1}{\epsilon})$ term is due to the bias error introduced by Markovian sampling, whereas \cite{agarwal2019optimality} analyzes i.i.d.\ sampling. 

We provide a sketch of the proof of \Cref{thm4} below and relegate the detailed proof to \Cref{proof1}.
\begin{proof}[Proof Sketch of \Cref{thm4}]
The proof of \Cref{thm4} consists of two steps as we briefly describe as follows.

\textbf{Step 1.} \emph{Decomposing convergence error.} We show that the incremental change of the objective function values can be bounded by the changes of the KL-distance between the iterating policy and globally optimal policy, tracking error (which has been bounded in \Cref{critic_rate}), non-vanishing approximation error, and the variance error (which is upper bounded by a constant).

\textbf{Step 2.} \emph{Analyzing convergence rate under various stepsizes.} We analyze the error bounds on the convergence rate under various stepsize settings for fast and time time scales. It turns out the relative scaling of the stepsizes of the two time scales determines which error term dominates the final convergence rate, and we identify the dominating error terms for each setting.
\end{proof}

\section{Conclusion}
In this paper, we provided the first non-asymptotic convergence analysis for two time-scale AC and NAC algorithms under Markovian sampling. In particular, we showed that two time-scale AC converges to a first-order stationary point, and two time-scale NAC converges to a neighborhood of the globally optimal solution. We showed that the overall sample complexity of two time-scale AC outperforms the best existing result of single-sample nested-loop AC by a factor of $\mathcal{O}(\frac{1}{\epsilon^{0.5}})$ \cite{qiu2019finite}, and the overall sample complexity of two time-scale NAC is as good as that of nested-loop NAC \cite{agarwal2019optimality}. We developed new techniques to analyze the bias errors of linear and nonlinear SA, with dynamically changing base functions and a time-varying transition kernel.
Our techniques can be further applied to study other two time-scale RL algorithms in the future.

\newpage
\appendix
\noindent {\Large \textbf{Appendices}}
\section{Supporting Lemmas}\label{sc: list_lemma_2}

In this section, we provide supporting lemmas which are useful for the proof of the main theorems. The detailed proofs of these lemmas are relegated to Section \ref{sc: lemmaproof2}.

\begin{lemma}[Lemma 2 in \cite{xu2020improving}]\label{lemma: visit_dis}
	Considering the initialization distribution $\eta(\cdot)$ and transition kernel $\mathsf{P}(\cdot|s,a)$. Let $\eta(\cdot)=\zeta(\cdot)$ or $\mathsf{P}(\cdot|\hat{s},\hat{a})$ for any given $(\hat{s},\hat{a})\in \mathcal{S}\times\mathcal{A}$. Denote $\nu_{\pi_w,\eta}(\cdot,\cdot)$ as the state-action visitation distribution of MDP with policy $\pi_w$ and the initialization distribution $\eta(\cdot)$. Suppose Assumption \ref{ass2} holds. Then we have 
	\begin{flalign*}
	\lTV{\nu_{\pi_w,\eta}-\nu_{\pi_{w^\prime},\eta}}\leq C_\nu\ltwo{w-w^\prime}
	\end{flalign*}
	for all $w,w^\prime\in \mR^d$, where $C_\nu=C_\pi\left( 1 +  \lceil \log_\rho m^{-1} \rceil + \frac{1}{1-\rho} \right)$.
\end{lemma}

\begin{lemma}[Lemma 3 in \cite{xu2020improving}]\label{valuefunction}
	Suppose Assumptions \ref{ass1} and \ref{ass2} hold, for any $w,w^\prime\in \mR^d$ and any state-action pair $(s,a)\in \mathcal{S}\times\mathcal{A}$. We have
	\begin{flalign*}
	\lone{Q_{\pi_w}(s,a) - Q_{\pi_{w^\prime}}(s,a)}\leq L_Q\ltwo{w-w^\prime},
	\end{flalign*} 
	and
	\begin{flalign*}
	\lone{V_{\pi_w}(s) - V_{\pi_{w^\prime}}(s)}\leq L_V\ltwo{w-w^\prime},
	\end{flalign*}
	where $L_Q=\frac{2r_{\max}C_\nu}{1-\gamma}$ and $L_V=\frac{r_{\max} (C_\pi + 2C_\nu)}{1-\gamma}$.
\end{lemma}

\begin{lemma}[Lemma 4 in \cite{xu2020improving}]\label{lemma: liptz2}
	There exists a constant $L_\phi$ such that 
	\begin{flalign*}
	\ltwo{\nabla_w \mE_{\nu_{\pi^*}}\Big[\log \pi_w(a,s)\Big]-\nabla_w \mE_{\nu_{\pi^*}}\Big[\log \pi_{w^\prime}(a,s)\Big]}\leq L_\phi\ltwo{w-w^\prime},
	\end{flalign*}
	holds for all $w,w^\prime \in \mR^d$.
\end{lemma}

\begin{lemma}\label{lemma: variance}
	Suppose Assumption \ref{ass1} hold. For any $t\geq 0$, we have
	\begin{flalign*}
		\ltwo{g_t(\theta_t)}^2\leq C_1\ltwo{\theta_t-\theta^{\lambda*}_t}^2+C_2,
	\end{flalign*}
	where $C_1=2(C_\phi+\lambda)^2$ and $C_2=2\big[(C_\phi+\lambda)R_\theta+ \frac{4C_\phi r_{\max}}{1-\gamma} \big]^2$.
\end{lemma}

\begin{lemma}\label{lemma: fixtracking}
	Suppose Assumptions \ref{ass1} and \ref{ass2} hold. For any $w,w^\prime\in \mR^d$, we have
	\begin{flalign*}
		\ltwo{\theta^{\lambda*}_{w}-\theta^{\lambda*}_{w^\prime}}\leq L_\theta\ltwo{w-w^\prime},
	\end{flalign*}
	 where $L_\theta= \frac{r_{\max}}{\lambda_P(1-\gamma)} \Big[ 6C_\phi C_\nu +  L_\phi + C_\phi C_\pi  + \frac{2C^2_\phi}{\lambda_P} (L_\phi+C_\phi C_\nu)\Big] $.
\end{lemma}

\begin{lemma}\label{lemma: support1}
	Consider $\hat{t}>0$ such that $(C_\phi^2+\lambda)\beta_{t-{\tau_t}}\tau^2_t\leq \frac{1}{4}$  for all $t>\hat{t}$. Then, for $0<\hat{t}<t$, we have 
	\begin{flalign*}
	\ltwo{\theta_t-\theta^{\lambda*}_t}\leq \Big( 1+2\beta_{t-{\tau_t}}\tau_t(C_\phi^2+\lambda)\Big)\ltwo{\theta_{t-\tau_t}-\theta^{\lambda*}_{t-\tau_t}} + 2C_4\beta_{t-{\tau_t}}\tau_t,
	\end{flalign*}
	with $C_4=(C_\phi^2+\lambda)R_\theta+\frac{2C_\phi r_{\max}}{1-\gamma} + \frac{C_3C_\alpha}{C_\beta}$.
\end{lemma}

\begin{lemma}\label{lemma: support2}
	For $0<\hat{t}<t$, we have
	\begin{flalign*}
		\ltwo{\theta_t-\theta_{t-\tau_t}}\leq \frac{3}{2}\beta_{t-{\tau_t}}\tau_t(C_\phi^2+\lambda)\ltwo{\theta_{t-\tau_t}-\theta^{\lambda*}_{t-\tau_t}}  + C_5\beta_{t-{\tau_t}}\tau_t,
	\end{flalign*}
	 where $C_5=\Big(\frac{1}{2}C_4+(C_\phi^2+\lambda)R_\theta+\frac{2C_\phi r_{\max}}{1-\gamma}\Big)$, and $\hat{t}$ is a positive constant that when satisfying $(C_\phi^2+\lambda)\beta_{t-{\tau_t}}\tau^2_t\leq \frac{1}{4}$ for all $t>\hat{t}$.
\end{lemma}

\begin{lemma}\label{lemma: support3}
	For any $\hat{t}>0$, we have
	\begin{flalign*}
		\ltwo{\theta_{\hat{t}}-\theta^{\lambda*}_{\hat{t}}}^2 \leq C_{16}\ltwo{\theta_0-\theta^{\lambda*}_0}^2 + C_{17},
	\end{flalign*}
	where $C_{16}=3 + \frac{27}{2}\beta^2_0\hat{t}^2(C_\phi^2+\lambda)^2$ and $C_{17}=3C^2_3 R^2_\theta \max\{1,C^4_\phi\} C^2_\alpha \hat{t}^2  + 6C^2_5\beta^2_0\hat{t}^2$.
\end{lemma}

\begin{lemma}\label{bias_supp1}
	For any $w, w^\prime \in \mR^d$ and any $(s,a)\in \mcs\times\mca$, we have 
	\begin{flalign}
	\ltwo{P^\lambda_{w}(s,a) - P^\lambda_{w^\prime}(s,a)}\leq 2C_\phi L_\phi \ltwo{w-w^\prime},
	\end{flalign}
	and
	\begin{flalign}
	\ltwo{b_{w}(s,a) - b_{w^\prime}(s,a)} \leq \left[\frac{L_\phi r_{\max}}{1-\gamma} + C_\phi (L_Q + L_V) \right] \ltwo{w-w^\prime}.
	\end{flalign}
\end{lemma}

\begin{lemma}\label{lemma: delta}
	For all $t>\hat{t}$, we have
	\begin{flalign*}
		\Delta_{P,\tau_t}=\ltwo{\overline{P}^\lambda_{w_{t-\tau_t}}-\mE\big[P^\lambda_{w_{t-\tau_t}}|\mf_{t-\tau_t}\big]}\leq C_6\alpha_{t-\tau_t}\tau^2_t,
	\end{flalign*}
	and
	\begin{flalign*}
		\Delta_{b,\tau_t}=\ltwo{\overline{b}_{w_{t-\tau_t}}-\mE\big[b_{w_{t-\tau_t}}|\mf_{t-\tau_t}\big]}\leq  C_7\alpha_{t-\tau_t}\tau^2_t,
	\end{flalign*}
	where $C_6=2C^2_\phi\big[ 2C_\pi R_\theta\max\{1,C^2_\phi\} + 1\big]$ and $C_7=\frac{2C_\phi r_{\max}}{1-\gamma}\big[2C_\pi R_\theta\max\{1,C^2_\phi\} + 1\big]$.
\end{lemma}

\begin{lemma}\label{bias_supp2}
	For any $w, w^\prime \in \mR^d$ and any $(s,a)\in \mcs\times\mca$, we have
	\begin{flalign*}
	\ltwo{\overline{P}^\lambda_{w}(s,a) - \overline{P}^\lambda_{w^\prime}(s,a)}\leq 2(C_\phi^2C_\nu + C_\phi L_\phi) \ltwo{w-w^\prime},
	\end{flalign*}
	and 
	\begin{flalign*}
	\ltwo{\overline{b}_{w}(s,a) - \overline{b}_{w^\prime}(s,a)} \leq \left[ \frac{2C_\phi C_\nu r_{\max}}{1-\gamma} + \frac{L_\phi r_{\max}}{1-\gamma} + C_\phi (L_Q + L_V) \right]\ltwo{w-w^\prime}.
	\end{flalign*}
\end{lemma}

\begin{lemma}\label{lemma: bias}
	For all $t>\hat{t}$, we have
	\begin{flalign*}
		&\mE[\xi_t(\theta_t)|\mf_{t-\tau_t}] \nonumber\\
		&= \mE[\langle g_t(\theta_t)-\overline{g}_t(\theta_t), \theta_t-\theta^{\lambda*}_t \rangle|\mf_{t-\tau_t}]\leq C_{12}\beta_{t-{\tau_t}}\tau^2_t\mE\left[\ltwo{\theta_t-\theta^{\lambda*}_t}^2|\mf_{t-\tau_t}\right] + C_{13}\beta_{t-{\tau_t}}\tau^2_t,
	\end{flalign*}
	 where $0<C_{12}<\infty$ and $0<C_{13}<\infty$ are constants independent of $t$.
\end{lemma}

\begin{lemma}\label{lemma: accumulate1}
	It holds that
	\begin{flalign*}
	\sum_{i=\hat{t}}^{t-1} e^{-\lambda_P \sum_{k=i+1}^{t-1}\beta_k} \beta^2_i\tau^2_i\leq C_{18}\tau^2_t e^{-\frac{\lambda_P C_\beta }{2(1-\nu)} [(t+1)^{1-\nu} - (\hat{t}+1)^{1-\nu}]} + C_{19} \tau^2_t \beta_{t-1},
	\end{flalign*}
	where $C_{18}=\frac{ 2 e^{ \frac{\lambda_P C_\beta}{2} } }{ \lambda_P }\max_{i\in [\hat{t},i_\beta-1]}\{ e^{\frac{\lambda_P}{2} \sum_{k=i+1}^{t-1}\beta_k} \beta_i \}$ and $C_{19}=\frac{ 2 e^{ \frac{\lambda_P C_\beta}{2} } }{ \lambda_P }$ with $i_\beta =  (\frac{2\nu}{C_\beta \lambda_P})^{\frac{1}{1-\nu}}$.
\end{lemma}

\begin{lemma}\label{lemma: accumulate2}
	It following that
	\begin{flalign*}
	\sum_{i=\hat{t}}^{t-1} e^{-\lambda_P \sum_{k=i+1}^{t-1}\beta_k} \alpha_i\leq C_{20}e^{-\frac{\lambda_P C_\beta }{2(1-\nu)} [(t+1)^{1-\nu} - (\hat{t}+1)^{1-\nu}]} + \frac{C_{21}}{t^{\sigma-\nu}}
	\end{flalign*}
	where
	\begin{flalign*}
		C_{20}= \frac{ 2C_\alpha e^{ \frac{\lambda_P C_\beta}{2} } }{ C_\beta\lambda_P }\max_{i\in [\hat{t},i_\alpha-1]}\Big\{ e^{\frac{\lambda_P}{2} \sum_{k=i+1}^{t-1}\beta_k} \frac{1}{(1+i)^{\sigma-\nu}} \Big\},
	\end{flalign*}
	 with $i_\alpha =  (\frac{2(\sigma-\nu)}{C_\beta \lambda_P})^{\frac{1}{1-(\sigma-\nu)}}$, and $C_{21}=\frac{ 2C_\alpha e^{ \frac{\lambda_P C_\beta}{2} } }{ C_\beta\lambda_P}$.
\end{lemma}

\begin{lemma}[Lemma 5 in \cite{xu2020improving}]\label{lemma: regularization}
	For any $w\in\mR^d$, define $\theta^{\lambda*}_w= (F(w)+\lambda)^{-1}\nabla J(w) $ and $\theta^{*}_w=F(w)^\dagger\nabla J(w)$. We have $\ltwo{\theta^{*}_w - \theta^{\lambda*}_w}\leq C_\lambda \lambda$, where $0<C_\lambda<\infty$ is a constant independent from $\lambda$.
\end{lemma}

\section{Proof of Theorem \ref{critic_rate}}\label{proof7}

We provide the proof of Theorem \ref{critic_rate} in three major steps.

\textbf{Step 1.} \emph{Decomposing tracking error.} We define the tracking error as $(\theta_t-\theta^{\lambda*}_t)$, and bound the recursion of the tracking error as follows. For any $t\geq 0$, we derive
\begin{flalign}
&\ltwo{\theta_{t+1}-\theta^{\lambda*}_{t+1}}^2\nonumber\\
&=\ltwo{\mcpi_{R_\theta}(\theta_t+\beta_t g_t(\theta_t))-\theta^{\lambda*}_{t+1}}^2\nonumber\\
&=\ltwo{\mcpi_{R_\theta}(\theta_t+\beta_t g_t(\theta_t))-\mcpi_{R_\theta}(\theta^{\lambda*}_{t+1})}^2\nonumber\\
&\overset{(i)}{\leq}\ltwo{\theta_t+\beta_t g_t(\theta_t)-\theta^{\lambda*}_{t+1}}^2\nonumber\\
&=\ltwo{\theta_t-\theta^{\lambda*}_t+\beta_t g_t(\theta_t) + \theta^{\lambda*}_t -\theta^{\lambda*}_{t+1}}^2\nonumber\\
&=\ltwo{\theta_t-\theta^{\lambda*}_t}^2 + 2\beta_t \langle g_t(\theta_t), \theta_t-\theta^{\lambda*}_t \rangle + 2\langle \theta_t-\theta^{\lambda*}_t, \theta^{\lambda*}_t -\theta^{\lambda*}_{t+1} \rangle + 2\beta_t \langle g_t(\theta_t), \theta^{\lambda*}_t -\theta^{\lambda*}_{t+1} \rangle \nonumber\\
&\quad + \ltwo{\theta^{\lambda*}_t -\theta^{\lambda*}_{t+1}}^2 + \beta_t^2\ltwo{g_t(\theta_t)}^2\nonumber\\
&=\ltwo{\theta_t-\theta^{\lambda*}_t}^2 + 2\beta_t \langle \overline{g}_t(\theta_t), \theta_t-\theta^{\lambda*}_t \rangle + 2\beta_t \langle g_t(\theta_t)-\overline{g}_t(\theta_t), \theta_t-\theta^{\lambda*}_t \rangle + 2\langle \theta_t-\theta^{\lambda*}_t, \theta^{\lambda*}_t -\theta^{\lambda*}_{t+1} \rangle \nonumber\\
&\quad + 2\beta_t \langle g_t(\theta_t), \theta^{\lambda*}_t -\theta^{\lambda*}_{t+1} \rangle + \ltwo{\theta^{\lambda*}_t - \theta^{\lambda*}_{t+1}}^2 + \beta_t^2\ltwo{g_t(\theta_t)}^2\nonumber\\
&\overset{(ii)}{\leq} (1-2\beta_t\lambda_P)\ltwo{\theta_t-\theta^{\lambda*}_t}^2 + 2\beta_t \langle g_t(\theta_t)-\overline{g}_t(\theta_t), \theta_t-\theta^{\lambda*}_t \rangle + 2\langle \theta_t-\theta^{\lambda*}_t, \theta^{\lambda*}_t -\theta^{\lambda*}_{t+1} \rangle \nonumber\\
&\quad + 2\beta_t \langle g_t(\theta_t), \theta^{\lambda*}_t -\theta^{\lambda*}_{t+1} \rangle + \ltwo{\theta^{\lambda*}_t - \theta^{\lambda*}_{t+1}}^2 + \beta_t^2\ltwo{g_t(\theta_t)}^2,\label{part1}
\end{flalign}
where $(i)$ follows from the expansive property of the projection operator $\mcpi_{R_\theta}(\cdot)$, $(ii)$ follows from the fact that
\begin{flalign*}
	\langle \overline{g}_t(\theta_t), \theta_t-\theta^{\lambda*}_t \rangle=\langle \overline{P}^\lambda_{w_t}(\theta_t-\theta^{\lambda*}_t), \theta_t-\theta^{\lambda*}_t \rangle\leq \lambda_P\ltwo{\theta_t-\theta^{\lambda*}_t}^2.
\end{flalign*}

\textbf{Step 2.} \emph{Bounding three error terms.} In \cref{part1}, we decompose the tracking error into an exponentially decaying term, a variance term, a bias error term, a fixed-point shift error term, and a slow drift error term, and now we bound each term individually.

The third error term in \cref{part1} is the slow drift term, which can be upper bounded as following:
\begin{flalign}
\langle \theta_t-\theta^{\lambda*}_t, \theta^{\lambda*}_t -\theta^{\lambda*}_{t+1} \rangle &\leq \ltwo{\theta_t-\theta^{\lambda*}_t} \ltwo{\theta^{\lambda*}_t -\theta^{\lambda*}_{t+1}}\overset{(i)}{\leq} L_\theta R_\theta \max\{1, C^2_\phi \} \alpha_t \ltwo{\theta_t -\theta^{\lambda*}_{t}}\nonumber\\
&\overset{(ii)}{\leq} \frac{1}{2}L_\theta R_\theta \max\{1, C^2_\phi \} \alpha_t (\ltwo{\theta_t -\theta^{\lambda*}_{t}}^2+1), \label{part1_subbound1}
\end{flalign}
where $(i)$ follows from Lemma \ref{lemma: fixtracking} and $(ii)$ follows from the fact that $x\leq \frac{1}{2}(x^2+1)$ for all $x$. The slow-drift error term is caused by the two time-scale nature of the algorithm and it terms out that this error term diminishes as the tracking error diminishes. 

The forth term in \cref{part1} is the fixed-point shift error term. This error is caused by the dynamically changing base functions $\phi_{w_t}(s,a)$ and the dynamically changing transition kernel as $w_t$ updates. We derive an upper bound as following:
\begin{flalign}
&\langle g_t(\theta_t), \theta^{\lambda*}_t -\theta^{\lambda*}_{t+1} \rangle \nonumber\\
&= \langle -P^\lambda_{w_t}\theta_t + b_{w_t}, \theta^{\lambda*}_t -\theta^{\lambda*}_{t+1}  \rangle \nonumber\\
&= \langle -P^\lambda_{w_t}(\theta_t-\theta^{\lambda*}_t), \theta^{\lambda*}_t -\theta^{\lambda*}_{t+1}  \rangle + \langle -P^\lambda_{w_t}\theta^{\lambda*}_t + b_{w_t}, \theta^{\lambda*}_t -\theta^{\lambda*}_{t+1}  \rangle \nonumber\\
&\leq \ltwo{P^\lambda_{w_t}}\ltwo{\theta_t-\theta^{\lambda*}_t}\ltwo{\theta^{\lambda*}_t -\theta^{\lambda*}_{t+1}} + (\ltwo{P^\lambda_{w_t}}\ltwo{\theta^{\lambda*}_t}+\ltwo{b_{w_t}})\ltwo{\theta^{\lambda*}_t -\theta^{\lambda*}_{t+1}}\nonumber\\
&\overset{(i)}{\leq} (C_\phi^2+\lambda)L_\theta R_\theta \max\{1, C^2_\phi \}\alpha_t\ltwo{\theta_t-\theta^{\lambda*}_t} + \left[(C_\phi^2+\lambda)R_\theta + \frac{2r_{\max}C_\phi}{1-\gamma} \right]L_\theta R_\theta \max\{1, C^2_\phi \}\alpha_t\nonumber\\
&\leq \frac{1}{2} (C_\phi^2+\lambda)L_\theta R_\theta \max\{1, C^2_\phi \}\alpha_t(\ltwo{\theta_t-\theta^{\lambda*}_t}^2+1) + \left[(C_\phi^2+\lambda)R_\theta + \frac{2r_{\max}C_\phi}{1-\gamma} \right]L_\theta R_\theta \max\{1, C^2_\phi \}\alpha_t \nonumber\\
&\overset{(ii)}{=}\frac{1}{2} (C_\phi^2+\lambda)C_3\alpha_t \ltwo{\theta_t-\theta^{\lambda*}_t}^2 + \left[(C_\phi^2+\lambda)(R_\theta+\frac{1}{2}) + \frac{2r_{\max}C_\phi}{1-\gamma} \right]C_3\alpha_t, \label{part1_subbound2}
\end{flalign}
where $(i)$ follows from Lemma \ref{lemma: fixtracking}, in which we derive the $L_\theta$-Lipschitz condition for the fixed-point. In $(ii)$ we define $C_3=L_\theta R_\theta \max\{1, C^2_\phi \}$. Substituting \cref{part1_subbound1} and \cref{part1_subbound2} to \cref{part1} and applying Lemma \ref{lemma: fixtracking} and Lemma \ref{lemma: variance} yield the following,
\begin{flalign}
\ltwo{\theta_{t+1}-\theta^{\lambda*}_{t+1}}^2
&\overset{(i)}{\leq} (1-2\beta_t\lambda_P)\ltwo{\theta_t-\theta^{\lambda*}_t}^2 + 2\beta_t \xi_t(\theta_t) + C_3\alpha_t (\ltwo{\theta_t -\theta^{\lambda*}_t}^2+1) \nonumber \\
&\quad + (C_\phi^2+\lambda)C_3\alpha_t\beta_t \ltwo{\theta_t-\theta^{\lambda*}_t}^2 +  2\left[(C_\phi^2+\lambda)(R_\theta+\frac{1}{2}) + \frac{2r_{\max}C_\phi}{1-\gamma} \right]C_3\alpha_t\beta_t \nonumber\\
&\quad + C_3^2\alpha_t^2 + \beta_t^2\big(C_1\ltwo{\theta_t-\theta^{\lambda*}_t}^2 + C_2\big) \nonumber\\
&\leq \Big(1-2\beta_t\lambda_P + C_3\alpha_t + (C_\phi^2+\lambda)C_3\alpha_t\beta_t + C_1\beta_t^2\Big) \ltwo{\theta_t-\theta^{\lambda*}_t}^2 + \beta_t \zeta_t(\theta_t) \nonumber\\
&\quad + C_3\alpha_t + 2\left[(C_\phi^2+\lambda)(R_\theta+\frac{1}{2}) + \frac{2r_{\max}C_\phi}{1-\gamma} \right]C_3\alpha_t\beta_t + C_3^2\alpha_t^2+C_2\beta_t^2, \label{part1_sub2}
\end{flalign}
where in $(i)$ we define the bias error as $\xi_t(\theta_t) = \langle g_t(\theta_t)-\overline{g}_t(\theta_t), \theta_t-\theta^{\lambda*}_t \rangle$. Taking expectation on both sides of \cref{part1_sub2} conditioned on the filtration $\mf_{t-\tau_t}$, we have
\begin{flalign}
&\mE\big[\ltwo{\theta_{t+1}-\theta^{\lambda*}_{t+1}}^2|\mf_{t-\tau_t}\big] \nonumber\\
&\leq \Big(1-2\beta_t\lambda_P + C_3\alpha_t + (C_\phi^2+\lambda)C_3\alpha_t\beta_t + C_1\beta_t^2\Big) \mE\big[\ltwo{\theta_t-\theta^{\lambda*}_t}^2|\mf_{t-\tau_t}\big] + \beta_t  \mE\big[\xi_t(\theta_t)|\mf_{t-\tau_t}\big] \nonumber\\
&\quad + C_3\alpha_t + 2\left[(C_\phi^2+\lambda)(R_\theta+\frac{1}{2}) + \frac{2r_{\max}C_\phi}{1-\gamma} \right]C_3\alpha_t\beta_t + C_3^2\alpha_t^2+C_2\beta_t^2\nonumber\\
& \overset{(i)}{\leq}  \Big(1-2\beta_t\lambda_P + C_3\alpha_t + (C_\phi^2+\lambda)C_3\alpha_t\beta_t + C_1\beta_t^2\Big) \mE\big[\ltwo{\theta_t-\theta^{\lambda*}_t}^2|\mf_{t-\tau_t}\big]   \nonumber\\
&\quad + C_{12}\beta_t\beta_{t-{\tau_t}}\tau^2_t\mE\left[\ltwo{\theta_t-\theta^{\lambda*}_t}^2|\mf_{t-\tau_t}\right] + C_{13}\beta_t\beta_{t-{\tau_t}}\tau^2_t + C_3\alpha_t \nonumber\\
&\quad + 2\left[(C_\phi^2+\lambda)(R_\theta+\frac{1}{2}) + \frac{2r_{\max}C_\phi}{1-\gamma} \right]C_3\alpha_t\beta_t + C_3^2\alpha_t^2+C_2\beta_t^2 \nonumber\\
&\leq  \Big(1-2\beta_t\lambda_P + C_3\alpha_t + (C_\phi^2+\lambda)C_3\alpha_t\beta_t + C_1\beta_t^2 + C_{12}\beta^2_{t-{\tau_t}}\tau^2_t\Big) \mE\big[\ltwo{\theta_t-\theta^{\lambda*}_t}^2|\mf_{t-\tau_t}\big]  \nonumber\\
&\quad + \left[C_2 + C_{13} + \frac{C_3^2 C_\alpha^2}{C_\beta^2} +2(C_\phi^2+\lambda)(R_\theta+\frac{1}{2})\frac{C_3 C_\alpha}{C_\beta} + \frac{4r_{\max}C_\phi C_3 C_\alpha}{(1-\gamma)C_\beta} \right]\beta^2_{t-\tau_t}\tau^2_t + C_3\alpha_t.\nonumber\\
&\leq \left(1-2\beta_t\lambda_P + C_3\alpha_t + \left[\frac{(C_\phi^2+\lambda)C_\phi C_3}{C_\beta} + C_1 + C_{12}\right]\beta^2_{t-{\tau_t}}\tau^2_t\right) \mE\big[\ltwo{\theta_t-\theta^{\lambda*}_t}^2|\mf_{t-\tau_t}\big]  \nonumber\\
&\quad + \left[C_2 + C_{13} + \frac{C_3^2 C_\alpha^2}{C_\beta^2} +2(C_\phi^2+\lambda)(R_\theta+\frac{1}{2})\frac{C_\alpha C_3}{C_\beta} + \frac{4r_{\max}C_\phi C_3 C_\alpha}{(1-\gamma)C_\beta} \right]\beta^2_{t-\tau_t}\tau^2_t + C_3\alpha_t \nonumber\\
&\overset{(ii)}{=}\left(1-2\beta_t\lambda_P + C_3\alpha_t + C_{14}\beta^2_{t-{\tau_t}}\tau^2_t\right) \mE\big[\ltwo{\theta_t-\theta^{\lambda*}_t}^2|\mf_{t-\tau_t}\big] + C_{15}\beta^2_{t-\tau_t}\tau^2_t + C_3\alpha_t\nonumber\\
&\overset{(iii)}{\leq} \left(1-\beta_t\lambda_P \right) \mE\big[\ltwo{\theta_t-\theta^{\lambda*}_t}^2|\mf_{t-\tau_t}\big] + 4C_{15}\beta^2_t\tau^2_t + C_3\alpha_t\label{part1_sub3}
\end{flalign}
where $(i)$ follows from Lemma \ref{lemma: bias}, in which we derive an upper bound for the bias error. In $(ii)$ we define $C_{14}=\left[\frac{(C_\phi^2+\lambda)C_\phi C_3}{C_\beta} + C_1 + C_{12}\right]$ and $C_{15} = C_2 + C_{13} + \frac{C_3^2 C_\alpha^2}{C_\beta^2} +2(C_\phi^2+\lambda)(R_\theta+\frac{1}{2})\frac{C_3 C_\alpha }{C_\beta} + \frac{4r_{\max}C_\phi C_3 C_\alpha}{(1-\gamma)C_\beta} $. $(iii)$ follows from the fact that $C_3\alpha_t + C_{14}\beta^2_{t-{\tau_t}}\tau^2_t\leq \beta_t\lambda_\theta$ and $\beta_{t-{\tau_t}}\leq 2\beta_t$ for all $t>\hat{t}$. Taking expectation on both sides of \cref{part1_sub3} yields
\begin{flalign}
\mE\big[\ltwo{\theta_{t+1}-\theta^{\lambda*}_{t+1}}^2 \big] \leq \left(1-\beta_t\lambda_P \right) \mE\big[\ltwo{\theta_t-\theta^{\lambda*}_t}^2\big] + 4C_{15}\beta^2_t\tau^2_t + C_3\alpha_t.\label{part1_sub4}
\end{flalign}
Applying \cref{part1_sub4} recursively yields
\begin{flalign}
\mE\big[\ltwo{\theta_t-\theta^{\lambda*}_t}^2\big] &\leq \left[ \prod_{i=\hat{t}}^{t-1}(1-\beta_i\lambda_P) \right]\mE\big[\ltwo{\theta_{\hat{t}}-\theta^{\lambda*}_{\hat{t}}}^2 \big]+ 4C_{15}\sum_{i=\hat{t}}^{t-1} \left[ \prod_{k=i+1}^{t-1}(1-\beta_k\lambda_P) \right]\beta^2_i\tau^2_i\nonumber\\
&\quad + C_3\sum_{i=\hat{t}}^{t-1} \left[ \prod_{k=i+1}^{t-1}(1-\beta_k\lambda_P) \right]\alpha_i\nonumber\\
&\overset{(i)}{\leq} \left[ \prod_{i=\hat{t}}^{t-1}(1-\beta_i\lambda_P) \right] \Big(C_{16}\ltwo{\theta_0-\theta^{\lambda*}_0}^2 + C_{17}\Big) + 4C_{15}\sum_{i=\hat{t}}^{t-1} \left[ \prod_{k=i+1}^{t-1}(1-\beta_k\lambda_P) \right]\beta^2_i\tau^2_i\nonumber\\
&\quad + C_3\sum_{i=\hat{t}}^{t-1} \left[ \prod_{k=i+1}^{t-1}(1-\beta_k\lambda_P) \right]\alpha_i\nonumber\\
&\overset{(ii)}{\leq} e^{-\lambda_P \sum_{i=\hat{t}}^{t-1}\beta_i} \Big(C_{16}\ltwo{\theta_0-\theta^{\lambda*}_0}^2 + C_{17}\Big) + 4C_{15}\sum_{i=\hat{t}}^{t-1} e^{-\lambda_P \sum_{k=i+1}^{t-1}\beta_i} \beta^2_k\tau^2_i\nonumber\\
&\quad + C_3\sum_{i=\hat{t}}^{t-1} e^{-\lambda_P \sum_{k=i+1}^{t-1}\beta_i}\alpha_i\nonumber\\
&\overset{(iii)}{\leq} e^{-\frac{\lambda_P C_\beta }{1-\nu} [(t+1)^{1-\nu} - (\hat{t}+1)^{1-\nu}]} \Big(C_{16}\ltwo{\theta_0-\theta^{\lambda*}_0}^2 + C_{17}\Big) \nonumber\\
&\quad + 4C_{15}\Big(C_{18}\tau^2_t e^{-\frac{\lambda_P C_\beta }{2(1-\nu)} [(t+1)^{1-\nu} - (\hat{t}+1)^{1-\nu}]} + C_{19} \tau^2_{t} \beta_{t-1}\Big)\nonumber\\
&\quad + C_3\Big( C_{20}e^{-\frac{\lambda_P C_\beta }{2(1-\nu)} [(t+1)^{1-\nu} - (\hat{t}+1)^{1-\nu}]} + \frac{C_{21}}{t^{\sigma-\nu}} \Big)\label{part_sub4}
\end{flalign}
where $(i)$ follows from Lemma \ref{lemma: support3}, $(ii)$ follows from the fact that $1-\beta_i\lambda_\theta e^{-\lambda_\theta\beta_i}$, and $(iii)$ follows from Lemmas \ref{lemma: accumulate1} and \ref{lemma: accumulate2}. Taking expectation over $\mf_{t-\tau_t}$ on both sides of \cref{part_sub4} yields the following preliminary bound:
\begin{flalign}
\mE\big[\ltwo{\theta_t-\theta^{\lambda*}_t}^2\big] = \mathcal{O}\left(\frac{1}{(1-\gamma)t^{\sigma-\nu}}\right).\label{preliminary}
\end{flalign}

\textbf{Step 3.} \emph{Recursively refining tracking error bound.} The bound we obtain in \cref{preliminary} is not tight. Since the slow-drift error term diminishes as the tracking error diminishes. By recursively substituting the preliminary bound of $\mE[\ltwo{\theta_t-\theta^{\lambda*}_t}^2]$ into the slow-drift term, we can obtain a refined decay rate of the tracking error. We proceed as follows.

First, \cref{preliminary} implies that there exists a constant $D_1<\infty$ such that
\begin{flalign}
\mE\big[\ltwo{\theta_t-\theta^{\lambda*}_t}^2\big] \leq \frac{D_1}{(1-\gamma)(t+1)^{\sigma-\nu}}.\label{part2_subbound1}
\end{flalign}

By substituting \cref{part2_subbound1} into the third term on the right hand side of \cref{part1}, we can derive a tigher bound on this term as follows
\begin{flalign}
\mE\big[\langle \theta_t-\theta^{\lambda*}_t, \theta^{\lambda*}_t -\theta^{\lambda*}_{t+1} \rangle] &\leq \sqrt{\mE\big[\ltwo{\theta_t-\theta^{\lambda*}_t}^2] \mE\big[\ltwo{\theta^{\lambda*}_t -\theta^{\lambda*}_{t+1}}^2]} \nonumber\\
&\overset{(i)}{\leq} \sqrt{\frac{C^2_3 D_1}{(1-\gamma)(t+1)^{\sigma-\nu}}\alpha^2_t} = \frac{C_3\sqrt{ C_\alpha D_1}}{(1-\gamma)^{0.5}(t+1)^{1.5\sigma-0.5\nu}},\label{part2_subbound2}
\end{flalign}
where $(i)$ follows from \Cref{lemma: fixtracking}. Rewrite \cref{part1_sub3} as follows
\begin{flalign}
&\ltwo{\theta_{t+1}-\theta^{\lambda*}_{t+1}}^2\nonumber\\
&\leq \Big(1-2\beta_t\lambda_P + (C_\phi^2+\lambda)C_3\alpha_t\beta_t + C_1\beta_t^2\Big) \ltwo{\theta_t-\theta^{\lambda*}_t}^2 + \beta_t \zeta_t(\theta_t) + 2\langle \theta_t-\theta^{\lambda*}_t, \theta^{\lambda*}_t -\theta^{\lambda*}_{t+1} \rangle \nonumber\\
&\quad + 2\left[(C_\phi^2+\lambda)(R_\theta+\frac{1}{2}) + \frac{2r_{\max}C_\phi}{1-\gamma} \right]C_3\alpha_t\beta_t + C_3^2\alpha_t^2+C_2\beta_t^2. \label{part2_sub2}
\end{flalign}
Taking expectation on both sides of \cref{part2_sub2} conditioned on the filtration $\mf_{t-\tau_t}$ yields
\begin{flalign}
&\mE\Big[\ltwo{\theta_{t+1}-\theta^{\lambda*}_{t+1}}^2|\mf_{t-\tau_t}\Big]\nonumber\\
&\leq \Big(1-2\beta_t\lambda_P + (C_\phi^2+\lambda)C_3\alpha_t\beta_t + C_1\beta_t^2\Big) \mE\Big[\ltwo{\theta_t-\theta^{\lambda*}_t}^2|\mf_{t-\tau_t}\Big] + \beta_t \mE\Big[\zeta_t(\theta_t)|\mf_{t-\tau_t}\Big] \nonumber\\
&\quad + 2\mE\Big[\langle \theta_t-\theta^{\lambda*}_t, \theta^{\lambda*}_t -\theta^{\lambda*}_{t+1} \rangle|\mf_{t-\tau_t}\Big] + 2\left[(C_\phi^2+\lambda)(R_\theta+\frac{1}{2}) + \frac{2r_{\max}C_\phi}{1-\gamma} \right]C_3\alpha_t\beta_t \nonumber\\
&\quad + C_3^2\alpha_t^2+C_2\beta_t^2.\nonumber
\end{flalign}
Following the steps similar to those in \textbf{Step 1}, we obtain
\begin{flalign}
\mE\left[\ltwo{\theta_{t+1}-\theta^{\lambda*}_{t+1}}^2|\mf_{t-\tau_t}\right]&\leq \left(1-\beta_t\lambda_P \right) \mE\left[\ltwo{\theta_t-\theta^{\lambda*}_t}^2|\mf_{t-\tau_t}\right] + 4C_{15}\beta^2_t\tau^2_t \nonumber\\
&\quad + 2\mE\Big[\langle \theta_t-\theta^{\lambda*}_t, \theta^{\lambda*}_t -\theta^{\lambda*}_{t+1} \rangle|\mf_{t-\tau_t}\Big].\label{part2_sub3}
\end{flalign}
Taking expectation on both side of \cref{part2_sub3} yields
\begin{flalign}
\mE\Big[\ltwo{\theta_{t+1}-\theta^{\lambda*}_{t+1}}^2\Big]&\leq \left(1-\beta_t\lambda_P \right) \mE\left[\ltwo{\theta_t-\theta^{\lambda*}_t}^2 \right] + 4C_{15}\beta^2_t\tau^2_t + 2\mE\Big[\langle \theta_t-\theta^{\lambda*}_t, \theta^{\lambda*}_t -\theta^{\lambda*}_{t+1} \rangle \Big]\nonumber\\
&\overset{(i)}{\leq} \left(1-\beta_t\lambda_P \right) \mE\left[\ltwo{\theta_t-\theta^{\lambda*}_t}^2 \right] + 4C_{15}\beta^2_t\tau^2_t + \frac{2C_3\sqrt{ C_\alpha D_1}}{(1-\gamma)^{0.5}(t+1)^{1.5\sigma-0.5\nu}},\label{part2_sub4}
\end{flalign}
where $(i)$ follows from \cref{part2_subbound2}. Applying \cref{part2_sub4} recursively yields
\begin{flalign}
\mE\big[\ltwo{\theta_t-\theta^{\lambda*}_t}^2\big] &\leq \left[ \prod_{i=\hat{t}}^{t-1}(1-\beta_i\lambda_P) \right]\mE\big[\ltwo{\theta_{\hat{t}}-\theta^{\lambda*}_{\hat{t}}}^2 \big]+ 4C_{15}\sum_{i=\hat{t}}^{t-1} \left[ \prod_{k=i+1}^{t-1}(1-\beta_k\lambda_\theta) \right]\beta^2_i\tau^2_i\nonumber\\
&\quad + \frac{2C_3}{(1-\gamma)^{0.5}}\sqrt{ C_\alpha D_1}\sum_{i=\hat{t}}^{t-1} \left[ \prod_{k=i+1}^{t-1}(1-\beta_k\lambda_P) \right]\frac{1}{(i+1)^{1.5\sigma-0.5\nu}}\nonumber\\
&\leq \left[ \prod_{i=\hat{t}}^{t-1}(1-\beta_i\lambda_P) \right] \Big(C_{16}\ltwo{\theta_0-\theta^{\lambda*}_0}^2 + C_{17}\Big) + 4C_{15}\sum_{i=\hat{t}}^{t-1} \left[ \prod_{k=i+1}^{t-1}(1-\beta_k\lambda_P) \right]\beta^2_i\tau^2_i\nonumber\\
&\quad + \frac{2C_3}{(1-\gamma)^{0.5}}\sqrt{ C_\alpha D_1}\sum_{i=\hat{t}}^{t-1} \left[ \prod_{k=i+1}^{t-1}(1-\beta_k\lambda_P) \right]\frac{1}{(i+1)^{1.5\sigma-0.5\nu}}  \nonumber\\
&\leq e^{-\lambda_P \sum_{i=\hat{t}}^{t-1}\beta_i} \Big(C_{16}\ltwo{\theta_0-\theta^{\lambda*}_0}^2 + C_{17}\Big) + 4C_{15}\sum_{i=\hat{t}}^{t-1} e^{-\lambda_P \sum_{k=i+1}^{t-1}\beta_k} \beta^2_i\tau^2_i\nonumber\\
&\quad + \frac{2C_3}{(1-\gamma)^{0.5}}\sqrt{ C_\alpha D_1}\sum_{i=\hat{t}}^{t-1} e^{-\lambda_P \sum_{k=i+1}^{t-1}\beta_k}\frac{1}{(i+1)^{1.5\sigma-0.5\nu}}\nonumber\\
&\leq e^{-\frac{\lambda_P C_\beta }{1-\nu} [(t+1)^{1-\nu} - (\hat{t}+1)^{1-\nu}]} \Big(C_{16}\ltwo{\theta_0-\theta^{\lambda*}_0}^2 + C_{17}\Big) \nonumber\\
&\quad + C_{15}\Big(C_{18}\tau^2_t e^{-\frac{\lambda_P C_\beta }{2(1-\nu)} [(t+1)^{1-\nu} - (\hat{t}+1)^{1-\nu}]} + C_{19} \tau^2_{t-1} \beta_{t-1}\Big)\nonumber\\
&\quad + \frac{2C_3}{(1-\gamma)^{0.5}}\sqrt{ C_\alpha D_1}\Big( C_{22}e^{-\frac{\lambda_P C_\beta }{2(1-\nu)} [(t+1)^{1-\nu} - (\hat{t}+1)^{1-\nu}]} + \frac{C_{23}}{t^{1.5\sigma-1.5\nu}} \Big), \label{part2_sub5}
\end{flalign}
where $0<C_{22}<\infty$ and $0<C_{23}<\infty$ are constants defined similarly to Lemmas \ref{lemma: accumulate1} and \ref{lemma: accumulate2}. If $1.5(\sigma-\nu)> \nu$, then we have
\begin{flalign}
\mE\left[\ltwo{\theta_t-\theta^{\lambda*}_t}^2\right] = \mathcal{O}\left(\frac{\log^2(t)}{(1-\gamma)t^{\nu}}\right),\nonumber
\end{flalign}
and otherwise
\begin{flalign}
\mE\left[\ltwo{\theta_t-\theta^{\lambda*}_t}^2\right] = \mathcal{O}(\frac{1}{(1-\gamma)^{1.5}t^{1.5(\sigma-\nu)}}),\nonumber
\end{flalign}
which yields a tighter bound than that in \textbf{Step 1}. Applying the steps similar to those from \cref{part2_subbound1} to \cref{part2_sub5} for finite times, we can eventually obtain the following bound
\begin{flalign*}
\mE\left[\ltwo{\theta_t-\theta^{\lambda*}_t}^2\right]=\left\{
\begin{array}{lr}
\mathcal{O}\big(\frac{\log^2 t}{(1-\gamma)t^\nu}\big), & \sigma> 1.5\nu, \\
\mathcal{O}\big(\frac{\log^2 t}{(1-\gamma)^2t^\nu}\big), & \sigma= 1.5\nu, \\
\mathcal{O}\big(\frac{1}{(1-\gamma)^2t^{2(\sigma-\nu)}}\big), &  \nu<\sigma<1.5\nu.
\end{array}
\right.
\end{flalign*}
which implies the result in Theorem \ref{critic_rate}.

\section{Proof of Theorem \ref{thm3}}\label{proof5}
We provide the proof in three major steps.

\textbf{Step 1.} \emph{Decomposing convergence error.} Define $P_{w_t}(s,a) = \phi_{w_{t}}(s,a) \phi_{w_{t}}(s,a)^\top$. Due to the $L_J$-Lipschitz condition of the objective function indicated by \Cref{lemma: sum}, we can obtain
\begin{flalign}
J(w_{t+1})&\geq J(w_t)+\langle \nabla_w J(w_t), w_{t+1}-w_t \rangle -\frac{L_J}{2}\ltwo{w_{t+1}-w_{t}}^2\nonumber\\
&= J(w_t)+\alpha_t \langle \nabla_w J(w_t), P_{w_t}(s_t,a_t) \theta_t \rangle -\frac{L_J \alpha^2_t}{2}\ltwo{P_{w_t}\theta_{t}}^2\nonumber\\
&= J(w_t)+\alpha_t \langle \nabla_w J(w_t), P_{w_t}(s_t,a_t)\theta^{\lambda*}_t \rangle + \alpha_t \langle \nabla_w J(w_t), P_{w_t}(s_t,a_t)(\theta_t-\theta^{\lambda*}_t) \rangle \nonumber\\
&\quad  -\frac{L_J \alpha^2_t}{2}\ltwo{P_{w_t}(s_t,a_t)\theta_{t}}^2\nonumber\\
&\geq J(w_t)+\alpha_t \ltwo{\nabla_w J(w_t)}^2 +\alpha_t \langle \nabla_w J(w_t), P_{w_t}(s_t,a_t)\theta^{\lambda*}_t - \nabla_w J(w_t) \rangle  \nonumber\\
&\quad + \alpha_t \langle \nabla_w J(w_t), P_{w_t}(s_t,a_t)(\theta_t-\theta^{\lambda*}_t) \rangle -\frac{L_J \alpha^2_t}{2}\ltwo{P_{w_t}(s_t,a_t)}^2\ltwo{\theta_{t}}^2 \nonumber\\
&\geq J(w_t)+\alpha_t \ltwo{\nabla_w J(w_t)}^2 +\alpha_t \langle \nabla_w J(w_t), P_{w_t}(s_t,a_t)\theta^{\lambda*}_t - \nabla_w J(w_t) \rangle - \frac{1}{4} \alpha_t \ltwo{\nabla_w J(w_t)}^2   \nonumber\\
&\quad - \alpha_t\ltwo{P_{w_t}(s_t,a_t)(\theta_t-\theta^{\lambda*}_t)}^2 - \frac{L_J \alpha^2_t}{2}C^4_{\phi} R^2_\theta\nonumber\\
&\geq J(w_t)+ \frac{3}{4}\alpha_t \ltwo{\nabla_w J(w_t)}^2 +\alpha_t \langle \nabla_w J(w_t), P_{w_t}(s_t,a_t)\theta^{\lambda*}_t - \nabla_w J(w_t) \rangle  \nonumber\\
&\quad - \alpha_t\ltwo{P_{w_t}(s_t,a_t)}^2\ltwo{\theta_t-\theta^{\lambda*}_t}^2  - \frac{L_J \alpha^2_t}{2}C^4_{\phi} R^2_\theta\nonumber\\
&\geq J(w_t)+ \frac{3}{4}\alpha_t \ltwo{\nabla_w J(w_t)}^2 +\alpha_t \langle \nabla_w J(w_t), P_{w_t}(s_t,a_t)\theta^{\lambda*}_t - \nabla_w J(w_t) \rangle \nonumber\\
&\quad - \alpha_tC^4_\phi\ltwo{\theta_t-\theta^{\lambda*}_t}^2 - \frac{L_J \alpha^2_t}{2}C^4_{\phi} R^2_\theta.\label{part6_sub1}
\end{flalign}
Rearranging \cref{part6_sub1} and taking expectation on both sides yield
\begin{flalign}
\frac{3}{4}\alpha_t \mE[\ltwo{\nabla_w J(w_t)}^2] &\leq \mE[J(w_{t+1})] - \mE[J(w_t)] - \alpha_t \mE[\langle \nabla_w J(w_t), P_{w_t}(s_t,a_t)\theta^{\lambda*}_t - \nabla_w J(w_t) \rangle] \nonumber\\
&\quad + \alpha_tC^4_\phi \mE[\ltwo{\theta_t-\theta^{\lambda*}_t}^2]  + \frac{L_J \alpha^2_t}{2}C^4_{\phi} R^2_\theta. \label{part6_sub2}
\end{flalign}
\Cref{part6_sub2} shows that the gradient $\mE[\ltwo{\nabla_w J(w_t)}^2]$ can be bounded by the difference between the objective function values, the bias error of actor's update, the tracking error, and the variance error (which is bounded by a constant).

\textbf{Step 2.} \emph{Bounding bias error of actor's update.} We bound the bias error
 ($\mE[\langle \nabla_w J(w_t), P_{w_t}\theta^{\lambda*}_t - \nabla_w J(w_t) \rangle]$) in \cref{part6_sub2}. Such a bias error is caused by the correlation between samples due to the dynamically changing Markovian sampling. We develop a new proof to bound such a bias error in a nonlinear SA update due to the nonlinear parameterization of the policy, which is different from the bias error of linear SA that we studied in \Cref{critic_rate} and we proceed as follows. 
 
Note that $\nabla_w J(w_t)=\mE_{\nu_{\pi_{w_t}}}[P_{w_t}(s,a)]\theta^*_t$, and we denote $\overline{P}_{w_t} = \mE_{\nu_{\pi_{w_t}}}[P_{w_t}(s,a)]$ and $P_{w_t} = P_{w_t}(s_t,a_t)$ for simplicity. Then, we can obtain the following
\begin{flalign}
&\langle \nabla_w J(w_t), P_{w_t}\theta^{\lambda*}_t - \nabla_w J(w_t) \rangle \nonumber\\
&=\langle \nabla_w J(w_{t-\tau_t}), P_{w_t}\theta^{\lambda*}_t - \nabla_w J(w_t) \rangle + \langle \nabla_w J(w_t) - \nabla_w J(w_{t-\tau_t}), P_{w_t}\theta^{\lambda*}_t - \nabla_w J(w_t) \rangle\nonumber\\
&= \langle \nabla_w J(w_{t-\tau_t}), (P_{w_{t-\tau_t}}- \overline{P}_{w_{t-\tau_t}}) \theta^{\lambda*}_{t-\tau_t} \rangle + \langle \nabla_w J(w_t)-\nabla_w J(w_{t-\tau_t}), P_{w_t}\theta^{\lambda*}_t - \nabla_w J(w_t) \rangle \nonumber\\
&\quad +  \langle \nabla_w J(w_{t-\tau_t}), (P_{w_{t}}- P_{w_{t-\tau_t}}) \theta^{\lambda*}_t \rangle + \langle \nabla_w J(w_{t-\tau_t}), P_{w_{t-\tau_t}} (\theta^{\lambda*}_t - \theta^{\lambda*}_{t-\tau_t}) \rangle \nonumber\\
&\quad  + \langle \nabla_w J(w_{t-\tau_t}), \overline{P}_{w_{t-\tau_t}} (\theta^{\lambda*}_t - \theta^{\lambda*}_{t-\tau_t}) \rangle + \langle \nabla_w J(w_{t-\tau_t}), \overline{P}_{w_{t}} (\theta^{\lambda*}_t - \theta^{*}_t) \rangle \nonumber\\
&\leq \langle \nabla_w J(w_{t-\tau_t}), (P_{w_{t-\tau_t}}- \overline{P}_{w_{t-\tau_t}}) \theta^{\lambda*}_{t-\tau_t} \rangle +  \ltwo{\nabla_w J(w_t)-\nabla_w J(w_{t-\tau_t})}  \ltwo{P_{w_t}\theta^{\lambda*}_t - \nabla_w J(w_t)}  \nonumber\\
&\quad +   \ltwo{\nabla_w J(w_{t-\tau_t})}  \ltwo{P_{w_{t}}- P_{w_{t-\tau_t}}}  \ltwo{\theta^{\lambda*}_t}  +  \ltwo{\nabla_w J(w_{t-\tau_t})} \ltwo{P_{w_{t-\tau_t}}} \ltwo{\theta^{\lambda*}_t - \theta^{\lambda*}_{t-\tau_t}}  \nonumber\\
&\quad  + \ltwo{ \nabla_w J(w_{t-\tau_t})} \ltwo{\overline{P}_{w_{t-\tau_t}}} \ltwo{\theta^{\lambda*}_t - \theta^{\lambda*}_{t-\tau_t} } +  \ltwo{\nabla_w J(w_{t-\tau_t})} \ltwo{\overline{P}_{w_{t}}} \ltwo{\theta^{\lambda*}_t - \theta^{*}_t } \nonumber\\
&\overset{(i)}{\leq} \langle \nabla_w J(w_{t-\tau_t}), (P_{w_{t-\tau_t}}- \overline{P}_{w_{t-\tau_t}}) \theta^{\lambda*}_{t-\tau_t} \rangle +  L_J  \left(C^2_\phi R_\theta + \frac{C_\phi r_{\max}}{1-\gamma}\right)\ltwo{w_t-w_{t-\tau_t}}  \nonumber\\
&\quad +   \frac{2C^2_\phi L_\phi R_\theta r_{\max}}{1-\gamma}   \ltwo{w_t-w_{t-\tau_t}} +  \frac{2C^3_\phi L_\theta r_{\max}}{1-\gamma} \ltwo{w_t-w_{t-\tau_t}}  +  \frac{C^3_\phi C_\lambda r_{\max}}{1-\gamma} \lambda \nonumber\\
&\overset{(ii)}{\leq}\langle \nabla_w J(w_{t-\tau_t}), (P_{w_{t-\tau_t}}- \overline{P}_{w_{t-\tau_t}}) \theta^{\lambda*}_{t-\tau_t} \rangle + C_{27}\alpha_{t-\tau_t}\tau_t + \frac{C^3_\phi C_\lambda r_{\max}}{1-\gamma} \lambda, \label{part6_sub3}
\end{flalign}
where $(i)$ follows from Lemmas \ref{lemma: fixtracking}, \ref{bias_supp1} and \ref{lemma: fixtracking}, and in $(ii)$ we define
\begin{flalign*}
C_{27}=\left[ L_J  \left(C^2_\phi R_\theta + \frac{C_\phi r_{\max}}{1-\gamma}\right) + \frac{2C^2_\phi L_\phi R_\theta r_{\max}}{1-\gamma} + \frac{2C^3_\phi L_\theta r_{\max}}{1-\gamma}  \right] R_\theta.
\end{flalign*}
We next consider $\mE[\langle \nabla_w J(w_{t-\tau_t}), (P_{w_{t-\tau_t}}- \overline{P}_{w_{t-\tau_t}}) \theta^{\lambda*}_t \rangle|\mf_{t-\tau_t}]$, and have
\begin{flalign}
\mE[\langle \nabla_w& J(w_{t-\tau_t}), (P_{w_{t-\tau_t}}- \overline{P}_{w_{t-\tau_t}}) \theta^{\lambda*}_{t-\tau_t} \rangle|\mf_{t-\tau_t}]\nonumber\\
&=\langle \nabla_w J(w_{t-\tau_t}), (\mE[P_{w_{t-\tau_t}}|\mf_{t-\tau_t}]- \overline{P}_{w_{t-\tau_t}}) \theta^{\lambda*}_{t-\tau_t} \rangle\nonumber\\
&\leq \ltwo{\nabla_w J(w_{t-\tau_t})}  \ltwo{\mE[P_{w_{t-\tau_t}}|\mf_{t-\tau_t}]- \overline{P}_{w_{t-\tau_t}}} \ltwo{\theta^{\lambda*}_{t-\tau_t}}\nonumber\\
&\leq \frac{C_\phi R_\theta r_{\max}}{1-\gamma}\Delta_{P,\tau_t} \overset{(i)}{\leq} \frac{C_6C_\phi R_\theta r_{\max}}{1-\gamma} \alpha_{t-\tau_t}\tau^2_t,\label{part6_sub4}
\end{flalign}
where $(i)$ follows from Lemma \ref{lemma: delta}. Take expectation over $\mf_{t-\tau_t}$ on both sides of \cref{part6_sub4}, and recall that $\alpha_{t-\tau_t}\leq 2\alpha_t$ for all $t\geq \hat{t}$. Then we have the following bound
\begin{flalign}
\mE[\langle \nabla_w J(w_t), P_{w_t}\theta^{\lambda*}_t - \nabla_w J(w_t) \rangle]\leq C_{28}\alpha_{t}\tau^2_t + \frac{C^3_\phi C_\lambda r_{\max}}{1-\gamma} \lambda,\label{part6_sub5}
\end{flalign}
where $C_{28}=\frac{2C_6C_\phi R_\theta r_{\max}}{1-\gamma}+C_{27}$. Substituting \cref{part6_sub5} into \cref{part6_sub2} yields
\begin{flalign}
\frac{3}{4}\alpha_t \mE[\ltwo{\nabla_w J(w_t)}^2] &\leq \mE[J(w_{t+1})] - \mE[J(w_t)] + C_{28}\alpha^2_{t}\tau^2_t + \frac{C^3_\phi C_\lambda r_{\max}}{1-\gamma} \lambda\alpha_{t} \nonumber\\
&\quad + \alpha_tC^4_\phi \mE[\ltwo{\theta_t-\theta^{\lambda*}_t}^2]  + \frac{L_J \alpha^2_t}{2}C^4_{\phi} R^2_\theta.\nonumber
\end{flalign}
Telescoping the above inequality from $t$ to $\hat{t}$ yields
\begin{flalign}
\frac{3}{4} \sum_{i=\hat{t}}^{t}\alpha_i \mE[\ltwo{\nabla_w J(w_i)}^2] &\leq \mE[J(w_{t+1})] - \mE[J(w_{\hat{t}})] + C_{28}\sum_{i=\hat{t}}^{t}\alpha^2_{i}\tau^2_i + \frac{C^3_\phi C_\lambda r_{\max}}{1-\gamma} \lambda\sum_{i=\hat{t}}^{t}\alpha_{i} \nonumber\\
&\quad + C^4_\phi \sum_{i=\hat{t}}^{t}\alpha_i \mE[\ltwo{\theta_i-\theta^{\lambda*}_i}^2]  + \frac{L_J }{2}C^4_{\phi} R^2_\theta \sum_{i=\hat{t}}^{t}\alpha^2_i,\nonumber
\end{flalign}
which implies
\begin{flalign}
\frac{3}{4} \sum_{i=0}^{t}\alpha_i \mE[\ltwo{\nabla_w J(w_i)}^2] &\leq \frac{r_{\max}}{1-\gamma} + \frac{3C_\phi r_{\max}}{4(1-\gamma)} \sum_{i=0}^{\hat{t}-1}\alpha_i  + (C_{28} + \frac{L_J }{2}C^4_{\phi} R^2_\theta )\sum_{i=0}^{t}\alpha^2_{i}\tau^2_i \nonumber\\
&\quad + C^4_\phi \sum_{i=0}^{t}\alpha_i \mE[\ltwo{\theta_i-\theta^{\lambda*}_i}^2]  + \frac{C^3_\phi C_\lambda r_{\max}}{1-\gamma} \lambda\sum_{i=0}^{t}\alpha_{i}.\label{part3_sub3}
\end{flalign}

\textbf{Step 3.} \emph{Analyzing convergence rate under various stepsizes.} We then analyze the error bounds on the convergence rate under various stepsize settings for fast and slow time scales. 

The result of critic's convergence rate in Theorem \ref{critic_rate} implies that $\mE\big[\ltwo{\theta_i-\theta^{\lambda*}_i}^2\big]$ achieves the fastest convergence rate with a fixed $\sigma$ when $1.5\nu=\sigma$. Thus, here we consider the case when $1.5\nu=\sigma$, and have $\mE\big[\ltwo{\theta_i-\theta^{\lambda*}_i}^2\big]\leq D_2\log^2(i)/[(1-\gamma)^2(1+i)^{\frac{2}{3}\sigma}]$ for a positive constant $D_2<\infty$. We denote the distribution $P_{J,t}(i)=\text{Prob}(x=i)=\frac{\alpha_i}{\sum_{i=0}^{t}\alpha_i}$. Dividing both sides of \cref{part3_sub3} by $\sum_{i=0}^{t}\alpha_i$ and letting $\tilde{t}\sim P_J(\tilde{t}=i)$, we obtain
\begin{flalign}
\frac{3}{4}\mE\ltwo{\nabla_w J(w_{\tilde{t}})}^2 &\leq \frac{r_{\max}}{1-\gamma} + \frac{3C_\phi r_{\max}}{4(1-\gamma)} \frac{\sum_{i=0}^{\hat{t}-1}\alpha_i}{\sum_{i=0}^{t}\alpha_i}  + (C_{28} + \frac{L_J }{2}C^4_{\phi} R^2_\theta ) \frac{\sum_{i=0}^{t}\alpha^2_{i}\tau^2_i}{\sum_{i=0}^{t}\alpha_i} \nonumber\\
&\quad + C^4_\phi \frac{\sum_{i=0}^{t}\alpha_i \mE[\ltwo{\theta_i-\theta^{\lambda*}_i}^2]}{\sum_{i=0}^{t}\alpha_i} + \frac{C^3_\phi C_\lambda r_{\max}}{1-\gamma} \lambda.\nonumber
\end{flalign}
\textbf{Case 1:} $\sigma>\frac{3}{5}$. We first have
\begin{flalign*}
\sum_{i=0}^{t} \alpha_i\mE\big[\ltwo{\theta_i-\theta^{\lambda*}_i}^2\big]\leq \frac{D_2}{(1-\gamma)^2}C_\alpha \Big[\int_{0}^{t}(1+x)^{-\frac{5}{3}\sigma}dx+1\Big]\log^2(t) \leq  \frac{5\sigma}{5\sigma-3}\frac{D_2}{(1-\gamma)^2}C_\alpha\log^2(t).
\end{flalign*}
Note that $\sum_{i=0}^{t} \alpha_i\geq \frac{C_\alpha}{1-\sigma}[(t+1)^{1-\sigma}-1]$ and $\sum_{i=0}^{t} \alpha^2_i\leq \frac{2C^2_\phi \sigma}{2\sigma-1}$. We obtain
\begin{flalign}
\frac{3}{4}\mE\ltwo{\nabla_w J(w_{\tilde{t}})}^2 &\leq \frac{r_{\max}[1/(1-\gamma)+3C_\phi\sum_{i=0}^{\hat{t}-1}\alpha_i/4]}{ \frac{C_\alpha}{1-\sigma}[(t+1)^{1-\sigma}-1] } + (C_{28} + \frac{L_J }{2}C^4_{\phi} R^2_\theta ) \frac{\frac{2C^2_\phi \sigma}{2\sigma-1}\tau^2_t}{\frac{C_\alpha}{1-\sigma}[(t+1)^{1-\sigma}-1]} \nonumber\\
&\quad + \frac{C^4_\phi}{(1-\gamma)^2} \frac{\frac{5\sigma}{5\sigma-3}D_2C_\alpha\log^2(t)}{\frac{C_\alpha}{1-\sigma}[(t+1)^{1-\sigma}-1]}  + \frac{C^3_\phi C_\lambda r_{\max}}{1-\gamma} \lambda.\nonumber
\end{flalign}
\textbf{Case 2:} $\sigma=\frac{3}{5}$. We have
\begin{flalign*}
\sum_{i=0}^{t} \alpha_i\mE\big[\ltwo{\theta_i-\theta^{\lambda*}_i}^2\big]\leq \frac{D_2}{(1-\gamma)^2}C_\alpha \Big[\int_{0}^{t}\frac{1}{1+x}dx+1\Big]\log^2(t) \leq  \frac{2D_2}{(1-\gamma)^2}C_\alpha\log^3(t).
\end{flalign*}
Thus,
\begin{flalign}
\frac{3}{4}\mE\ltwo{\nabla_w J(w_{\tilde{t}})}^2 &\leq \frac{r_{\max}[1/(1-\gamma)+3C_\phi\sum_{i=0}^{\hat{t}-1}\alpha_i/4]}{ \frac{C_\alpha}{1-\sigma}[(t+1)^{\frac{2}{5}}-1] } + (C_{28} + \frac{L_J }{2}C^4_{\phi} R^2_\theta ) \frac{\frac{2C^2_\phi \sigma}{2\sigma-1}\tau^2_t}{\frac{C_\alpha}{1-\sigma}[(t+1)^{\frac{2}{5}}-1]} \nonumber\\
&\quad + \frac{C^4_\phi}{(1-\gamma)^2} \frac{2D_2C_\alpha\log^3(t)}{\frac{C_\alpha}{1-\sigma}[(t+1)^{\frac{2}{5}}-1]} + \frac{C^3_\phi C_\lambda r_{\max}}{1-\gamma} \lambda.\nonumber
\end{flalign}
\textbf{Case 3:} $\frac{1}{2}<\sigma<\frac{3}{5}$. We have
\begin{flalign*}
\sum_{i=0}^{t} \alpha_i\mE\big[\ltwo{\theta_i-\theta^{\lambda*}_i}^2\big]&\leq \frac{D_2}{(1-\gamma)^2}C_\alpha \Big[\int_{0}^{t}(1+x)^{-\frac{5}{3}\sigma}dx+1\Big]\log^2(t) \nonumber\\
&\leq  \frac{3}{3-5\sigma}\frac{D_2}{(1-\gamma)^2}C_\alpha\log^2(t)(t+1)^{1-\frac{5}{3}\sigma}.
\end{flalign*}
Thus
\begin{flalign}
\frac{3}{4}\mE\ltwo{\nabla_w J(w_{\tilde{t}})}^2 &\leq \frac{r_{\max}[1/(1-\gamma)+3C_\phi\sum_{i=0}^{\hat{t}-1}\alpha_i/4]}{ \frac{C_\alpha}{1-\sigma}[(t+1)^{1-\sigma}-1] } + (C_{28} + \frac{L_J }{2}C^4_{\phi} R^2_\theta ) \frac{\frac{2C^2_\phi \sigma}{2\sigma-1}\tau^2_t}{\frac{C_\alpha}{1-\sigma}[(t+1)^{1-\sigma}-1]}  \nonumber\\
&\quad + \frac{C^4_\phi}{(1-\gamma)^2} \frac{\frac{3}{3-5\sigma}D_2C_\alpha\log^2(t)(t+1)^{1-\frac{5}{3}\sigma}}{\frac{C_\alpha}{1-\sigma}[(t+1)^{1-\sigma}-1]} + \frac{C^3_\phi C_\lambda r_{\max}}{1-\gamma} \lambda.\nonumber
\end{flalign}
\textbf{Case 4:} $\sigma=\frac{1}{2}$. We have $\sum_{i=0}^{t} \alpha^2_i\leq 2C^2_\phi \log(t+1)$. Thus,
\begin{flalign}
\frac{3}{4}\mE\ltwo{\nabla_w J(w_{\tilde{t}})}^2 &\leq \frac{r_{\max}[1/(1-\gamma)+3C_\phi\sum_{i=0}^{\hat{t}-1}\alpha_i/4]}{ \frac{C_\alpha}{1-\sigma}[(t+1)^{\frac{1}{2}}-1] } + (C_{28} + \frac{L_J }{2}C^4_{\phi} R^2_\theta ) \frac{2C^2_\phi \log(t+1)\tau^2_t}{\frac{C_\alpha}{1-\sigma}[(t+1)^{\frac{1}{2}}-1]}  \nonumber\\
&\quad + \frac{C^4_\phi}{(1-\gamma)^2} \frac{\frac{3}{3-5\sigma}D_2C_\alpha\log^2(t)(t+1)^{1-\frac{5}{3}\sigma}}{\frac{C_\alpha}{1-\sigma}[(t+1)^{\frac{1}{2}}-1]} + \frac{C^3_\phi C_\lambda r_{\max}}{1-\gamma} \lambda.\nonumber
\end{flalign}
\textbf{Case 5:} $\sigma<\frac{1}{2}$. We have $\sum_{i=0}^{t} \alpha^2_i\leq \frac{C^2_\phi}{1-2\sigma}(t+1)^{1-2\sigma}$. Thus,
\begin{flalign}
\frac{3}{4}\mE\ltwo{\nabla_w J(w_{\tilde{t}})}^2 &\leq \frac{r_{\max}[1/(1-\gamma)+3C_\phi\sum_{i=0}^{\hat{t}-1}\alpha_i/4]}{ \frac{C_\alpha}{1-\sigma}[(t+1)^{1-\sigma}-1] } + (C_{28} + \frac{L_J }{2}C^4_{\phi} R^2_\theta ) \frac{\frac{C^2_\phi}{1-2\sigma}(t+1)^{1-2\sigma}}{\frac{C_\alpha}{1-\sigma}[(t+1)^{1-\sigma}-1]}  \nonumber\\
&\quad + \frac{C^4_\phi}{(1-\gamma)^2} \frac{\frac{3}{3-5\sigma}D_2C_\alpha\log^2(t)(t+1)^{1-\frac{5}{3}\sigma}}{\frac{C_\alpha}{1-\sigma}[(t+1)^{1-\sigma}-1]} + \frac{C^3_\phi C_\lambda r_{\max}}{1-\gamma} \lambda.\nonumber
\end{flalign}
To summarize all the above cases, we have
\begin{flalign*}
\mE\ltwo{\nabla_w J(w_{\tilde{t}})}^2=\mathcal{O}(\lambda)+\left\{
\begin{array}{lr}
\mathcal{O}\big(\frac{\log^2 t}{(1-\gamma)^2t^{1-\sigma}}\big), & \sigma> \frac{3}{5}, \\
\mathcal{O}\big(\frac{\log^3 t}{(1-\gamma)^2t^{\frac{2}{5}}}\big), &  \sigma= \frac{3}{5}, \\
\mathcal{O}\big(\frac{\log^2 t}{(1-\gamma)^2t^{\frac{2}{3}\sigma}}\big), &  \sigma<\frac{3}{5}, \\
\end{array}
\right.
\end{flalign*}
The above bound implies that the optimal convergence rate can be obtained when $\sigma=\frac{3}{5}$. Then it requires at least $\mathcal{O}(\frac{1}{(1-\gamma)^2\epsilon^{2.5}}\log^3\frac{1}{\epsilon})$ iterations to obtain $\mE\ltwo{\nabla_w J(w_{\tilde{t}})}^2\leq \epsilon + \mathcal{O}(\lambda)$.

\section{Proof of Theorem \ref{thm4}}\label{proof1}
We provide the proof of \Cref{thm4} in two major steps.

\textbf{Step 1.} \emph{Decomposing convergence error.} 
 Denote $D(w) = D_{KL} \big(\pi^*(\cdot|s), \pi_w(\cdot|s  )\big) = \mE_{\nu_{\pi^*}}\Big[ \log\frac{\pi^*(a|s)}{\pi_w(a|s)} \Big]$. We derive as follows.
\begin{flalign}
&D(w_t)-D(w_{t+1}) \nonumber\\
&=\mE_{\nu_{\pi^*}}\Big[ \log(\pi_{w_{t+1}}(a|s)) - \log(\pi_{w_t}(a|s)) \Big]\nonumber\\
&\overset{(i)}{\geq} \mE_{\nu_{\pi^*}}\Big[ \nabla_w \log(\pi_{w_t}(a|s)) \Big]^\top (w_{t+1}-w_t) - \frac{L_\phi}{2}\ltwo{w_{t+1}-w_t}^2 \nonumber\\
&= \mE_{\nu_{\pi^*}}\Big[ \phi_{w_t}(s,a) \Big]^\top (w_{t+1}-w_t) - \frac{L_\phi}{2}\ltwo{w_{t+1}-w_t}^2 \nonumber\\
&= \alpha_t\mE_{\nu_{\pi^*}}\Big[ \phi_{w_t}(s,a) \Big]^\top \theta_t - \frac{L_\phi}{2} \alpha^2_t \ltwo{\theta_t}^2 \nonumber\\
&= \alpha_t\mE_{\nu_{\pi^*}}\Big[ \phi_{w_t}(s,a) \Big]^\top \theta^*_t +  \alpha_t\mE_{\nu_{\pi^*}}\Big[ \phi_{w_t}(s,a) \Big]^\top (\theta_t - \theta^{\lambda*}_t) +  \alpha_t\mE_{\nu_{\pi^*}}\Big[ \phi_{w_t}(s,a) \Big]^\top (\theta^{\lambda*}_t - \theta^*_t) \nonumber\\
&\quad - \frac{L_\phi}{2} \alpha^2_t \ltwo{\theta_t}^2 \nonumber\\
&= \alpha_t\mE_{\nu_{\pi^*}}\Big[ A_{\pi_{w_t}}(s,a) \Big]+  \alpha_t\mE_{\nu_{\pi^*}}\Big[ \phi_{w_t}(s,a) \Big]^\top (\theta_t - \theta^{\lambda*}_t) +  \alpha_t\mE_{\nu_{\pi^*}}\Big[ \phi_{w_t}(s,a) \Big]^\top (\theta^{\lambda*}_t - \theta^*_t) \nonumber\\
&\quad + \alpha_t\mE_{\nu_{\pi^*}}\Big[ \phi_{w_t}(s,a)^\top \theta^*_t -  A_{w_t}(s,a) \Big] - \frac{L_\phi}{2} \alpha^2_t \ltwo{\theta_t}^2 \nonumber\\
&= (1-\gamma)\alpha_t\Big(J(\pi^*)-J(\pi_{w_t})\Big)+  \alpha_t\mE_{\nu_{\pi^*}}\Big[ \phi_{w_t}(s,a) \Big]^\top (\theta_t - \theta^{\lambda*}_t) +  \alpha_t\mE_{\nu_{\pi^*}}\Big[ \phi_{w_t}(s,a) \Big]^\top (\theta^{\lambda*}_t - \theta^*_t) \nonumber\\
&\quad + \alpha_t\mE_{\nu_{\pi^*}}\Big[ \phi_{w_t}(s,a)^\top \theta^*_t -  A_{\pi_{w_t}}(s,a) \Big] - \frac{L_\phi}{2} \alpha^2_t \ltwo{\theta_t}^2 \nonumber\\
&\geq (1-\gamma)\alpha_t\Big(J(\pi^*)-J(\pi_{w_t})\Big)+  \alpha_t\mE_{\nu_{\pi^*}}\Big[ \phi_{w_t}(s,a) \Big]^\top (\theta_t - \theta^{\lambda*}_t) +  \alpha_t\mE_{\nu_{\pi^*}}\Big[ \phi_{w_t}(s,a) \Big]^\top (\theta^{\lambda*}_t - \theta^*_t) \nonumber\\
&\quad - \alpha_t\sqrt{\mE_{\nu_{\pi^*}}\big[ \phi_{w_t}(s,a)^\top \theta^*_t -  A_{\pi_{w_t}}(s,a) \big]^2} - \frac{L_\phi}{2} \alpha^2_t \ltwo{\theta_t}^2 \nonumber\\
&\geq (1-\gamma)\alpha_t\Big(J(\pi^*)-J(\pi_{w_t})\Big)+  \alpha_t\mE_{\nu_{\pi^*}}\Big[ \phi_{w_t}(s,a) \Big]^\top (\theta_t - \theta^{\lambda*}_t) +  \alpha_t\mE_{\nu_{\pi^*}}\Big[ \phi_{w_t}(s,a) \Big]^\top (\theta^{\lambda*}_t - \theta^*_t) \nonumber\\
&\quad - \sqrt{\linf{\frac{\nu_{\pi^*}}{\nu_{\pi_{w_t}}}}}\alpha_t\sqrt{\mE_{\nu_{\pi_{w_t}}}\big[ \phi_{w_t}(s,a)^\top \theta^*_t -  A_{\pi_{w_t}}(s,a) \big]^2} - \frac{L_\phi}{2} \alpha^2_t \ltwo{\theta_t}^2 \nonumber\\
&\overset{(ii)}{\geq} (1-\gamma)\alpha_t\Big(J(\pi^*)-J(\pi_{w_t})\Big)+  \alpha_t\mE_{\nu_{\pi^*}}\Big[ \phi_{w_t}(s,a) \Big]^\top (\theta_t - \theta^{\lambda*}_t) +  \alpha_t\mE_{\nu_{\pi^*}}\Big[ \phi_{w_t}(s,a) \Big]^\top (\theta^{\lambda*}_t - \theta^*_t) \nonumber\\
&\quad - \sqrt{\frac{1}{1-\gamma}  \linf{\frac{\nu_{\pi^*}}{\nu_{\pi_{w_0}}}}}\alpha_t\sqrt{\mE_{\nu_{\pi_{w_t}}}\big[ \phi_{w_t}(s,a)^\top \theta^*_t -  A_{\pi_{w_t}}(s,a) \big]^2} - \frac{L_\phi}{2} \alpha^2_t \ltwo{\theta_t}^2.\nonumber\\
&\overset{(iii)}{\geq} (1-\gamma)\alpha_t\Big(J(\pi^*)-J(\pi_{w_t})\Big) - \alpha_t C_\phi \ltwo{\theta_t - \theta^{\lambda*}_t} -  \alpha_t C_\phi C_{\lambda} \lambda \nonumber\\
&\quad - \sqrt{\frac{1}{1-\gamma}  \linf{\frac{\nu_{\pi^*}}{\nu_{\pi_{w_0}}}}}\alpha_t\sqrt{\mE_{\nu_{\pi_{w_t}}}\big[ \phi_{w_t}(s,a)^\top \theta^*_t -  A_{\pi_{w_t}}(s,a) \big]^2} - \frac{L_\phi}{2} \alpha^2_t \ltwo{\theta_t}^2.\label{recatch1}
\end{flalign}

where $(i)$ follows from the $L_\phi$ gradient Lipschitz condition given by Lemma \ref{lemma: liptz2}, $(ii)$ follows from the fact that $\nu_{\pi_{w_t}}\leq (1-\gamma)\nu_{\pi_{w_0}}$ \cite{agarwal2019optimality}, and $(iii)$ follows from Lemma \ref{lemma: regularization}. \Cref{recatch1} yields
\begin{flalign}
D(w_t)-D(w_{t+1})&\geq (1-\gamma)\alpha_t\Big(J(\pi^*)-J(\pi_{w_t})\Big) - \alpha_t C_\phi \ltwo{\theta_t - \theta^{\lambda*}_t} -  \alpha_t C_\phi C_{\lambda} \lambda \nonumber\\
&\quad - \alpha_t \sqrt{\frac{1}{1-\gamma}  \linf{\frac{\nu_{\pi^*}}{\nu_{\pi_{w_0}}}}} \sqrt{\zeta_{approx}} - \frac{L_\phi}{2} \alpha^2_t R_\theta^2. \label{part3_sub10}
\end{flalign}
Taking expectation on both side of \cref{part3_sub10} and rearranging the terms yield
\begin{flalign}
(1-\gamma)\alpha_t\Big(J(\pi^*)-\mE\big[J(\pi_{w_t})\big]\Big) &\leq \mE\big[D(w_t)\big]-\mE\big[D(w_{t+1})\big] + \alpha_t C_\phi \ltwo{\theta_t - \theta^{\lambda*}_t} + \alpha_t C_\phi C_{\lambda} \lambda \nonumber\\
&\quad + \alpha_t \sqrt{\frac{1}{1-\gamma}  \linf{\frac{\nu_{\pi^*}}{\nu_{\pi_{w_0}}}}} \sqrt{\zeta_{approx}} + \frac{L_\phi}{2} \alpha^2_t R_\theta^2. \label{part3_sub11}
\end{flalign}
\cref{part3_sub11} shows that the incremental change of the objective function values can be bounded by the changes of the KL-distance between the iterating policy and globally optimal policy, tracking error (which has been bounded in \Cref{critic_rate}), non-vanishing approximation error, and the variance error (which is upper bounded by a constant).

\textbf{Step 2.} \emph{Analyzing convergence rate under various stepsizes.} Summing \cref{part3_sub11} over iterations up to $t$-th step yields
\begin{flalign}
(1-\gamma)\sum_{i=0}^{t}&\alpha_i\Big(J(\pi^*)-\mE\big[J(\pi_{w_i})\big]\Big)\nonumber\\
&\leq \sum_{i=0}^{t}\left(\mE\big[D(w_i)\big]-\mE\big[D(w_{i+1})\big]\right) + \alpha_t C_\phi \mE\big[\ltwo{\theta_t - \theta^{\lambda*}_t}\big]  \nonumber\\
&\quad + C_\phi C_{\lambda} \lambda \sum_{i=0}^{t}\alpha_i + \sqrt{\frac{1}{1-\gamma}  \linf{\frac{\nu_{\pi^*}}{\nu_{\pi_{w_0}}}}} \sqrt{\zeta_{approx}} \sum_{i=0}^{t}\alpha_i \nonumber\\
&\quad + \frac{L_\phi R_\theta^2}{2}  \sum_{i=0}^{t}\alpha^2_i \nonumber\\
&\leq D(w_0) + \alpha_t C_\phi \sqrt{\mE\big[\ltwo{\theta_t - \theta^{\lambda*}_t}^2\big] }+ C_\phi C_{\lambda} \lambda \sum_{i=0}^{t}\alpha_i \nonumber\\
&\quad + \sqrt{\frac{1}{1-\gamma}  \linf{\frac{\nu_{\pi^*}}{\nu_{\pi_{w_0}}}}} \sqrt{\zeta_{approx}} \sum_{i=0}^{t}\alpha_i + \frac{L_\phi R_\theta^2}{2}  \sum_{i=0}^{t}\alpha^2_i.
\label{part3_sub12}
\end{flalign}
Dividing both sides of \cref{part3_sub12} by $\sum_{i=0}^{t}\alpha_i$ and letting $\tilde{t}\sim P_J(\tilde{t}=i)$, we obtain
\begin{flalign}
&(1-\gamma)\Big(J(\pi^*)-\mE\big[J(\pi_{w_{\tilde{t}}})\big]\Big) \nonumber\\
&\leq \frac{D_0}{\sum_{i=0}^{t}\alpha_i} + \frac{\sum_{i=0}^{t}\alpha_i \sqrt{\mE\big[\ltwo{\theta_i - \theta^{\lambda*}_i}^2\big] }}{\sum_{i=0}^{t}\alpha_i} + C_\phi C_{\lambda}\lambda  + \sqrt{\frac{1}{1-\gamma} \linf{\frac{\nu_{\pi^*}}{\nu_{\pi_{w_0}}}}} \sqrt{\zeta_{approx}} \nonumber\\
&\quad + \frac{L_\phi R_\theta^2}{2} \frac{\sum_{i=0}^{t}\alpha^2_i}{\sum_{i=0}^{t}\alpha_i}. \label{part3_sub13}
\end{flalign}
Similarly, we consider the case when $1.5\nu=\sigma$. We next analyze the error bounds on the convergence rate under various settings for $\sigma$.

\textbf{Case 1:} $\sigma>\frac{3}{4}$. We have
\begin{flalign}
&\sum_{i=0}^{t}\alpha_i \sqrt{\mE\big[\ltwo{\theta_i - \theta^{\lambda*}_i}^2\big] } \nonumber\\
&\leq C_\alpha\frac{\sqrt{D_2}}{1-\gamma} \sum_{i=0}^{t} \frac{\log(i)}{(1+i)^{\frac{4}{3}\sigma}}\leq C_\alpha\frac{\sqrt{D_2}}{1-\gamma}\Big[\int_{0}^{t}(1+x)^{-\frac{4}{3}\sigma}dx+1\Big] \log(t) \leq \frac{4\sigma}{4\sigma-3}C_\alpha\frac{\sqrt{D_2}}{1-\gamma}\log(t).\nonumber
\end{flalign}
Thus
\begin{flalign}
(1-\gamma)\Big(J(\pi^*)-\mE\big[J(\pi_{w_{\tilde{t}}})\big]\Big) &\leq \frac{D_0}{\frac{C_\alpha}{1-\sigma}[(t+1)^{1-\sigma}-1]} + \frac{\frac{4\sigma}{4\sigma-3}C_\alpha\sqrt{D_2}\log(t)}{\frac{C_\alpha}{1-\sigma}[(t+1)^{1-\sigma}-1](1-\gamma)} + C_\phi C_{\lambda}\lambda  \nonumber\\
&\quad + \sqrt{\frac{1}{1-\gamma}  \linf{\frac{\nu_{\pi^*}}{\nu_{\pi_{w_0}}}}} \sqrt{\zeta_{approx}} + \frac{L_\phi R_\theta^2}{2} \frac{\frac{2C^2_\phi \sigma}{2\sigma-1}}{\frac{C_\alpha}{1-\sigma}[(t+1)^{1-\sigma}-1]}. \label{part3_sub14}
\end{flalign}
\textbf{Case 2:} $\sigma=\frac{3}{4}$. We have
\begin{flalign}
&\sum_{i=0}^{t}\alpha_i \sqrt{\mE\big[\ltwo{\theta_i - \theta^{\lambda*}_i}^2\big] }\nonumber\\
& \leq C_\alpha\frac{\sqrt{D_2}}{1-\gamma} \sum_{i=0}^{t} \frac{\log(i)}{1+i}\leq C_\alpha\frac{\sqrt{D_2}}{1-\gamma}\Big[\int_{0}^{t}\frac{1}{1+x}dx+1\Big] \log(t) \leq \frac{8\sigma}{4\sigma-3}\frac{C_\alpha}{1-\gamma}\sqrt{D_2}\log^2(t).\nonumber
\end{flalign}
Thus,
\begin{flalign}
(1-\gamma)\Big(J(\pi^*)-\mE\big[J(\pi_{w_{\tilde{t}}})\big]\Big) &\leq \frac{D_0}{\frac{C_\alpha}{1-\sigma}[(t+1)^{\frac{1}{4}}-1]} + \frac{\frac{8\sigma}{4\sigma-3}C_\alpha\sqrt{D_2}\log^2(t)}{\frac{C_\alpha}{1-\sigma}[(t+1)^{\frac{1}{4}}-1](1-\gamma)} + C_\phi C_{\lambda}\lambda  \nonumber\\
&\quad + \sqrt{\frac{1}{1-\gamma}  \linf{\frac{\nu_{\pi^*}}{\nu_{\pi_{w_0}}}}} \sqrt{\zeta_{approx}} + \frac{L_\phi R_\theta^2}{2} \frac{\frac{2C^2_\phi \sigma}{2\sigma-1}}{\frac{C_\alpha}{1-\sigma}[(t+1)^{\frac{1}{4}}-1]}. \label{part3_sub15}
\end{flalign}
{Case 3:} $\frac{1}{2}<\sigma<\frac{3}{4}$. We have
\begin{flalign}
\sum_{i=0}^{t}\alpha_i \sqrt{\mE\big[\ltwo{\theta_i - \theta^{\lambda*}_i}^2\big] } &\leq C_\alpha\frac{\sqrt{D_2}}{1-\gamma} \sum_{i=0}^{t} \frac{\log(i)}{(1+i)^{\frac{4}{3}\sigma}}\leq C_\alpha\frac{\sqrt{D_2}}{1-\gamma}\Big[\int_{0}^{t}(1+x)^{-\frac{4}{3}\sigma}dx+1\Big] \log(t) \nonumber\\
&\leq \frac{3}{3-4\sigma}C_\alpha\frac{\sqrt{D_2}}{1-\gamma}\log(t)(t+1)^{1-\frac{4}{3}\sigma}.\nonumber
\end{flalign}
Thus,
\begin{flalign}
(1-\gamma)\Big(J(\pi^*)-\mE\big[J(\pi_{w_{\tilde{t}}})\big]\Big) &\leq \frac{D_0}{\frac{C_\alpha}{1-\sigma}[(t+1)^{1-\sigma}-1]} + \frac{\frac{3}{3-4\sigma}C_\alpha\sqrt{D_2}\log(t)(t+1)^{1-\frac{4}{3}\sigma}}{\frac{C_\alpha}{1-\sigma}[(t+1)^{1-\sigma}-1](1-\gamma)} + C_\phi C_{\lambda}\lambda  \nonumber\\
&\quad + \sqrt{\frac{1}{1-\gamma}  \linf{\frac{\nu_{\pi^*}}{\nu_{\pi_{w_0}}}}} \sqrt{\zeta_{approx}} + \frac{L_\phi R_\theta^2}{2} \frac{\frac{2C^2_\phi \sigma}{2\sigma-1}}{\frac{C_\alpha}{1-\sigma}[(t+1)^{1-\sigma}-1]}. \label{part3_sub16}
\end{flalign}
\textbf{Case 4:} $\sigma=\frac{1}{2}$. We have $\sum_{i=0}^{t} \alpha^2_i\leq 2C^2_\phi \log(t+1)$ and
\begin{flalign}
\sum_{i=0}^{t}\alpha_i \sqrt{\mE\big[\ltwo{\theta_i - \theta^{\lambda*}_i}^2\big] } \leq \frac{3}{3-4\sigma}C_\alpha\frac{\sqrt{D_2}}{1-\gamma}\log(t)(t+1)^{\frac{1}{3}}.\nonumber
\end{flalign}
Thus,
\begin{flalign}
(1-\gamma)\Big(J(\pi^*)-\mE\big[J(\pi_{w_{\tilde{t}}})\big]\Big) &\leq \frac{D_0}{\frac{C_\alpha}{1-\sigma}[(t+1)^{1-\sigma}-1]} + \frac{\frac{3}{3-4\sigma}C_\alpha\sqrt{D_2}\log(t)(t+1)^{\frac{1}{3}}}{\frac{C_\alpha}{1-\sigma}[(t+1)^{\frac{1}{2}}-1](1-\gamma)} + C_\phi C_{\lambda}\lambda  \nonumber\\
&\quad + \sqrt{\frac{1}{1-\gamma}  \linf{\frac{\nu_{\pi^*}}{\nu_{\pi_{w_0}}}}} \sqrt{\zeta_{approx}} + \frac{L_\phi R_\theta^2}{2} \frac{2C^2_\phi \log(t+1)}{\frac{C_\alpha}{1-\sigma}[(t+1)^{\frac{1}{2}}-1]}. \label{part3_sub17}
\end{flalign}
\textbf{Case 5:} $\sigma<\frac{1}{2}$. We have $\sum_{i=0}^{t} \alpha^2_i\leq \frac{C^2_\phi}{1-2\sigma}(t+1)^{1-2\sigma}$. Thus,
\begin{flalign}
(1-\gamma)\Big(J(\pi^*)-\mE\big[J(\pi_{w_{\tilde{t}}})\big]\Big) &\leq \frac{D_0}{\frac{C_\alpha}{1-\sigma}[(t+1)^{1-\sigma}-1]} + \frac{\frac{3}{3-4\sigma}C_\alpha\sqrt{D_2}\log(t)(t+1)^{1-\frac{4}{3}\sigma}}{\frac{C_\alpha}{1-\sigma}[(t+1)^{1-\sigma}-1](1-\gamma)} + C_\phi C_{\lambda}\lambda  \nonumber\\
&\quad + \sqrt{\frac{1}{1-\gamma}  \linf{\frac{\nu_{\pi^*}}{\nu_{\pi_{w_0}}}}} \sqrt{\zeta_{approx}} + \frac{L_\phi R_\theta^2}{2} \frac{\frac{C^2_\phi}{1-2\sigma}(t+1)^{1-2\sigma}}{\frac{C_\alpha}{1-\sigma}[(t+1)^{1-\sigma}-1]}. \label{part3_sub18}
\end{flalign}
Summarizing \cref{part3_sub14}, \cref{part3_sub15}, \cref{part3_sub16}, \cref{part3_sub17} and \cref{part3_sub18} yields
\begin{flalign*}
J(\pi^*)-\mE\big[J(\pi_{w_{\tilde{t}}})\big]\leq \sqrt{\frac{1}{(1-\gamma)^3}  \linf{\frac{\nu_{\pi^*}}{\nu_{\pi_{w_0}}}}} \sqrt{\zeta_{approx}} + \mathcal{O}(\lambda) + \left\{
\begin{array}{lr}
\mathcal{O}\big(\frac{\log t}{(1-\gamma)^2t^{1-\sigma}}\big), & \sigma\geq \frac{3}{4}, \\
\mathcal{O}\big(\frac{\log^2 t}{(1-\gamma)^2t^{\frac{1}{4}}}\big), &  \sigma= \frac{3}{4}, \\
\mathcal{O}\big(\frac{\log t}{(1-\gamma)^2t^{\frac{1}{3}\sigma}}\big), &  \sigma<\frac{3}{4}. \\
\end{array}
\right.
\end{flalign*}
The above bound implies that the optimal convergence rate can be obtained when $\sigma=\frac{3}{4}$. Then it requires at least $\mathcal{O}(\frac{1}{(1-\gamma)^8\epsilon^{4}}\log^2\frac{1}{\epsilon})$ iterations to obtain 
\begin{flalign*}
	J(\pi^*)-\mE\big[J(\pi_{w_{\tilde{t}}})\big]\leq \sqrt{\frac{1}{(1-\gamma)^3} \linf{\frac{\nu_{\pi^*}}{\nu_{\pi_{w_0}}}}} \sqrt{\zeta_{approx}} + \mathcal{O}(\lambda) + \epsilon.
\end{flalign*}

\section{Proof of Supporting Lemmas in Section \ref{sc: list_lemma_2}}\label{sc: lemmaproof2}

\begin{proof}[\textbf{Proof of Lemma \ref{valuefunction}}]
	The first part of \Cref{valuefunction}, which indicate the $L_Q$-Lipschitz property of state-action value function, has been established in \cite{xu2020improving}, here we show how to derive the second part of the result. By definition, we have $V_{\pi_w}(s)=\frac{1}{1-\gamma}\int_{a}\int_{(\hat{s},\hat{a})}r(\hat{s},\hat{a})dP^{w_{t+1}}_{(s,a)}(\hat{s},\hat{a})\pi_{w_{t+1}}(a|s)da$. It follows that
	\begin{flalign}
	&\lone{V_{\pi_w}(s) - V_{\pi_{w^\prime}}(s)}\nonumber\\
	&=\frac{1}{1-\gamma}\lone{\int_{a}\int_{(\hat{s},\hat{a})}r(\hat{s},\hat{a})dP^{\pi_{w}}_{(s,a)}(\hat{s},\hat{a})\pi_{w}(da|s) - \int_{a}\int_{(\hat{s},\hat{a})}r(\hat{s},\hat{a})dP^{\pi_{w^\prime}}_{(s,a)}(\hat{s},\hat{a})\pi_{w^\prime}(da|s)}\nonumber\\
	&\leq \frac{1}{1-\gamma}\left|\int_{a}\int_{(\hat{s},\hat{a})}r(\hat{s},\hat{a})dP^{\pi_{w}}_{(s,a)}(\hat{s},\hat{a})\pi_{w}(da|s) - \int_{a}\int_{(\hat{s},\hat{a})}r(\hat{s},\hat{a})dP^{\pi_{w}}_{(s,a)}(\hat{s},\hat{a})\pi_{w^\prime}(da|s) \notag\right. \nonumber\\
	& \phantom{=\;\;} \left. \qquad + \int_{a}\int_{(\hat{s},\hat{a})}r(\hat{s},\hat{a})dP^{\pi_{w}}_{(s,a)}(\hat{s},\hat{a})\pi_{w^\prime}(da|s) - \int_{a}\int_{(\hat{s},\hat{a})}r(\hat{s},\hat{a})dP^{\pi_{w^\prime}}_{(s,a)}(\hat{s},\hat{a})\pi_{w^\prime}(da|s)\right|\nonumber\\
	&\leq \frac{1}{1-\gamma} \lone{\int_{a}\int_{(\hat{s},\hat{a})}r(\hat{s},\hat{a})dP^{\pi_{w}}_{(s,a)}(\hat{s},\hat{a})\pi_{w}(da|s)- \int_{a}\int_{(\hat{s},\hat{a})}r(\hat{s},\hat{a})dP^{\pi_{w}}_{(s,a)}(\hat{s},\hat{a})\pi_{w^\prime}(da|s)}\nonumber\\
	&\quad + \frac{1}{1-\gamma}\lone{\int_{a}\int_{(\hat{s},\hat{a})}r(\hat{s},\hat{a})dP^{\pi_{w}}_{(s,a)}(\hat{s},\hat{a})\pi_{w^\prime}(da|s) - \int_{a}\int_{(\hat{s},\hat{a})}r(\hat{s},\hat{a})dP^{\pi_{w^\prime}}_{(s,a)}(\hat{s},\hat{a})\pi_{w^\prime}(da|s)}\nonumber\\
	&\leq \frac{1}{1-\gamma} \int_{a}\lone{\int_{(\hat{s},\hat{a})}r(\hat{s},\hat{a})dP^{\pi_{w}}_{(s,a)}(\hat{s},\hat{a})}\lone{\pi_{w}(da|s)-\pi_{w^\prime}(da|s)}\nonumber\\
	&\quad + \frac{1}{1-\gamma}\int_{a}\lone{\int_{(\hat{s},\hat{a})}r(\hat{s},\hat{a})dP^{\pi_w}_{(s,a)}(\hat{s},\hat{a}) - \int_{(\hat{s},\hat{a})}r(\hat{s},\hat{a})dP^{w^\prime}_{(s,a)}(\hat{s},\hat{a})} \pi_{w^\prime}(da|s)\nonumber\\
	&\leq \frac{r_{\max}C_\pi}{1-\gamma}\ltwo{w-w^\prime} + \frac{1}{1-\gamma}\int_{a}\left[\int_{(\hat{s},\hat{a})}r(\hat{s},\hat{a})\lone{dP^{w_{t+1}}_{(s,a)}(\hat{s},\hat{a})-dP^{w_t}_{(s,a)}(\hat{s},\hat{a})}\right] \pi_{w_t}(da|s)\nonumber\\
	&\leq \frac{r_{\max}C_\pi}{1-\gamma}\ltwo{w-w^\prime} + \frac{2r_{\max}}{1-\gamma}\int_{a}\lTV{P^{w_{t+1}}_{(s,a)}-P^{w_t}_{(s,a)}} \pi_{w_t}(da|s)\nonumber\\
	&\leq \frac{r_{\max}C_\pi}{1-\gamma}\ltwo{w-w^\prime} + \frac{2r_{\max}}{1-\gamma}\int_{a}C_\nu\ltwo{w-w^\prime} \pi_{w_t}(da|s)\nonumber\\
	&\leq L_V\ltwo{w-w^\prime},
	\end{flalign}
	which complete the proof.
\end{proof}

\begin{proof}[\textbf{Proof of Lemma \ref{lemma: support1}}]
	Consider critic's update at step $t-1$. We have
	\begin{flalign}
	\ltwo{\theta_t-\theta_{t-1}}&=\ltwo{\mcpi_{R_\theta}\big(\theta_{t-1} + \beta_{t-1}g_{t-1}(\theta_{t-1})\big)-\theta_{t-1}}\nonumber\\
	&\leq \ltwo{\theta_{t-1} + \beta_{t-1}g_{t-1}(\theta_{t-1})-\theta_{t-1}}\nonumber\\
	&=\beta_{t-1}\ltwo{g_{t-1}(\theta_{t-1})}\nonumber\\
	&=\beta_{t-1}\ltwo{-P^\lambda_{w_{t-1}}\theta_{t-1}+b_{w_{t-1}}}\nonumber\\
	&=\beta_{t-1}\ltwo{-P^\lambda_{w_{t-1}}(\theta_{t-1}-\theta^{\lambda*}_{t-1})-P^\lambda_{w_{t-1}}\theta^{\lambda*}_{t-1}+b_{w_{t-1}}}\nonumber\\
	&\leq \beta_{t-1}\Big(\ltwo{P^\lambda_{w_{t-1}}}\ltwo{\theta_{t-1}-\theta^{\lambda*}_{t-1}}+\ltwo{P^\lambda_{w_{t-1}}}\ltwo{\theta^{\lambda*}_{t-1}}+\ltwo{b_{w_{t-1}}}\Big)\nonumber\\
	&\leq \beta_{t-1}\Big[(C_\phi^2+\lambda)\ltwo{\theta_{t-1}-\theta^{\lambda*}_{t-1}}+(C_\phi^2+\lambda)R_\theta+\frac{2C_\phi r_{\max}}{1-\gamma}\Big].\label{incre1}
	\end{flalign}
	We can also obtain
	\begin{flalign}
	&\ltwo{\theta_t-\theta^{\lambda*}_t}\nonumber\\
	&=\ltwo{\theta_{t} - \theta_{t-1} + \theta_{t-1}-\theta^{\lambda*}_{t-1} + \theta^{\lambda*}_{t-1} - \theta^{\lambda*}_t }\nonumber\\
	&\leq \ltwo{\theta_{t} - \theta_{t-1}} + \ltwo{\theta_{t-1}-\theta^{\lambda*}_{t-1}} + \ltwo{\theta^{\lambda*}_{t-1} - \theta^{\lambda*}_t}\nonumber\\
	&\overset{(i)}{\leq} \beta_{t-1}\Big[(C_\phi^2+\lambda)\ltwo{\theta_{t-1}-\theta^{\lambda*}_{t-1}}+(C_\phi^2+\lambda)R_\theta+\frac{2C_\phi r_{\max}}{1-\gamma}\Big] + \ltwo{\theta_{t-1}-\theta^{\lambda*}_{t-1}} + C_3\alpha_{t-1}\nonumber\\
	&=[1+\beta_{t-1}(C_\phi^2+\lambda)]\ltwo{\theta_{t-1}-\theta^{\lambda*}_{t-1}} + \Big[(C_\phi^2+\lambda)R_\theta+\frac{2C_\phi r_{\max}}{1-\gamma}\Big]\beta_{t-1} + C_3\alpha_{t-1}\nonumber\\
	&\overset{(ii)}{\leq} [1+\beta_{t-1}(C_\phi^2+\lambda)]\ltwo{\theta_{t-1}-\theta^{\lambda*}_{t-1}} + \Big[(C_\phi^2+\lambda)R_\theta+\frac{2C_\phi r_{\max}}{1-\gamma} + \frac{C_3C_\alpha}{C_\beta}\Big]\beta_{t-1},\label{incre2}
	\end{flalign}
	where $C_3=L_\theta R_\theta \max\{1, C^2_\phi \}$, $(i)$ follows from \cref{incre1} and Lemma \ref{lemma: fixtracking}, $(ii)$ follows from the fact that $\alpha_t\leq \frac{C_\alpha}{C_\beta}\beta_t$ for all $t\geq0$. Let $C_4=(C_\phi^2+\lambda)R_\theta+\frac{2C_\phi r_{\max}}{1-\gamma} + \frac{C_3C_\alpha}{C_\beta}$. Applying \cref{incre2} recursively yields
	\begin{flalign}
	\ltwo{\theta_t-\theta^{\lambda*}_t}&\leq \left[ \prod_{i=1}^{\tau_t}\Big(1+\beta_{t-i}(C_\phi^2+\lambda)\Big)\right]\ltwo{\theta_{t-\tau_t}-\theta^{\lambda*}_{t-\tau_t}} + C_4\sum_{k=1}^{\tau_t}\left[ \prod_{i=1}^{k}\Big(1+\beta_{t-i}(C_\phi^2+\lambda)\Big)\right]\beta_{t-k}\nonumber\\
	&\overset{(i)}{\leq} \Big(1+\beta_{t-{\tau_t}}(C_\phi^2+\lambda)\Big)^{\tau_t}\ltwo{\theta_{t-\tau_t}-\theta^{\lambda*}_{t-\tau_t}} + C_4\beta_{t-{\tau_t}}\sum_{k=1}^{\tau_t}\Big(1+\beta_{t-{\tau_t}}(C_\phi^2+\lambda)\Big)^k\nonumber\\
	&\leq \Big(1+\beta_{t-{\tau_t}}(C_\phi^2+\lambda)\Big)^{\tau_t}\ltwo{\theta_{t-\tau_t}-\theta^{\lambda*}_{t-\tau_t}} + \frac{C_4}{C^2_\phi+\lambda}\left[ \Big(1+\beta_{t-{\tau_t}}(C_\phi^2+\lambda)\Big)^{\tau_t}-1 \right],\label{incre3}
	\end{flalign}
	where $(i)$ follows from the fact that $\{\beta_t\}$ is non-increasing. When $t$ is large enough, i.e., $t>\hat{t}$, such that $(C_\phi^2+\lambda)\beta_{t-{\tau_t}}\tau^2_t\leq \frac{1}{4}$, we have $(1+\beta_{t-{\tau_t}}(C_\phi^2+\lambda)\Big)^{\tau_t}\leq 1+2\beta_{t-{\tau_t}}\tau_t(C_\phi^2+\lambda)$ \cite{srikant2019finite}. Then we can further upper-bound \cref{incre3} as the follows
	\begin{flalign}
	\ltwo{\theta_t-\theta^{\lambda*}_t}&\leq \Big(1+\beta_{t-{\tau_t}}(C_\phi^2+\lambda)\Big)^{\tau_t}\ltwo{\theta_{t-\tau_t}-\theta^{\lambda*}_{t-\tau_t}} + 2C_4\beta_{t-{\tau_t}}\tau_t\label{incre4}\\
	&\leq\Big( 1+2\beta_{t-{\tau_t}}\tau_t(C_\phi^2+\lambda)\Big)\ltwo{\theta_{t-\tau_t}-\theta^{\lambda*}_{t-\tau_t}} + 2C_4\beta_{t-{\tau_t}}\tau_t.\nonumber
	\end{flalign}
\end{proof}

\begin{proof}[\textbf{Proof of Lemma \ref{lemma: support2}}]
	Following from critic's update in Algorithm \ref{algorithm_ttsac}, we have
	\begin{flalign}
	& \ltwo{\theta_t-\theta_{t-\tau_t}}\nonumber\\
	&\leq \sum_{i=t-\tau_t}^{t-1}\ltwo{\theta_{i+1}-\theta_i}=\sum_{i=t-\tau_t}^{t-1}\ltwo{\mcpi_{R_\theta}(\theta_i+\beta_i g_i(\theta_i))-\theta_i}\leq \sum_{i=t-\tau_t}^{t-1}\beta_i\ltwo{g_i(\theta_i)}\nonumber\\
	&\overset{(i)}{\leq} \sum_{i=t-\tau_t}^{t-1}\beta_i\Big[(C_\phi^2+\lambda)\ltwo{\theta_i-\theta^{\lambda*}_i}+(C_\phi^2+\lambda)R_\theta+\frac{2C_\phi r_{\max}}{1-\gamma}\Big]\nonumber\\
	&\leq \beta_{t-{\tau_t}}(C_\phi^2+\lambda)\Big[\sum_{i=t-\tau_t}^{t-1}\ltwo{\theta_i-\theta^{\lambda*}_i}\Big] + \Big((C_\phi^2+\lambda)R_\theta+\frac{2C_\phi r_{\max}}{1-\gamma}\Big)\beta_{t-{\tau_t}}\tau_t\nonumber\\
	&\overset{(ii)}{\leq} \beta_{t-{\tau_t}}(C_\phi^2+\lambda)\sum_{i=t-\tau_t}^{t-1}\Big[ \Big(1+\beta_{t-{\tau_t}}(C_\phi^2+\lambda)\Big)^{i-t+\tau_t}\ltwo{\theta_{t-\tau_t}-\theta^{\lambda*}_{t-\tau_t}} + 2C_4\beta_{t-{\tau_t}}(i-t+\tau_t) \Big]\nonumber\\
	&\quad + \Big((C_\phi^2+\lambda)R_\theta+\frac{2C_\phi r_{\max}}{1-\gamma}\Big)\beta_{t-{\tau_t}}\tau_t\nonumber\\
	&\overset{(iii)}{\leq} \beta_{t-{\tau_t}}(C_\phi^2+\lambda)\sum_{i=t-\tau_t}^{t-1}\Big[ \Big(1+\beta_{t-{\tau_t}}(C_\phi^2+\lambda)\Big)^{\tau_t-1}\ltwo{\theta_{t-\tau_t}-\theta^{\lambda*}_{t-\tau_t}} + 2C_4\beta_{t-{\tau_t}}(\tau_t-1) \Big]\nonumber\\
	&\quad + \Big((C_\phi^2+\lambda)R_\theta+\frac{2C_\phi r_{\max}}{1-\gamma}\Big)\beta_{t-{\tau_t}}\tau_t\nonumber\\
	&\overset{(iv)}{\leq} \beta_{t-{\tau_t}}\tau_t(C_\phi^2+\lambda)\Big[ \Big(1+2\beta_{t-{\tau_t}}\tau_t(C_\phi^2+\lambda)\Big)\ltwo{\theta_{t-\tau_t}-\theta^{\lambda*}_{t-\tau_t}} + 2C_4\beta_{t-{\tau_t}}\tau_t \Big]\nonumber\\
	&\quad + \Big((C_\phi^2+\lambda)R_\theta+\frac{2C_\phi r_{\max}}{1-\gamma}\Big)\beta_{t-{\tau_t}}\tau_t\nonumber\\
	&= \beta_{t-{\tau_t}}\tau_t(C_\phi^2+\lambda) \Big(1+2\beta_{t-{\tau_t}}\tau_t(C_\phi^2+\lambda)\Big)\ltwo{\theta_{t-\tau_t}-\theta^{\lambda*}_{t-\tau_t}} + 2C_4(C_\phi^2+\lambda)\beta^2_{t-{\tau_t}}\tau^2_t \nonumber\\
	&\quad + \Big((C_\phi^2+\lambda)R_\theta+\frac{2C_\phi r_{\max}}{1-\gamma}\Big)\beta_{t-{\tau_t}}\tau_t\nonumber\\
	&\overset{(v)}{\leq} \frac{3}{2}\beta_{t-{\tau_t}}\tau_t(C_\phi^2+\lambda)\ltwo{\theta_{t-\tau_t}-\theta^{\lambda*}_{t-\tau_t}}  + \Big(\frac{1}{2}C_4+(C_\phi^2+\lambda)R_\theta+\frac{2C_\phi r_{\max}}{1-\gamma}\Big)\beta_{t-{\tau_t}}\tau_t\nonumber\\
	&=\frac{3}{2}\beta_{t-{\tau_t}}\tau_t(C_\phi^2+\lambda)\ltwo{\theta_{t-\tau_t}-\theta^{\lambda*}_{t-\tau_t}}  + C_5\beta_{t-{\tau_t}}\tau_t,\nonumber
	\end{flalign}
	where $(i)$ follows from \cref{incre1}, $(ii)$ follows from \cref{incre4}, $(iii)$ follows from the fact that $\{ \beta_t \}$ is non-increasing, and $(iv)$ and $(v)$ follow from the fact that $(1+\beta_{t-{\tau_t}}(C_\phi^2+\lambda)\Big)^{\tau_t}\leq 1+2\beta_{t-{\tau_t}}\tau_t(C_\phi^2+\lambda)$ and $(C_\phi^2+\lambda)\beta_{t-{\tau_t}}\tau^2_t\leq \frac{1}{4}$ for all $t>\hat{t}$.
\end{proof}

\begin{proof}[\textbf{Proof of Lemma \ref{lemma: support3}}]
	Following from the update rule in Algorithm \ref{algorithm_ttsac}, we obtain
	\begin{flalign}
	\ltwo{\theta_{\hat{t}}-\theta^{\lambda*}_{\hat{t}}}^2&= \ltwo{\theta_0-\theta^{\lambda*}_0 + \theta^{\lambda*}_0 - \theta^{\lambda*}_{\hat{t}} + \theta_{\hat{t}} - \theta_0 }^2\nonumber\\
	&\leq 3\ltwo{\theta_0-\theta^{\lambda*}_0}^2 + 3\ltwo{\theta^{\lambda*}_0 - \theta^{\lambda*}_{\hat{t}}}^2 + 3\ltwo{\theta_{\hat{t}} - \theta_0}^2.\label{incre24}
	\end{flalign}
	Following from Lemma \ref{lemma: support2}, we obtain
	\begin{flalign}
	\ltwo{\theta_{\hat{t}}-\theta_0}\leq \frac{3}{2}\beta_0\hat{t}(C_\phi^2+\lambda)\ltwo{\theta_0-\theta^{\lambda*}_0}  + C_5\beta_0\hat{t}.\label{incre25}
	\end{flalign}
	Following from the fact that $a\leq b+c$ for positive $a$, $b$, $c$, which implies $a^2\leq 2b^2+2c^2$, we obtain
	\begin{flalign}
	\ltwo{\theta_{\hat{t}}-\theta_0}^2\leq \frac{9}{2}\beta^2_0\hat{t}^2(C_\phi^2+\lambda)^2\ltwo{\theta_0-\theta^{\lambda*}_0}^2  + 2C^2_5\beta^2_0\hat{t}^2.\label{incre26}
	\end{flalign}
	Following from Lemma \ref{lemma: fixtracking}, we have
	\begin{flalign}
	\ltwo{\theta^{\lambda*}_0 - \theta^{\lambda*}_{\hat{t}}}^2 &\leq (\sum_{i=0}^{\hat{t}-1}\ltwo{\theta^{\lambda*}_i - \theta^{\lambda*}_{i+1}})^2\leq C^2_3(\sum_{i=0}^{\hat{t}-1}\ltwo{w_i - w_{i+1}})^2\leq C^2_3 R^2_\theta \max\{1,C^4_\phi\} (\sum_{i=0}^{\hat{t}-1}\alpha_i)^2\nonumber\\
	&\leq C^2_3 R^2_\theta \max\{1,C^4_\phi\} C^2_\alpha \hat{t}^2.\label{incre27}
	\end{flalign}
	Substituting \cref{incre26} and \cref{incre27} into \cref{incre24} yields
	\begin{flalign}
	\ltwo{\theta_{\hat{t}}-\theta^{\lambda*}_{\hat{t}}}^2&\leq \Big[3 + \frac{27}{2}\beta^2_0\hat{t}^2(C_\phi^2+\lambda)^2\Big]\ltwo{\theta_0-\theta^{\lambda*}_0}^2 + 3C^2_3 R^2_\theta \max\{1,C^4_\phi\} C^2_\alpha \hat{t}^2  + 6C^2_5\beta^2_0\hat{t}^2\nonumber\\
	&= C_{16}\ltwo{\theta_0-\theta^{\lambda*}_0}^2 + C_{17}.
	\end{flalign}
\end{proof}

\begin{proof}[\textbf{Proof of Lemma \ref{bias_supp1}}]
	By the definitions of $P^\lambda_{w}(s,a)$ and $P^\lambda_{w^\prime}(s,a)$, we have
	\begin{flalign}
	P^\lambda_{w}(s,a) - P^\lambda_{w^\prime}(s,a)
	&= \phi_w(s,a) \phi_w(s,a)^\top - \phi_{w^\prime}(s,a) \phi_{w^\prime}(s,a)^\top \nonumber\\
	& = \phi_w(s,a) \phi_w(s,a)^\top - \phi_{w^
		\prime}(s,a) \phi_w(s,a)^\top + \phi_{w^\prime}(s,a) \phi_w(s,a)^\top  - \phi_{w^\prime}(s,a) \phi_{w^\prime}(s,a)^\top \nonumber\\
	& = ( \phi_w(s,a) - \phi_{w^\prime}(s,a) )\phi_w(s,a)^\top + \phi_{w}(s,a) (\phi_{w^\prime}(s,a) - \phi_{w^\prime}(s,a) )^\top, \nonumber
	\end{flalign}
	which implies
	\begin{flalign}
	&\ltwo{P^\lambda_{w}(s,a) - P^\lambda_{w^\prime}(s,a)}\nonumber\\
	&\leq \ltwo{( \phi_w(s,a) - \phi_{w^\prime}(s,a) )\phi_w(s,a)^\top} + \ltwo{\phi_{w}(s,a) (\phi_{w^\prime}(s,a) - \phi_{w^\prime}(s,a) )^\top}\nonumber\\
	&\leq \lF{( \phi_w(s,a) - \phi_{w^\prime}(s,a) )\phi_w(s,a)^\top} + \lF{\phi_{w}(s,a) (\phi_{w^\prime}(s,a) - \phi_{w^\prime}(s,a) )^\top} \nonumber\\
	&\leq 2C_\phi \ltwo{\phi_w(s,a) - \phi_{w^\prime}(s,a)} \nonumber\\
	&\leq 2C_\phi L_\phi \ltwo{w-w^\prime}. \nonumber
	\end{flalign}
	By the definitions of $b_{w}$ and $b_{w^\prime}$, we have
	\begin{flalign}
	&\hat{b}_{w}(s,a) -\hat{b}_{w^\prime}(s,a)\nonumber\\
	&= \phi_w(s,a)A_{\pi_w}(s,a) - \phi_{w^\prime}(s,a)A_{\pi_{w^\prime}}(s,a)\nonumber\\
	&= \phi_w(s,a)A_{\pi_w}(s,a) - \phi_{w^\prime}(s,a)A_{\pi_w}(s,a) + \phi_{w^\prime}(s,a)A_{\pi_w}(s,a) - \phi_{w^\prime}(s,a)A_{\pi_{w^\prime}}(s,a)\nonumber\\
	&= (\phi_w(s,a) - \phi_{w^\prime}(s,a))A_{\pi_w}(s,a) + \phi_{w^\prime}(s,a)(A_{\pi_w}(s,a) - A_{\pi_{w^\prime}}(s,a)).\nonumber\\
	&= (\phi_w(s,a) - \phi_{w^\prime}(s,a))A_{\pi_w}(s,a) + \phi_{w^\prime}(s,a)[Q_{\pi_w}(s,a) - Q_{\pi_{w^\prime}}(s,a) +  V_{\pi_{w^\prime}}(s) - V_{\pi_w}(s)],\nonumber\\
	\end{flalign}
	which implies
	\begin{flalign}
	&\ltwo{\hat{b}_{w}(s,a) -\hat{b}_{w^\prime}(s,a)}\nonumber\\
	&\leq \ltwo{\phi_w(s,a) - \phi_{w^\prime}(s,a)} A_{\pi_w}(s,a) + \ltwo{\phi_{w^\prime}(s,a)} \big[\lone{Q_{\pi_w}(s,a) - Q_{\pi_{w^\prime}}(s,a)} +  \lone{V_{\pi_{w^\prime}}(s) - V_{\pi_w}(s)} \big] \nonumber\\
	&\leq \left[\frac{L_\phi r_{\max}}{1-\gamma} + C_\phi (L_Q + L_V) \right]\ltwo{w-w^\prime}.
	\end{flalign}
\end{proof}

\begin{proof}[\textbf{Proof of Lemma \ref{bias_supp2}}]
	By the definitions of $\overline{P}^\lambda_{w}$ and $ \overline{P}^\lambda_{w^\prime}$, we have
	\begin{flalign}
	\overline{P}^\lambda_{w} - \overline{P}^\lambda_{w^\prime} &= \int_{(s,a)} \phi_w(s,a) \phi_w(s,a)^\top d\nu_{\pi_{w}} - \int_{(s,a)} \phi_{w^\prime}(s,a) \phi_{w^\prime}(s,a)^\top d\nu_{\pi_{w^\prime}} \nonumber\\
	&= \int_{(s,a)} \phi_w(s,a) \phi_w(s,a)^\top d\nu_{\pi_{w}} - \int_{(s,a)} \phi_w(s,a) \phi_w(s,a)^\top d\nu_{\pi_{w^\prime}} \nonumber\\
	&\quad + \int_{(s,a)} \phi_w(s,a) \phi_w(s,a)^\top d\nu_{\pi_{w^\prime}} - \int_{(s,a)} \phi_{w^\prime}(s,a) \phi_{w^\prime}(s,a)^\top d\nu_{\pi_{w^\prime}} \nonumber\\
	&= \int_{(s,a)} \phi_w(s,a) \phi_w(s,a)^\top (d\nu_{\pi_{w}} - d\nu_{\pi_{w^\prime}}) \nonumber\\
	&\quad + \int_{(s,a)} (\phi_w(s,a) \phi_w(s,a)^\top - \phi_{w^\prime}(s,a) \phi_{w^\prime}(s,a)^\top) d\nu_{\pi_{w^\prime}},\nonumber
	\end{flalign}
	which implies
	\begin{flalign}
	&\ltwo{\overline{P}^\lambda_{w}(s,a) - \overline{P}^\lambda_{w^\prime}(s,a)} \nonumber\\
	&\leq \ltwo{\int_{(s,a)} \phi_w(s,a) \phi_w(s,a)^\top (d\nu_{\pi_{w}} - d\nu_{\pi_{w^\prime}})} \nonumber\\
	&\quad + \ltwo{\int_{(s,a)} (\phi_w(s,a) \phi_w(s,a)^\top - \phi_{w^\prime}(s,a) \phi_{w^\prime}(s,a)^\top) d\nu_{\pi_{w^\prime}}}\nonumber\\
	&\leq \int_{(s,a)} \ltwo{\phi_w(s,a) \phi_w(s,a)^\top} \lone{d\nu_{\pi_{w}} - d\nu_{\pi_{w^\prime}}} \nonumber\\
	&\quad + \int_{(s,a)} \ltwo{\phi_w(s,a) \phi_w(s,a)^\top - \phi_{w^\prime}(s,a) \phi_{w^\prime}(s,a)^\top} d\nu_{\pi_{w^\prime}} \nonumber\\
	&\leq 2C^2_\phi \lTV{\nu_{\pi_{w}} - \nu_{\pi_{w^\prime}} }+ \int_{(s,a)} 2C_\phi L_\phi \ltwo{w-w^\prime} d\nu_{\pi_{w^\prime}} \nonumber\\
	&\overset{(i)}{\leq} 2(C_\phi^2C_\nu + C_\phi L_\phi) \ltwo{w-w^\prime},\nonumber
	\end{flalign}
	where $(i)$ follows from Lemma \ref{lemma: visit_dis}. By the definitions of $\overline{b}_{w}$ and $\overline{b}_{w^\prime}$, we have
	\begin{flalign}
	\overline{b}_{w} - \overline{b}_{w^\prime}&= \int_{(s,a)} \phi_{w}(s,a)A_{\pi_w}(s,a) d\nu_{\pi_{w}}(s,a) - \int_{(s,a)} \phi_{w^\prime}(s,a)A_{\pi_{w^\prime}}(s,a) d\nu_{\pi_{w^\prime}}(s,a) \nonumber\\
	&= \int_{(s,a)} \phi_{w}(s,a)A_{\pi_w}(s,a) d\nu_{\pi_{w}}(s,a) - \int_{(s,a)} \phi_{w}(s,a)A_{\pi_w}(s,a) d\nu_{\pi_{w^\prime}}(s,a) \nonumber\\
	&\quad + \int_{(s,a)} \phi_{w}(s,a)A_{\pi_w}(s,a) d\nu_{\pi_{w^\prime}}(s,a) - \int_{(s,a)} \phi_{w^\prime}(s,a)A_{\pi_{w^\prime}}(s,a) d\nu_{\pi_{w^\prime}}(s,a) \nonumber\\
	&= \int_{(s,a)} \phi_{w}(s,a)A_{\pi_w}(s,a) (d\nu_{\pi_{w}}(s,a) - d\nu_{\pi_{w^\prime}}(s,a)) \nonumber\\
	&\quad + \int_{(s,a)} (\phi_{w}(s,a)A_{\pi_w}(s,a) - \phi_{w^\prime}(s,a)A_{\pi_{w^\prime}}(s,a) )d\nu_{\pi_{w^\prime}}(s,a), \nonumber
	\end{flalign}
	which implies
	\begin{flalign}
	&\ltwo{\overline{b}_{w}(s,a) - \overline{b}_{w^\prime}(s,a)} \nonumber\\
	&\leq \ltwo{\int_{(s,a)} \phi_{w}(s,a)A_{\pi_w}(s,a) (d\nu_{\pi_{w}}(s,a) - d\nu_{\pi_{w^\prime}}(s,a))} \nonumber\\
	&\quad + \ltwo{\int_{(s,a)} (\phi_{w}(s,a)A_{\pi_w}(s,a) - \phi_{w^\prime}(s,a)A_{\pi_{w^\prime}}(s,a) )d\nu_{\pi_{w^\prime}}(s,a)}\nonumber\\
	&\leq \int_{(s,a)} \ltwo{\phi_{w}(s,a)A_{\pi_w}(s,a)} \lone{d\nu_{\pi_{w}}(s,a) - d\nu_{\pi_{w^\prime}}(s,a)}\nonumber\\
	&\quad + \int_{(s,a)} \ltwo{\phi_{w}(s,a)A_{\pi_w}(s,a) - \phi_{w^\prime}(s,a)A_{\pi_{w^\prime}}(s,a) }d\nu_{\pi_{w^\prime}}(s,a)\nonumber\\
	&\leq \frac{2C_\phi r_{\max}}{1-\gamma}\lTV{\nu_{\pi_{w}} - \nu_{\pi_{w^\prime}} } + \int_{(s,a)} \left[\frac{L_\phi r_{\max}}{1-\gamma} + C_\phi (L_Q + L_V) \right]\ltwo{w-w^\prime}d\nu_{\pi_{w^\prime}}(s,a)\nonumber\\
	&\leq \left[ \frac{2C_\phi C_\nu r_{\max}}{1-\gamma} + \frac{L_\phi r_{\max}}{1-\gamma} + C_\phi (L_Q + L_V) \right]\ltwo{w-w^\prime}.\nonumber
	\end{flalign}
\end{proof}

\begin{proof}[\textbf{Proof of Lemma \ref{lemma: delta}}]
	Similarly to the proof of Theorem 1 in \cite{zou2019finite}, here we construct an auxiliary MDP from $(t-\tau_t)$-th step to $t$-th step by following a fixed policy $\pi_{w_{t-\tau_t}}$. We denote the samples as $\{(\bs_i,\ba_i)\}$ ($t-\tau_t+1\leq i\leq t$):
	\begin{flalign}\label{auxMDP}
	(s_{t-\tau_t},a_{t-\tau_t})\overset{\pi_{w_{t-\tau_t}}}{\longrightarrow}(\bs_{t-\tau_t+1},\ba_{t-\tau_t+1})\overset{\pi_{w_{t-\tau_t}}}{\longrightarrow}\cdots\overset{\pi_{w_{t-\tau_t}}}{\longrightarrow}(\bs_t,\ba_t).
	\end{flalign}
	By definition we have
	\begin{flalign*}
	&\overline{P}^\lambda_{w_{t-\tau_t}}-\mE\big[P^\lambda_{w_{t-\tau_t}}|\mf_{t-\tau_t}\big]\\
	&=\int_{(s,a)}\phi_{w_{t-\tau_t}}(s,a)\phi_{w_{t-\tau_t}}(s,a)^\top d\nu_{\pi_{w_{t-\tau_t}}}(s,a)\nonumber\\
	&\quad - \int_{(s,a)}\phi_{w_{t-\tau_t}}(s,a)\phi_{w_{t-\tau_t}}(s,a)^\top d\msP(s_t=s,a_t=a|\mf_{t-\tau_t})\\
	&= \int_{(s,a)}\phi_{w_{t-\tau_t}}(s,a)\phi_{w_{t-\tau_t}}(s,a)^\top d\nu_{\pi_{w_{t-\tau_t}}}(s,a) \nonumber\\
	&\quad - \int_{(s,a)}\phi_{w_{t-\tau_t}}(s,a)\phi_{w_{t-\tau_t}}(s,a)^\top d\breve{\msP}(\bs_t=s,\ba_t=a|\mf_{t-\tau_t})\\
	&\quad + \int_{(s,a)}\phi_{w_{t-\tau_t}}(s,a)\phi_{w_{t-\tau_t}}(s,a)^\top d\breve{\msP}(\bs_t=s,\ba_t=a|\mf_{t-\tau_t}) \nonumber\\
	&\quad - \int_{(s,a)}\phi_{w_{t-\tau_t}}(s,a)\phi_{w_{t-\tau_t}}(s,a)^\top d\msP(s_t=s,a_t=a|\mf_{t-\tau_t})\\
	&=\int_{(s,a)}\phi_{w_{t-\tau_t}}(s,a)\phi_{w_{t-\tau_t}}(s,a)^\top [d\nu_{\pi_{w_{t-\tau_t}}}(s,a)-d\breve{\msP}(\bs_t=s,\ba_t=a|\mf_{t-\tau_t})]\\
	&\quad + \int_{(s,a)}\phi_{w_{t-\tau_t}}(s,a)\phi_{w_{t-\tau_t}}(s,a)^\top [d\breve{\msP}(\bs_t=s,\ba_t=a|\mf_{t-\tau_t})-d\msP(s_t=s,a_t=a|\mf_{t-\tau_t})],
	\end{flalign*}
	where $\msP(s_i,a_i|\mf)$ denotes the distribution of $(s_i,a_i)$ conditioned on the filtration $\mf$, and $\breve{\msP}(\breve{s}_i,\breve{a}_i|\mf)$ denotes the distribution of $(\breve{s}_i,\breve{a}_i)$ of the auxiliary MDP at $i$-th step in (\ref{auxMDP}) conditioned on the filtration $\mf$. Then we obtain
	\begin{flalign}
	&\ltwo{\overline{P}^\lambda_{w_{t-\tau_t}}-\mE\big[P^\lambda_{w_{t-\tau_t}}|\mf_{t-\tau_t}\big]}\nonumber\\
	&\leq \ltwo{\int_{(s,a)}\phi_{w_{t-\tau_t}}(s,a)\phi_{w_{t-\tau_t}}(s,a)^\top [d\nu_{\pi_{w_{t-\tau_t}}}(s,a)-d\msP(s_t=s,a_t=a|\mf_{t-\tau_t})]}\nonumber\\
	&\quad + \ltwo{\int_{(s,a)}\phi_{w_{t-\tau_t}}(s,a)\phi_{w_{t-\tau_t}}(s,a)^\top [d\breve{\msP}(\bs_t=s,\ba_t=a|\mf_{t-\tau_t})-d\msP(s_t=s,a_t=a|\mf_{t-\tau_t})]}\nonumber\\
	&\leq  \int_{(s,a)}\ltwo{\phi_{w_{t-\tau_t}}(s,a)\phi_{w_{t-\tau_t}}(s,a)^\top} \lone{d\nu_{\pi_{w_{t-\tau_t}}}(s,a)-d\msbP(\bs_t=s,\ba_t=a|\mf_{t-\tau_t})}\nonumber\\
	&\quad + \int_{(s,a)}\ltwo{\phi_{w_{t-\tau_t}}(s,a)\phi_{w_{t-\tau_t}}(s,a)^\top} \lone{d\breve{\msP}(\bs_t=s,\ba_t=a|\mf_{t-\tau_t})-d\msP(s_t=s,a_t=a|\mf_{t-\tau_t})}\nonumber\\
	&\leq 2C^2_\phi\Big( \lTV{\nu_{\pi_{w_{t-\tau_t}}}(\cdot,\cdot)-\breve{\msP}(\bs_t=s,\ba_t=a|\mf_{t-\tau_t})} \nonumber\\
	&\quad\qquad+ \lTV{\breve{\msP}(\bs_t=s,\ba_t=a|\mf_{t-\tau_t})-\msP(s_t=s,a_t=a|\mf_{t-\tau_t})} \Big).\label{incre11}
	\end{flalign}
	For the term $\lTV{\nu_{\pi_{w_{t-\tau_t}}}(\cdot,\cdot)-\breve{\msP}(\bs_t,\ba_t|\mf_{t-\tau_t})}$, due to Assumption \ref{ass2}, we have
	\begin{flalign}
	\lTV{\nu_{\pi_{w_{t-\tau_t}}}(\cdot,\cdot)-\breve{\msP}_t(\bs_t,\ba_t|\mf_{t-\tau_t})}\leq m\rho^{\tau_t}. \label{incre13}
	\end{flalign}
	For $\lTV{\breve{\msP}_t(\bs_t,\ba_t|\mf_{t-\tau_t})-\msP_t(s_t,a_t|\mf_{t-\tau_t})}$, by definition we obtain
	\begin{flalign}
	&\lTV{\msbP(\bs_t,\ba_t|\mf_{t-\tau_t})-\msP(s_t,a_t|\mf_{t-\tau_t})}\nonumber\\
	&\leq \lTV{\msbP(\bs_t|\mf_{t-\tau_t})\pi_{w_{t-\tau_t}}(\ba_t|\bs_t)-\msP(s_t|\mf_{t-\tau_t})\pi_{w_t}(a_t|s_t)}\nonumber\\
	&=\frac{1}{2}\int_{s}\int_{a} \lone{\msbP(\bs_t=s|\mf_{t-\tau_t})\pi_{w_{t-\tau_t}}(\ba_t=a|\bs_t=s)-\msP(s_t=s|\mf_{t-\tau_t})\pi_{w_t}(a_t=a|s_t=s)}dads\nonumber\\
	&=\frac{1}{2}\int_{s}\int_{a} \Big|\msbP(\bs_t=s|\mf_{t-\tau_t})\pi_{w_{t-\tau_t}}(\ba_t=a|\bs_t=s)-\msbP(\bs_t=s|\mf_{t-\tau_t})\pi_{w_t}(a_t=a|s_t=s) \nonumber\\
	&\qquad\qquad + \msbP(\bs_t=s|\mf_{t-\tau_t})\pi_{w_t}(a_t=a|s_t=s) - \msP(s_t=s|\mf_{t-\tau_t})\pi_{w_t}(a_t=a|s_t=s) \Big| dads\nonumber\\
	&\leq \frac{1}{2}\int_{s}\msbP(\bs_t=s|\mf_{t-\tau_t})\int_{a} \lone{\pi_{w_{t-\tau_t}}(\ba_t=a|\bs_t=s)-\pi_{w_t}(a_t=a|s_t=s)}dads\nonumber\\
	&\quad + \frac{1}{2}\int_{s}\lone{\msbP(\bs_t=s|\mf_{t-\tau_t})- \msP(s_t=s|\mf_{t-\tau_t})}\int_{a} \pi_{w_t}(a_t=a|s_t=s) dads\nonumber\\
	&\overset{(i)}{\leq} C_\pi\ltwo{w_{t-\tau_t}-w_t} + \frac{1}{2}\int_{s}\lone{\msbP(\bs_t=s|\mf_{t-\tau_t})- \msP(s_t=s|\mf_{t-\tau_t})}ds \nonumber\\
	&\overset{(ii)}{\leq} C_\pi R_\theta \max\{1,C^2_\phi\} \alpha_{t-\tau_t}\tau_t + \frac{1}{2}\int_{s}\lone{\msbP(\bs_t=s|\mf_{t-\tau_t})- \msP(s_t=s|\mf_{t-\tau_t})}ds\nonumber\\
	&= C_\pi R_\theta \max\{1,C^2_\phi\} \alpha_{t-\tau_t}\tau_t + \lTV{\msbP(\bs_t|\mf_{t-\tau_t})- \msP(s_t|\mf_{t-\tau_t})}. \label{incre14}
	\end{flalign}
	where $(i)$ follows from the third item in \Cref{ass1} and $(ii)$ follows from the fact that
	\begin{flalign*}
		\ltwo{w_{t-\tau_t}-w_t}\leq \sum_{i=t-\tau_t}^{t-1}\ltwo{w_{i+1}-w_i}\leq R_\theta\max\{1,C^2_\phi\}\sum_{i=t-\tau_t}^{t-1}\alpha_i \leq R_\theta\max\{1,C^2_\phi\}\alpha_{t-\tau_t}\tau_t.
	\end{flalign*}
	For $\msbP(\bs_t=s|\mf_{t-\tau_t})$. We then have
	\begin{flalign}
	&\msbP(\bs_t=\cdot|\mf_{t-\tau_t})\nonumber\\
	&=\int_{s} \msbP(\bs_{t-1}=s|\mf_{t-\tau_t})\msbP(\bs_t=\cdot|\bs_{t-1}=s)ds\nonumber\\
	&=\int_{s} \msbP(\bs_{t-1}=s|\mf_{t-\tau_t})\int_{a}\widetilde{\msP}(\bs_t=\cdot|\bs_{t-1}=s,\ba_{t-1}=a)\pi_{w_{t-\tau_t}}(\ba_{t-1}=a|\bs_{t-1}=s) dads.\label{incre15}
	\end{flalign}
	Similarly, for $\msP(s_t=s|\mf_{t-\tau_t})$, we have
	\begin{flalign}
	&\msP(s_t=\cdot|\mf_{t-\tau_t})\nonumber\\
	&=\int_{s} \msP(s_{t-1}=s|\mf_{t-\tau_t})\widetilde{P}(s_t=\cdot|s_{t-1}=s)ds\nonumber\\
	&=\int_{s} \msP(s_{t-1}=s|\mf_{t-\tau_t})\int_{a}\widetilde{\msP}(s_t=\cdot|s_{t-1}=s,a_{t-1}=a)\pi_{w_{t-\tau_t}}(a_{t-1}=a|s_{t-1}=s) dads.\label{incre16}
	\end{flalign}
	Then, combining \cref{incre15} and \cref{incre16} yields
	\begin{flalign}
	&\lone{\msbP(\bs_t=\cdot|\mf_{t-\tau_t})- \msP(s_t=\cdot|\mf_{t-\tau_t})}\nonumber\\
	&=\Bigg| \int_{s} \msbP(\bs_{t-1}=s|\mf_{t-\tau_t})\int_{a}\widetilde{\msP}(\bs_t=\cdot|\bs_{t-1}=s,\ba_{t-1}=a)\pi_{w_{t-\tau_t}}(\ba_{t-1}=a|\bs_{t-1}=s) dads\nonumber\\
	&\qquad - \int_{s} \msP(s_{t-1}=s|\mf_{t-\tau_t})\int_{a}\widetilde{\msP}(s_t=\cdot|s_{t-1}=s,a_{t-1}=a)\pi_{w_{t-1}}(a_{t-1}=a|s_{t-1}=s) dads \Bigg| \nonumber\\
	&\leq  \int_{s} \msbP(\bs_{t-1}=s|\mf_{t-\tau_t})  \int_{a}\Bigg|\widetilde{\msP}(\bs_t=\cdot|\bs_{t-1}=s,\ba_{t-1}=a)\pi_{w_{t-\tau_t}}(\ba_{t-1}=a|\bs_{t-1}=s) \nonumber\\
	&\quad\qquad - \int_{a}\widetilde{\msP}(s_t=\cdot|s_{t-1}=s,a_{t-1}=a)\pi_{w_{t-1}}(a_{t-1}=a|s_{t-1}=s) \Bigg| dads \nonumber\\
	&\quad + \int_{s} \Bigg|\msbP(\bs_{t-1}=s|\mf_{t-\tau_t})  - \msP(s_{t-1}=s|\mf_{t-\tau_t}) \Bigg|\nonumber\\
	&\qquad\int_{a}\widetilde{\msP}(s_t=\cdot|s_{t-1}=s,a_{t-1}=a)\pi_{w_{t-1}}(a_{t-1}=a|s_{t-1}=s) dads\nonumber\\
	&\leq  \int_{s} \msbP(\bs_{t-1}=s|\mf_{t-\tau_t})  \int_{a}\widetilde{\msP}(s_t=\cdot|s_{t-1}=s,a_{t-1}=a) |\pi_{w_{t-\tau_t}}(\ba_{t-1}=a|\bs_{t-1}=s) \nonumber\\
	&\qquad\qquad- \pi_{w_{t-1}}(a_{t-1}=a|s_{t-1}=s)|  dads \nonumber\\
	&\quad + \int_{s} \lone{\msbP(\bs_{t-1}=s|\mf_{t-\tau_t})  - \msP(s_{t-1}=s|\mf_{t-\tau_t})} ds\nonumber\\
	&\leq  \int_{s} \msbP(\bs_{t-1}=s|\mf_{t-\tau_t})  \int_{a} |\pi_{w_{t-\tau_t}}(\ba_{t-1}=a|\bs_{t-1}=s) - \pi_{w_{t-1}}(a_{t-1}=a|s_{t-1}=s)|  dads \nonumber\\
	&\quad + \int_{s} \lone{\msbP(\bs_{t-1}=s|\mf_{t-\tau_t})  - \msP(s_{t-1}=s|\mf_{t-\tau_t})} ds\nonumber\\
	&\leq 2C_\pi\ltwo{w_{t-\tau_t}-w_{t-1}} + \int_{s} \lone{\msbP(\bs_{t-1}=s|\mf_{t-\tau_t})  - \msP(s_{t-1}=s|\mf_{t-\tau_t})} ds\nonumber\\
	&\leq 2C_\pi\ltwo{w_{t-\tau_t}-w_{t-1}} +  2\lTV{\msbP(\bs_{t-1}|\mf_{t-\tau_t})  - \msP(s_{t-1}|\mf_{t-\tau_t})}.\nonumber
	\end{flalign}
	Then we have
	\begin{flalign}
	\lTV{\msbP(\bs_t|\mf_{t-\tau_t})- \msP(s_t|\mf_{t-\tau_t})}&=\frac{1}{2}\int_{s}\lone{\msbP(\bs_t=s|\mf_{t-\tau_t})- \msP(s_t=s|\mf_{t-\tau_t})}ds \nonumber\\
	&\leq C_\pi\ltwo{w_{t-\tau_t}-w_{t-1}} +  \lTV{\msbP(\bs_{t-1}|\mf_{t-\tau_t})  - \msP(s_{t-1}|\mf_{t-\tau_t})}. \label{incre17}
	\end{flalign}
	Applying \cref{incre17} recursively yields
	\begin{flalign}
	\lTV{\msbP(\bs_t|\mf_{t-\tau_t})- \msP(s_t|\mf_{t-\tau_t})}&\leq C_\pi\sum_{i=t-\tau_t}^{t-1}\ltwo{w_{t-\tau_t}-w_i}\leq C_\pi R_\theta \max\{1,C^2_\phi\} \sum_{i=t-\tau_t}^{t-1}\sum_{k=t-\tau_t}^{i}\alpha_k\nonumber\\
	&\leq C_\pi R_\theta \max\{1,C^2_\phi\} \alpha_{t-\tau_t}\tau^2_t.\label{incre18}
	\end{flalign}
	Substituting \cref{incre18} into \cref{incre14} and due to the fact $\tau_t\geq 1$, we obtain
	\begin{flalign}
	\lTV{\msbP(\bs_t,\ba_t|\mf_{t-\tau_t})-\msP(s_t,a_t|\mf_{t-\tau_t})}\leq 2C_\pi R_\theta\max\{1,C^2_\phi\}\alpha_{t-\tau_t}\tau^2_t.\label{incre19}
	\end{flalign}
	Substituting \cref{incre13} and \cref{incre19} into \cref{incre11} yields
	\begin{flalign}
	\ltwo{\overline{P}^\lambda_{w_t}-\mE\big[P^\lambda_{w_t}|\mf_{t-\tau_t}\big]}&\leq 2C^2_\phi \big[ m\rho^{\tau_t} + 2C_\pi R_\theta\max\{1,C^2_\phi\}\alpha_{t-\tau_t}\tau^2_t\big]\nonumber\\
	&\leq 2C^2_\phi \big[ 2C_\pi R_\theta\max\{1,C^2_\phi\} + 1\big]\alpha_{t-\tau_t}\tau^2_t.\nonumber
	\end{flalign}
	Then, consider $\Delta_{b,\tau_t}=\ltwo{\overline{b}_{w_{t-\tau_t}}-\mE\big[b_{w_{t-\tau_t}}|\mf_{t-\tau_t}\big]}$. By definition, we have
	\begin{flalign}
	&\overline{b}_{w_{t-\tau_t}}-\mE\big[b_{w_{t-\tau_t}}|\mf_{t-\tau_t}\big]\nonumber\\
	&=\int_{(s,a)}Q_{\pi_{w_{t-\tau_t}}}(s,a)\phi_{w_{t-\tau_t}}(s,a)d\nu_{w_{t-\tau_t}}(s,a)-\int_{s}V_{\pi_{w_{t-\tau_t}}}(s)\left[\int_{a}\phi_{w_{t-\tau_t}}(s,a)\pi_{w_{t-\tau_t}}(s,a)da\right] d\nu_{w_{t-\tau_t}}(s)\nonumber\\
	&\quad -\Bigg[\int_{(s,a)}Q_{\pi_{w_{t-\tau_t}}}(s,a)\phi_{w_{t-\tau_t}}(s,a)d\msP(s_t=s,a_t=a|\mf_{t-\tau_t}) \nonumber\\ &\quad -\int_{s}V_{\pi_{w_{t-\tau_t}}}(s)\left[\int_{a^\prime}\phi_{w_{t-\tau_t}}(s,a^\prime)\pi_{w_{t-\tau_t}}(s,a^\prime)da^\prime\right]d\msP(s_t=s|\mf_{t-\tau_t})\Bigg]\nonumber\\
	&=\int_{(s,a)}Q_{\pi_{w_{t-\tau_t}}}(s,a)\phi_{w_{t-\tau_t}}(s,a)\big[d\nu_{w_{t-\tau_t}}(s,a)-d\msP(s_t=s,a_t=a|\mf_{t-\tau_t})\big]\nonumber\\
	&\quad + \int_{(s,a)}V_{\pi_{w_{t-\tau_t}}}(s)\phi_{w_{t-\tau_t}}(s,a)\big[d\msP(s_t=s,a_t=a|\mf_{t-\tau_t})-d\nu_{w_{t-\tau_t}}(s,a)\big]\nonumber\\
	&=\int_{(s,a)}A_{\pi_{w_{t-\tau_t}}}(s,a)\phi_{w_{t-\tau_t}}(s,a)\big[d\nu_{w_{t-\tau_t}}(s,a)-d\msP(s_t=s,a_t=a|\mf_{t-\tau_t})\big]\nonumber\\
	&= \int_{(s,a)}A_{\pi_{w_{t-\tau_t}}}(s,a)\phi_{w_{t-\tau_t}}(s,a) [d\nu_{\pi_{w_{t-\tau_t}}}(s,a)-d\breve{\msP}(\bs_t=s,\ba_t=a|\mf_{t-\tau_t})]\nonumber\\
	&\quad + \int_{(s,a)}A_{\pi_{w_{t-\tau_t}}}(s,a)\phi_{w_{t-\tau_t}}(s,a) [d\breve{\msP}(\bs_t=s,\ba_t=a|\mf_{t-\tau_t})-d\msP(s_t=s,a_t=a|\mf_{t-\tau_t})].\nonumber
	\end{flalign}
	Then, we obtain
	\begin{flalign}
	&\ltwo{\overline{b}_{w_t}-\mE\big[b_{w_t}|\mf_{t-\tau_t}\big]}\nonumber\\
	&\leq \ltwo{\int_{(s,a)}A_{\pi_{w_{t-\tau_t}}}(s,a)\phi_{w_{t-\tau_t}}(s,a) [d\nu_{\pi_{w_{t-\tau_t}}}(s,a)-d\msP(s_t=s,a_t=a|\mf_{t-\tau_t})]}\nonumber\\
	&\quad + \ltwo{\int_{(s,a)}A_{\pi_{w_{t-\tau_t}}}(s,a)\phi_{w_{t-\tau_t}}(s,a) [d\breve{\msP}(\bs_t=s,\ba_t=a|\mf_{t-\tau_t})-d\msP(s_t=s,a_t=a|\mf_{t-\tau_t})]}\nonumber\\
	&\leq \int_{(s,a)}\ltwo{A_{\pi_{w_{t-\tau_t}}}(s,a)\phi_{w_{t-\tau_t}}(s,a)} \lone{d\nu_{\pi_{w_{t-\tau_t}}}(s,a)-d\msbP(\bs_t=s,\ba_t=a|\mf_{t-\tau_t})}\nonumber\\
	&\quad + \int_{(s,a)}\ltwo{A_{\pi_{w_{t-\tau_t}}}(s,a)\phi_{w_{t-\tau_t}}(s,a)} \lone{d\breve{\msP}(\bs_t=s,\ba_t=a|\mf_{t-\tau_t})-d\msP(s_t=s,a_t=a|\mf_{t-\tau_t})}\nonumber\\
	&\leq \frac{2C_\phi r_{\max}}{1-\gamma}\Big( \lTV{\nu_{\pi_{w_{t-\tau_t}}}(\cdot,\cdot)-\breve{\msP}(\bs_t=s,\ba_t=a|\mf_{t-\tau_t})} \nonumber\\
	&\quad + \lTV{\breve{\msP}(\bs_t=s,\ba_t=a|\mf_{t-\tau_t})-\msP(s_t=s,a_t=a|\mf_{t-\tau_t})} \Big).\label{incre20}
	\end{flalign}
	Substituting \cref{incre13} and \cref{incre19} into \cref{incre20} yields
	\begin{flalign}
	\ltwo{\overline{b}_{w_t}-\mE\big[b_{w_t}|\mf_{t-\tau_t}\big]}&\leq \frac{2C_\phi r_{\max}}{1-\gamma}\big[ m\rho^{\tau_t} + 2C_\pi R_\theta\max\{1,C^2_\phi\}\alpha_{t-\tau_t}\tau^2_t\big]\nonumber\\
	&\leq  \frac{2C_\phi r_{\max}}{1-\gamma}\big[ 2C_\pi R_\theta\max\{1,C^2_\phi\} + 1\big]\alpha_{t-\tau_t}\tau^2_t.\nonumber
	\end{flalign}
\end{proof}

\begin{proof}[\textbf{Proof of Lemma \ref{lemma: bias}}]
	By the definition of $\overline{g}_t(\theta^t)$ and $g_t(\theta_t)$, we obtain
	\begin{flalign}
	&\langle g_t(\theta_t)-\overline{g}_t(\theta_t), \theta_t-\theta^{\lambda*}_t \rangle\nonumber\\
	&\overset{(i)}{=}\langle \hat{g}_{t-\tau_t}(\theta_t)-\overline{g}_{t-\tau_t}(\theta_t), \theta_t-\theta^{\lambda*}_t \rangle + \langle g_t(\theta_t)-\hat{g}_{t-\tau_t}(\theta_t), \theta_t-\theta^{\lambda*}_t \rangle + \langle \overline{g}_{t-\tau_t}(\theta_t)-\overline{g}_t(\theta_t), \theta_t-\theta^{\lambda*}_t \rangle\nonumber\\
	&=\langle \hat{g}_{t-\tau_t}(\theta_t)-\overline{g}_{t-\tau_t}(\theta_t), \theta_t-\theta^{\lambda*}_{t-\tau_t} \rangle + \langle \hat{g}_{t-\tau_t}(\theta_t)-\overline{g}_{t-\tau_t}(\theta_t), \theta^{\lambda*}_{t-\tau_t} - \theta^{\lambda*}_t \rangle \nonumber\\
	&\quad + \langle g_t(\theta_t)-\hat{g}_{t-\tau_t}(\theta_t), \theta_t-\theta^{\lambda*}_t \rangle  + \langle \overline{g}_{t-\tau_t}(\theta_t)-\overline{g}_t(\theta_t), \theta_t-\theta^{\lambda*}_t \rangle \nonumber\\
	&=\langle -P^\lambda_{w_{t-\tau_t}}\theta_t + \hat{b}_{w_{t-\tau_t}} + \overline{P}^\lambda_{w_{t-\tau_t}}\theta_t - \overline{b}_{w_{t-\tau_t}}, \theta_t-\theta^{\lambda*}_{t-\tau_t} \rangle + \langle \hat{g}_{t-\tau_t}(\theta_t)-\overline{g}_{t-\tau_t}(\theta_t), \theta^{\lambda*}_{t-\tau_t} - \theta^{\lambda*}_t \rangle \nonumber\\
	&\quad + \langle g_t(\theta_t)-\hat{g}_{t-\tau_t}(\theta_t), \theta_t-\theta^{\lambda*}_t \rangle + \langle \overline{g}_{t-\tau_t}(\theta_t)-\overline{g}_t(\theta_t), \theta_t-\theta^{\lambda*}_t \rangle \nonumber\\
	&=\langle \big(\overline{P}^\lambda_{w_{t-\tau_t}}-P^\lambda_{w_{t-\tau_t}}\big)(\theta_t-\theta^{\lambda*}_{t-\tau_t}), \theta_t-\theta^{\lambda*}_{t-\tau_t} \rangle + \langle \big(\overline{P}^\lambda_{w_{t-\tau_t}}-P^\lambda_{w_{t-\tau_t}}\big)\theta^{\lambda*}_{t-\tau_t} + (\hat{b}_{w_{t-\tau_t}}-\overline{b}_{w_{t-\tau_t}}), \theta_t-\theta^{\lambda*}_{t-\tau_t} \rangle \nonumber\\
	&\quad + \langle \hat{g}_{t-\tau_t}(\theta_t)-\overline{g}_{t-\tau_t}(\theta_t), \theta^{\lambda*}_{t-\tau_t} - \theta^{\lambda*}_t \rangle + \langle g_t(\theta_t)-\hat{g}_{t-\tau_t}(\theta_t), \theta_t-\theta^{\lambda*}_t \rangle + \langle \overline{g}_{t-\tau_t}(\theta_t)-\overline{g}_t(\theta_t), \theta_t-\theta^{\lambda*}_t \rangle, \label{bias_bound1}
	\end{flalign}
	where in $(i)$ we define $\hat{g}_{t-\tau_t}(\theta_t)\coloneqq -P^\lambda_{w_{t-\tau_t}}\theta_t + \hat{b}_{w_{t-\tau_t}}$, with $P^\lambda_{w_{t-\tau_t}}\coloneqq\phi_{w_{t-\tau_t}}(s_t,a_t)\phi_{w_{t-\tau_t}}(s_t,a_t)^\top + \lambda I$ and $\hat{b}_{w_{t-\tau_t}}\coloneqq \phi_{w_{t-\tau_t}}(s_t,a_t) A_{\pi_{w_{t-\tau_t}}}(s_t,a_t)$. Then we bound the three terms in the right-hand side of \cref{bias_bound1} as follows. For the first term in \cref{bias_bound1} we have
	\begin{flalign}
	&\langle \big(\overline{P}^\lambda_{w_{t-\tau_t}}-P^\lambda_{w_{t-\tau_t}}\big)(\theta_t-\theta^{\lambda*}_{t-\tau_t}), \theta_t-\theta^{\lambda*}_{t-\tau_t} \rangle\nonumber\\
	&=\langle \big(\overline{P}^\lambda_{w_{t-\tau_t}}-P^\lambda_{w_{t-\tau_t}}\big)(\theta_{t-\tau_t}-\theta^{\lambda*}_{t-\tau_t}), \theta_t-\theta^{\lambda*}_{t-\tau_t} \rangle + \langle \big(\overline{P}^\lambda_{w_{t-\tau_t}}-P^\lambda_{w_{t-\tau_t}}\big)(\theta_t-\theta_{t-\tau_t}), \theta_t-\theta_{t-\tau_t} \rangle \nonumber\\
	&\quad + \langle \big(\overline{P}^\lambda_{w_{t-\tau_t}}-P^\lambda_{w_{t-\tau_t}}\big)(\theta_t-\theta_{t-\tau_t}), \theta_{t-\tau_t}-\theta^{\lambda*}_{t-\tau_t} \rangle \nonumber\\
	&=\langle \big(\overline{P}^\lambda_{w_{t-\tau_t}}-P^\lambda_{w_{t-\tau_t}}\big)(\theta_{t-\tau_t}-\theta^{\lambda*}_{t-\tau_t}), \theta_{t-\tau_t}-\theta^{\lambda*}_{t-\tau_t} \rangle + \langle \big(\overline{P}^\lambda_{w_{t-\tau_t}}-P^\lambda_{w_{t-\tau_t}}\big)(\theta_{t-\tau_t}-\theta^{\lambda*}_{t-\tau_t}), \theta_t-\theta_{t-\tau_t} \rangle \nonumber\\
	&\quad + \langle \big(\overline{P}^\lambda_{w_{t-\tau_t}}-P^\lambda_{w_{t-\tau_t}}\big)(\theta_t-\theta_{t-\tau_t}), \theta_t-\theta_{t-\tau_t} \rangle + \langle \big(\overline{P}^\lambda_{w_{t-\tau_t}}-P^\lambda_{w_{t-\tau_t}}\big)(\theta_t-\theta_{t-\tau_t}), \theta_{t-\tau_t}-\theta^{\lambda*}_{t-\tau_t} \rangle \nonumber \\
	&=\langle \big(\overline{P}^\lambda_{w_{t-\tau_t}}-P^\lambda_{w_{t-\tau_t}}\big)(\theta_{t-\tau_t}-\theta^{\lambda*}_{t-\tau_t}), \theta_{t-\tau_t}-\theta^{\lambda*}_{t-\tau_t} \rangle + 2\langle \big(\overline{P}^\lambda_{w_{t-\tau_t}}-P^\lambda_{w_{t-\tau_t}}\big)(\theta_{t-\tau_t}-\theta^{\lambda*}_{t-\tau_t}), \theta_t-\theta_{t-\tau_t} \rangle \nonumber\\
	&\quad + \langle \big(\overline{P}^\lambda_{w_{t-\tau_t}}-P^\lambda_{w_{t-\tau_t}}\big)(\theta_t-\theta_{t-\tau_t}), \theta_t-\theta_{t-\tau_t} \rangle \nonumber. \label{bias_bound2}\\
	\end{flalign}
	Taking expectation with respect to the filtration $\mf_{t-\tau_t}$ on both sides of \cref{bias_bound2} yields
	\begin{flalign}
	\mE\big[\langle \big(\overline{P}^\lambda_{w_{t-\tau_t}}&-P^\lambda_{w_{t-\tau_t}}\big)(\theta_{t-\tau_t}-\theta^{\lambda*}_{t-\tau_t}), \theta_{t-\tau_t}-\theta^{\lambda*}_{t-\tau_t} \rangle|\mf_{t-\tau_t}\big]\nonumber\\
	&\leq \langle \big(\overline{P}^\lambda_{w_{t-\tau_t}}-\mE\big[P^\lambda_{w_{t-\tau_t}}\big|\mf_{t-\tau_t}]\big)(\theta_{t-\tau_t}-\theta^{\lambda*}_{t-\tau_t}), \theta_{t-\tau_t}-\theta^{\lambda*}_{t-\tau_t} \rangle\nonumber\\
	&\leq \ltwo{\overline{P}^\lambda_{w_{t-\tau_t}}-\mE\big[P^\lambda_{w_{t-\tau_t}}\big|\mf_{t-\tau_t}]} \ltwo{ \theta_{t-\tau_t}-\theta^{\lambda*}_{t-\tau_t}}^2\nonumber\\
	&\overset{(i)}{=}\Delta_{P,\tau_t} \ltwo{ \theta_{t-\tau_t}-\theta^{\lambda*}_{t-\tau_t}}^2.\label{incre5}
	\end{flalign}
	where in $(i)$ we denote $\Delta_{P,\tau_t}\coloneqq \ltwo{\overline{P}^\lambda_{w_{t-\tau_t}}-\mE\big[P^\lambda_{w_{t-\tau_t}}|\mf_{t-\tau_t}\big]}$. Taking expectation on both sides of \cref{incre5} conditioned on $\mf_{t-\tau_t}$ yields
	\begin{flalign}
	\mE\big[\langle \big(\overline{P}^\lambda_{w_{t-\tau_t}}&-P^\lambda_{w_{t-\tau_t}}\big)(\theta_{t-\tau_t}-\theta^{\lambda*}_{t-\tau_t}), \theta_t-\theta_{t-\tau_t} \rangle |\mf_{t-\tau_t}\big]\nonumber\\
	&\leq \mE\left[ \ltwo{\overline{P}^\lambda_{w_{t-\tau_t}}-P^\lambda_{w_{t-\tau_t}}} \ltwo{\theta_t-\theta_{t-\tau_t}}  |\mf_{t-\tau_t}\right]\ltwo{\theta_{t-\tau_t}-\theta^{\lambda*}_{t-\tau_t}}\nonumber\\
	&\leq\mE\left[ \Big(\ltwo{\overline{P}^\lambda_{w_{t-\tau_t}}}+\ltwo{P^\lambda_{w_{t-\tau_t}}}\Big) \ltwo{\theta_t-\theta_{t-\tau_t}}  |\mf_{t-\tau_t}\right]\ltwo{\theta_{t-\tau_t}-\theta^{\lambda*}_{t-\tau_t}}\nonumber\\
	&\leq2(C_\phi^2+\lambda)\mE\left[ \ltwo{\theta_t-\theta_{t-\tau_t}}  |\mf_{t-\tau_t}\right]\ltwo{\theta_{t-\tau_t}-\theta^{\lambda*}_{t-\tau_t}}\nonumber\\
	&\overset{(i)}{\leq} 3(C_\phi^2+\lambda)^2\beta_{t-{\tau_t}}\tau_t\ltwo{\theta_{t-\tau_t}-\theta^{\lambda*}_{t-\tau_t}}^2  + 2C_5(C_\phi^2+\lambda)\beta_{t-{\tau_t}}\tau_t\ltwo{\theta_{t-\tau_t}-\theta^{\lambda*}_{t-\tau_t}}\nonumber\\
	&\overset{(ii)}{\leq} 3(C_\phi^2+\lambda)^2\beta_{t-{\tau_t}}\tau_t\ltwo{\theta_{t-\tau_t}-\theta^{\lambda*}_{t-\tau_t}}^2  + C_5(C_\phi^2+\lambda)\beta_{t-{\tau_t}}\tau_t\big(\ltwo{\theta_{t-\tau_t}-\theta^{\lambda*}_{t-\tau_t}}^2+1\big)\nonumber\\
	&=\Big[3(C_\phi^2+\lambda)+C_5\Big](C_\phi^2+\lambda)\beta_{t-{\tau_t}}\tau_t\ltwo{\theta_{t-\tau_t}-\theta^{\lambda*}_{t-\tau_t}}^2  + C_5(C_\phi^2+\lambda)\beta_{t-{\tau_t}}\tau_t,\label{incre6}
	\end{flalign}
	where $(i)$ follows from Lemma \ref{lemma: support2} and $(ii)$ follows from the fact that $2x\leq 1+x^2$. Furthermore,
	\begin{flalign}
	\mE\big[\langle \big(\overline{P}^\lambda_{w_{t-\tau_t}}&-P^\lambda_{w_{t-\tau_t}}\big)(\theta_t-\theta_{t-\tau_t}), \theta_t-\theta_{t-\tau_t} \rangle|\mf_{t-\tau_t}\big]\nonumber\\
	&\leq \mE\big[ \ltwo{\overline{P}^\lambda_{w_{t-\tau_t}}-P^\lambda_{w_{t-\tau_t}}}\ltwo{\theta_t-\theta_{t-\tau_t}}^2 |\mf_{t-\tau_t}\big]\nonumber\\
	&\leq  \mE\left[ \left(\ltwo{\overline{P}^\lambda_{w_{t-\tau_t}}}+\ltwo{P^\lambda_{w_{t-\tau_t}}}\right)\ltwo{\theta_t-\theta_{t-\tau_t}}^2 |\mf_{t-\tau_t}\right]\nonumber\\
	&\leq 2(C_\phi^2+\lambda)\mE\left[ \ltwo{\theta_t-\theta_{t-\tau_t}}^2 |\mf_{t-\tau_t}\right].\label{incre7}
	\end{flalign}
	Combining \cref{incre5}, \cref{incre6} and \cref{incre7} and applying Lemma \ref{lemma: delta} yield the following bound:
	\begin{flalign}
	\mE\big[\langle \big(\overline{P}^\lambda_{w_{t-\tau_t}}&-P^\lambda_{w_{t-\tau_t}}\big)(\theta_t-\theta^{\lambda*}_{t-\tau_t}), \theta_t-\theta^{\lambda*}_{t-\tau_t} \rangle|\mf_{t-\tau_t}\big]\nonumber\\
	&\leq \left\{C_6\alpha_{t-\tau_t}\tau^2_t+\left[\big(3(C_\phi^2+\lambda)+C_5\big)\right](C_\phi^2+\lambda)\beta_{t-{\tau_t}}\tau_t\right\}\ltwo{\theta_{t-\tau_t}-\theta^{\lambda*}_{t-\tau_t}}^2 \nonumber\\
	&\quad + 2(C_\phi^2+\lambda)\mE\left[ \ltwo{\theta_t-\theta_{t-\tau_t}}^2 |\mf_{t-\tau_t}\right] + C_5(C_\phi^2+\lambda)\beta_{t-{\tau_t}}\tau_t\nonumber\\
	&\leq \left\{\frac{C_6C_\alpha}{C_\beta}+\left[\big(3(C_\phi^2+\lambda)+C_5\big)\right](C_\phi^2+\lambda)\right\}\beta_{t-{\tau_t}}\tau^2_t\ltwo{\theta_{t-\tau_t}-\theta^{\lambda*}_{t-\tau_t}}^2 \nonumber\\
	&\quad + 2(C_\phi^2+\lambda)\mE\left[ \ltwo{\theta_t-\theta_{t-\tau_t}}^2 |\mf_{t-\tau_t}\right] + C_5(C_\phi^2+\lambda)\beta_{t-{\tau_t}}\tau^2_t\label{incre8}
	\end{flalign}
	Consider the second term in \cref{bias_bound1}, and we have
	\begin{flalign}
	\langle \big(\overline{P}^\lambda_{w_{t-\tau_t}}&-P^\lambda_{w_{t-\tau_t}}\big)\theta^{\lambda*}_{t-\tau_t} + (\hat{b}_{w_{t-\tau_t}}-\overline{b}_{w_{t-\tau_t}}), \theta_t-\theta^{\lambda*}_{t-\tau_t} \rangle\nonumber\\
	&=\langle \big(\overline{P}^\lambda_{w_{t-\tau_t}}-P^\lambda_{w_{t-\tau_t}}\big)\theta^{\lambda*}_{t-\tau_t}, \theta_t-\theta^{\lambda*}_{t-\tau_t} \rangle + \langle \hat{b}_{w_{t-\tau_t}}-\overline{b}_{w_{t-\tau_t}}, \theta_t-\theta^{\lambda*}_{t-\tau_t} \rangle\nonumber\\
	&=\langle \big(\overline{P}^\lambda_{w_{t-\tau_t}}-P^\lambda_{w_{t-\tau_t}}\big)\theta^{\lambda*}_{t-\tau_t}, \theta_t-\theta_{t-\tau_t} \rangle + \langle \big(\overline{P}^\lambda_{w_{t-\tau_t}}-P^\lambda_{w_{t-\tau_t}}\big)\theta^{\lambda*}_{t-\tau_t},\theta_{t-\tau_t}-\theta^{\lambda*}_{t-\tau_t} \rangle\nonumber\\
	&\quad + \langle \hat{b}_{w_{t-\tau_t}}-\overline{b}_{w_{t-\tau_t}}, \theta_t-\theta_{t-\tau_t} \rangle + \langle \hat{b}_{w_{t-\tau_t}}-\overline{b}_{w_{t-\tau_t}}, \theta_{t-\tau_t}-\theta^{\lambda*}_{t-\tau_t} \rangle. \label{bias_bound3}
	\end{flalign}
	Taking expectation on both sides of \cref{bias_bound3} conditioned on $\mf_{t-\tau_t}$ yields
	\begin{flalign}
	&\mE\big[\langle \big(\overline{P}^\lambda_{w_{t-\tau_t}}-P^\lambda_{w_{t-\tau_t}}\big)\theta^{\lambda*}_{t-\tau_t} + (\hat{b}_{w_{t-\tau_t}}-\overline{b}_{w_{t-\tau_t}}), \theta_t-\theta^{\lambda*}_{t-\tau_t} \rangle|\mf_{t-\tau_t}\big]\nonumber\\
	&=\mE\big[\langle \big(\overline{P}^\lambda_{w_{t-\tau_t}}-P^\lambda_{w_{t-\tau_t}}\big)\theta^{\lambda*}_{t-\tau_t}, \theta_t-\theta_{t-\tau_t} \rangle|\mf_{t-\tau_t}\big] + \mE\big[\langle \big(\overline{P}^\lambda_{w_{t-\tau_t}}-P^\lambda_{w_{t-\tau_t}}\big)\theta^{\lambda*}_{t-\tau_t},\theta_{t-\tau_t}-\theta^{\lambda*}_{t-\tau_t} \rangle|\mf_{t-\tau_t}\big]\nonumber\\
	&\quad + \mE\big[\langle \hat{b}_{w_{t-\tau_t}}-\overline{b}_{w_{t-\tau_t}}, \theta_t-\theta_{t-\tau_t} \rangle|\mf_{t-\tau_t}\big] + \mE\big[\langle \hat{b}_{w_{t-\tau_t}}-\overline{b}_{w_{t-\tau_t}}, \theta_{t-\tau_t}-\theta^{\lambda*}_{t-\tau_t} \rangle|\mf_{t-\tau_t}\big]\nonumber\\
	&\leq \mE\big[\ltwo{\overline{P}^\lambda_{w_{t-\tau_t}}-P^\lambda_{w_{t-\tau_t}}} \ltwo{\theta^{\lambda*}_{t-\tau_t}} \ltwo{\theta_t-\theta_{t-\tau_t}}|\mf_{t-\tau_t}\big] \nonumber\\
	&\quad +  \ltwo{\overline{P}^\lambda_{w_{t-\tau_t}}-\mE\big[P^\lambda_{w_{t-\tau_t}}|\mf_{t-\tau_t}\big]}\ltwo{\theta^{\lambda*}_{t-\tau_t}} \ltwo{\theta_{t-\tau_t}-\theta^{\lambda*}_{t-\tau_t}} \nonumber\\
	&\quad + \mE\big[\ltwo{\hat{b}_{w_{t-\tau_t}}-\overline{b}_{w_{t-\tau_t}}} \ltwo{\theta_t-\theta_{t-\tau_t}} |\mf_{t-\tau_t}\big] + \ltwo{\mE\big[\hat{b}_{w_{t-\tau_t}}|\mf_{t-\tau_t}\big]-\overline{b}_{w_{t-\tau_t}}} \ltwo{\theta_{t-\tau_t}-\theta^{\lambda*}_{t-\tau_t}} \nonumber\\
	&\overset{(i)}{\leq}\left[2(C^2_\phi+\lambda)R_\theta +\frac{3r_{\max}C_\phi}{1-\gamma} \right]\mE\big[ \ltwo{\theta_t-\theta_{t-\tau_t}}|\mf_{t-\tau_t}\big] + (\Delta_{P,\tau_t}R_\theta + \Delta_{b,\tau_t})\ltwo{\theta_{t-\tau_t}-\theta^{\lambda*}_{t-\tau_t}}\nonumber\\
	&\overset{(ii)}{\leq}\left[2(C^2_\phi+\lambda)R_\theta +\frac{3r_{\max}C_\phi}{1-\gamma} \right]\left[\frac{3}{2}\beta_{t-{\tau_t}}\tau_t(C_\phi^2+\lambda)\ltwo{\theta_{t-\tau_t}-\theta^{\lambda*}_{t-\tau_t}}  + C_5\beta_{t-{\tau_t}}\tau_t\right] \nonumber\\
	&\quad + (\Delta_{P,\tau_t}R_\theta + \Delta_{b,\tau_t})\ltwo{\theta_{t-\tau_t}-\theta^{\lambda*}_{t-\tau_t}}\nonumber\\
	&=\left\{\frac{3}{2}\beta_{t-{\tau_t}}\tau_t(C_\phi^2+\lambda)\left[2(C^2_\phi+\lambda)R_\theta +\frac{3r_{\max}C_\phi}{1-\gamma} \right] + \Delta_{P,\tau_t}R_\theta + \Delta_{b,\tau_t}\right\}\ltwo{\theta_{t-\tau_t}-\theta^{\lambda*}_{t-\tau_t}} \nonumber\\
	&\quad + C_5\beta_{t-{\tau_t}}\tau_t\left[2(C^2_\phi+\lambda)R_\theta +\frac{3r_{\max}C_\phi}{1-\gamma} \right]\nonumber\\
	&\overset{(iii)}{\leq}\left\{\frac{3}{4}\beta_{t-{\tau_t}}\tau_t(C_\phi^2+\lambda)\left[2(C^2_\phi+\lambda)R_\theta +\frac{3r_{\max}C_\phi}{1-\gamma} \right] + \frac{1}{2}\Delta_{P,\tau_t}R_\theta + \frac{1}{2}\Delta_{b,\tau_t}\right\}\ltwo{\theta_{t-\tau_t}-\theta^{\lambda*}_{t-\tau_t}}^2 \nonumber\\
	&\quad + C_5\beta_{t-{\tau_t}}\tau_t\left[2(C^2_\phi+\lambda)R_\theta +\frac{3r_{\max}C_\phi}{1-\gamma} \right] + \frac{3}{4}\beta_{t-{\tau_t}}\tau_t(C_\phi^2+\lambda)\left[2(C^2_\phi+\lambda)R_\theta +\frac{3r_{\max}C_\phi}{1-\gamma} \right] \nonumber\\
	&\quad + \frac{1}{2}\Delta_{P,\tau_t}R_\theta + \frac{1}{2}\Delta_{b,\tau_t}\nonumber\\
	&\overset{(iv)}{\leq}\left\{\frac{3}{4}(C_\phi^2+\lambda)\left[2(C^2_\phi+\lambda)R_\theta +\frac{3r_{\max}C_\phi}{1-\gamma} \right]\beta_{t-{\tau_t}}\tau_t + \frac{1}{2}(C_6R_\theta+C_7)\alpha_{t-\tau_t}\tau^2_t\right\}\ltwo{\theta_{t-\tau_t}-\theta^{\lambda*}_{t-\tau_t}}^2 \nonumber\\
	&\quad + \left[\frac{3}{2}(C^2_\phi+\lambda)^2 R_\theta +\frac{9r_{\max}C_\phi}{4(1-\gamma)}(C_\phi^2+\lambda) + 2C_5(C^2_\phi+\lambda)R_\theta +\frac{3r_{\max}C_\phi C_5}{1-\gamma} \right]\beta_{t-{\tau_t}}\tau_t\nonumber\\
	&\quad + \frac{1}{2}(C_6R_\theta+C_7)\alpha_{t-\tau_t}\tau^2_t\nonumber\\
	&\leq\left\{\frac{3}{4}\left[2(C^2_\phi+\lambda)R_\theta +\frac{3r_{\max}C_\phi}{1-\gamma} \right] + \frac{C_\alpha}{2C_\beta}(C_6R_\theta+C_7)\right\}\beta_{t-{\tau_t}}\tau^2_t\ltwo{\theta_{t-\tau_t}-\theta^{\lambda*}_{t-\tau_t}}^2 \nonumber\\
	&\quad + \left[\frac{3}{2}(C^2_\phi+\lambda)^2 R_\theta +\frac{9r_{\max}C_\phi}{4(1-\gamma)}(C_\phi^2+\lambda) + 2C_5(C^2_\phi+\lambda)R_\theta +\frac{3r_{\max}C_\phi C_5}{1-\gamma} + \frac{C_\alpha}{2C_\beta}(C_6R_\theta+C_7) \right]\beta_{t-{\tau_t}}\tau^2_t \label{incre9}
	\end{flalign}
	where in $(i)$ we denote  $\Delta_{b,\tau_t}\coloneqq \ltwo{\mE\big[\hat{b}_{w_{t-\tau_t}}|\mf_{t-\tau_t}\big]-\overline{b}_{w_{t-\tau_t}}}$, and $(ii)$ follows from Lemma \ref{lemma: support2}, $(iii)$ follows from the fact that $x\leq \frac{1}{2}(x^2+1)$, and $(iv)$ follows from Lemma \ref{lemma: delta}. Consider the third term in \cref{bias_bound1} and we have
	\begin{flalign}
	&\langle g_{t-\tau_t}(\theta_t)-\overline{g}_{t-\tau_t}(\theta_t), \theta^{\lambda*}_{t-\tau_t} - \theta^{\lambda*}_t \rangle\nonumber\\
	&\leq \langle \big(\overline{P}^\lambda_{w_{t-\tau_t}}-P^\lambda_{w_{t-\tau_t}}\big)\theta_t + (\hat{b}_{w_{t-\tau_t}}-\overline{b}_{w_{t-\tau_t}}), \theta^{\lambda*}_{t-\tau_t} - \theta^{\lambda*}_t \rangle\nonumber\\
	&\leq \langle \big(\overline{P}^\lambda_{w_{t-\tau_t}}-P^\lambda_{w_{t-\tau_t}}\big)(\theta_t-\theta^{\lambda*}_t), \theta^{\lambda*}_{t-\tau_t} - \theta^{\lambda*}_t \rangle + \langle \big(\overline{P}^\lambda_{w_{t-\tau_t}}-P^\lambda_{w_{t-\tau_t}}\big)\theta^{\lambda*}_t + (\hat{b}_{w_{t-\tau_t}}-\overline{b}_{w_{t-\tau_t}}), \theta^{\lambda*}_{t-\tau_t} - \theta^{\lambda*}_t \rangle\nonumber\\
	&\leq \ltwo{\overline{P}^\lambda_{w_{t-\tau_t}}-P^\lambda_{w_{t-\tau_t}}}\ltwo{\theta_t-\theta^{\lambda*}_t}\ltwo{\theta^{\lambda*}_{t-\tau_t} - \theta^{\lambda*}_t} \nonumber\\
	&\quad + \left(\ltwo{\overline{P}^\lambda_{w_{t-\tau_t}}-P^\lambda_{w_{t-\tau_t}}} \ltwo{\theta^{\lambda*}_t}	+ \ltwo{\hat{b}_{w_{t-\tau_t}}-\overline{b}_{w_{t-\tau_t}}} \right)\ltwo{\theta^{\lambda*}_{t-\tau_t} - \theta^{\lambda*}_t}\nonumber\\
	&\leq 2(C_\phi^2+\lambda)\ltwo{\theta_t-\theta^{\lambda*}_t} \Big( \sum_{i=t-\tau_t}^{t-1}\ltwo{\theta^{\lambda*}_i-\theta^{\lambda*}_{i+1}} \Big) \nonumber\\
	&\quad + \left( 2R_\theta(C^2_\phi+\lambda) + \frac{3r_{\max}C_\phi}{1-\gamma} \right)\Big( \sum_{i=t-\tau_t}^{t-1}\ltwo{\theta^{\lambda*}_i-\theta^{\lambda*}_{i+1}} \Big)\nonumber\\
	&\overset{(i)}{\leq} 2C_3(C_\phi^2+\lambda)\ltwo{\theta_t-\theta^{\lambda*}_t} \Big( \sum_{i=t-\tau_t}^{t-1} \alpha_i \Big) + C_3\left( 2R_\theta(C^2_\phi+\lambda) + \frac{3r_{\max}C_\phi}{1-\gamma} \right)\Big( \sum_{i=t-\tau_t}^{t-1} \alpha_i \Big)\nonumber\\
	&\overset{(ii)}{\leq} 2C_3(C_\phi^2+\lambda)  \alpha_{t-\tau_t}\tau_t \ltwo{\theta_t-\theta^{\lambda*}_t} + C_3\left( 2R_\theta(C^2_\phi+\lambda) + \frac{3r_{\max}C_\phi}{1-\gamma} \right)\alpha_{t-\tau_t}\tau_t \nonumber\\
	&\overset{(iii)}{\leq} C_3(C_\phi^2+\lambda)  \alpha_{t-\tau_t}\tau_t \ltwo{\theta_t-\theta^{\lambda*}_t}^2 + C_3\left( (2R_\theta+1)(C^2_\phi+\lambda) + \frac{3r_{\max}C_\phi}{1-\gamma} \right)\alpha_{t-\tau_t}\tau_t\nonumber\\
	&\leq \frac{C_3(C_\phi^2+\lambda)C_\alpha}{C_\beta} \beta_{t-\tau_t}\tau^2_t \ltwo{\theta_t-\theta^{\lambda*}_t}^2 + \frac{C_3C_\alpha}{C_\beta}\left( (2R_\theta+1)(C^2_\phi+\lambda) + \frac{3r_{\max}C_\phi}{1-\gamma} \right)\beta_{t-\tau_t}\tau^2_t.
	\label{incre10}
	\end{flalign}
	where $(i)$ follows from Lemma \ref{lemma: fixtracking}, $(ii)$ follows from the fact that $\{\alpha_t\}$ is non-increasing, $(iii)$ follows from the fact that $2x\leq 1+x^2$ and $(iv)$ follows from the fact that $(C^2_\phi+\lambda)\max\{ \alpha_t,\beta_t \}\tau^2_t\leq \frac{1}{4}$ for all $t\geq \hat{t}$. 
	
	Then consider the last two terms in \cref{bias_bound1}, and we have
	\begin{flalign}
	&\mE[\langle g_t(\theta_t)-g_{t-\tau_t}(\theta_t), \theta_t-\theta^{\lambda*}_t \rangle|\mf_{t-\tau_t}] \nonumber\\
	&=\mE[\langle -P^\lambda_{w_t}\theta_t + b_{w_t} + P^\lambda_{w_{t-\tau_t}}\theta_t - \hat{b}_{w_{t-\tau_t}}, \theta_t-\theta^{\lambda*}_t \rangle|\mf_{t-\tau_t}] \nonumber\\
	&\overset{(i)}{=}\mE[\langle -P^\lambda_{w_t}\theta_t + \hat{b}_{w_t} + P^\lambda_{w_{t-\tau_t}}\theta_t - \hat{b}_{w_{t-\tau_t}}, \theta_t-\theta^{\lambda*}_t \rangle|\mf_{t-\tau_t}] \nonumber\\
	&=\mE[\langle (P^\lambda_{w_{t-\tau_t}} - P^\lambda_{w_t} ) (\theta_t - \theta^{\lambda*}_t), \theta_t-\theta^{\lambda*}_t \rangle |\mf_{t-\tau_t} ]+ \mE[\langle (P^\lambda_{w_{t-\tau_t}} - P^\lambda_{w_t} ) \theta^{\lambda*}_t, \theta_t-\theta^{\lambda*}_t \rangle |\mf_{t-\tau_t}] \nonumber\\
	&\quad + \mE [\langle  \hat{b}_{w_t} - \hat{b}_{w_{t-\tau_t}}, \theta_t-\theta^{\lambda*}_t \rangle |\mf_{t-\tau_t}] \nonumber\\
	&\leq \mE[\ltwo{ P^\lambda_{w_{t-\tau_t}} - P^\lambda_{w_t} } \ltwo{\theta_t - \theta^{\lambda*}_t}^2 |\mf_{t-\tau_t} ]+ \mE[\ltwo{P^\lambda_{w_{t-\tau_t}} - P^\lambda_{w_t}} \ltwo{\theta^{\lambda*}_t} \ltwo{\theta_t-\theta^{\lambda*}_t } |\mf_{t-\tau_t}] \nonumber\\
	&\quad + \mE [ \ltwo{ \hat{b}_{w_t} - \hat{b}_{w_{t-\tau_t}}} \ltwo{\theta_t-\theta^{\lambda*}_t}  |\mf_{t-\tau_t}] \nonumber\\
	&\overset{(ii)}{\leq} 2C_\phi L_\phi  \mE [ \ltwo{ w_t - w_{t-\tau_t}} \ltwo{\theta_t-\theta^{\lambda*}_t}^2  |\mf_{t-\tau_t}] + 2C_\phi L_\phi R_\theta \ltwo{ w_t - w_{t-\tau_t}} \mE [ \ltwo{ w_t - w_{t-\tau_t}}  \ltwo{\theta_t-\theta^{\lambda*}_t}  |\mf_{t-\tau_t}] \nonumber\\
	&\quad + \left(\frac{L_\phi r_{\max}}{1-\gamma} + 2C_\phi L_z \right)  \mE [ \ltwo{ w_t - w_{t-\tau_t}} \ltwo{\theta_t-\theta^{\lambda*}_t}  |\mf_{t-\tau_t}] \nonumber\\
	&\leq 2C_\phi L_\phi R_\theta \max\{ 1, C^2_\phi \} \alpha_{t-\tau_t}\tau_t \mE [ \ltwo{\theta_t-\theta^{\lambda*}_t}^2  |\mf_{t-\tau_t}] + 2C_\phi L_\phi R^2_\theta \max\{ 1, C^2_\phi \} \alpha_{t-\tau_t}\tau_t \mE [ \ltwo{\theta_t-\theta^{\lambda*}_t}  |\mf_{t-\tau_t}] \nonumber\\
	&\quad + \left(\frac{L_\phi r_{\max}}{1-\gamma} + 2C_\phi L_z \right) R_\theta \max\{ 1, C^2_\phi \} \alpha_{t-\tau_t}\tau_t \mE [ \ltwo{\theta_t-\theta^{\lambda*}_t}  |\mf_{t-\tau_t}] \nonumber\\
	&\leq \left(2C_\phi L_\phi R_\theta + C_\phi L_\phi R^2_\theta +  \left(\frac{L_\phi r_{\max}}{2(1-\gamma)} + C_\phi L_z \right) R_\theta \right) \max\{ 1, C^2_\phi \} \alpha_{t-\tau_t}\tau_t \mE [ \ltwo{\theta_t-\theta^{\lambda*}_t}^2  |\mf_{t-\tau_t}] \nonumber\\
	&\quad + \left[C_\phi L_\phi R^2_\theta +  \left(\frac{L_\phi r_{\max}}{2(1-\gamma)} + C_\phi L_z \right) R_\theta \right]\max\{ 1, C^2_\phi \}\alpha_{t-\tau_t}\tau_t \nonumber\\
	&\overset{(ii)}{\leq} C_{23} \beta_{t-\tau_t}\tau_t \mE [ \ltwo{\theta_t-\theta^{\lambda*}_t}^2  |\mf_{t-\tau_t}] + C_{24}\beta_{t-\tau_t}\tau_t. \label{incre30}
	\end{flalign}
	where $(i)$ follows from the definition $\hat{b}_{w_{t}}=\phi_{w_{t}}(s_t,a_t) A_{\pi_{w_{t}}}(s_t,a_t)$ and by taking expectation with respect to the randomness of $\hat{Q}_{\pi_{w_{t}}}(s_t,a_t)$ and $a^\prime_t$, $(ii)$ follows from Lemma \ref{bias_supp1}, and the definition
	\begin{flalign*}
		C_{23}=\left(2C_\phi L_\phi R_\theta + C_\phi L_\phi R^2_\theta +  \left(\frac{L_\phi r_{\max}}{2(1-\gamma)} + C_\phi L_z \right) R_\theta \right) \max\{ 1, C^2_\phi \} \frac{C_\alpha}{C_\beta}.
	\end{flalign*}
	and
	\begin{flalign*}
		C_{24}=\left[C_\phi L_\phi R^2_\theta +  \left(\frac{L_\phi r_{\max}}{2(1-\gamma)} + C_\phi L_z \right) R_\theta \right] \max\{ 1, C^2_\phi \} \frac{C_\alpha}{C_\beta}
	\end{flalign*}
	We also have
	\begin{flalign}
	&\mE[\langle \overline{g}_t(\theta_t)-\overline{g}_{t-\tau_t}(\theta_t), \theta_t-\theta^{\lambda*}_t \rangle|\mf_{t-\tau_t}] \nonumber\\
	&=\mE[\langle -\overline{P}^\lambda_{w_t}\theta_t + \overline{b}_{w_t} + \overline{P}^\lambda_{w_{t-\tau_t}}\theta_t - \overline{b}_{w_{t-\tau_t}}, \theta_t-\theta^{\lambda*}_t \rangle|\mf_{t-\tau_t}] \nonumber\\
	&=\mE[\langle (\overline{P}^\lambda_{w_{t-\tau_t}} - \overline{P}^\lambda_{w_t} ) (\theta_t - \theta^{\lambda*}_t), \theta_t-\theta^{\lambda*}_t \rangle|\mf_{t-\tau_t}] + \mE[\langle (\overline{P}^\lambda_{w_{t-\tau_t}} - \overline{P}^\lambda_{w_t} ) \theta^{\lambda*}_t, \theta_t-\theta^{\lambda*}_t \rangle|\mf_{t-\tau_t}] \nonumber\\
	&\quad + \mE[\langle  \overline{b}_{w_t} - \overline{b}_{w_{t-\tau_t}}, \theta_t-\theta^{\lambda*}_t \rangle|\mf_{t-\tau_t}] \nonumber\\
	&\leq \mE\left[\ltwo{ \overline{P}^\lambda_{w_{t-\tau_t}} - \overline{P}^\lambda_{w_t} } \ltwo{\theta_t - \theta^{\lambda*}_t}^2  |\mf_{t-\tau_t} \right] + \mE\left[ \ltwo{\overline{P}^\lambda_{w_{t-\tau_t}} - \overline{P}^\lambda_{w_t}}  \ltwo{\theta^{\lambda*}_t} \ltwo{\theta_t-\theta^{\lambda*}_t} |\mf_{t-\tau_t}\right] \nonumber\\
	&\quad + \mE\left[ \ltwo{\overline{b}_{w_t} - \overline{b}_{w_{t-\tau_t}}}  \ltwo{\theta_t-\theta^{\lambda*}_t } |\mf_{t-\tau_t}\right] \nonumber\\
	&\overset{(i)}{\leq} 2(C_\phi^2C_\nu + C_\phi L_\phi)  \mE\left[ \ltwo{w-w^\prime} \ltwo{\theta_t - \theta^{\lambda*}_t}^2  |\mf_{t-\tau_t} \right] \nonumber\\
	&\quad + 2(C_\phi^2C_\nu + C_\phi L_\phi) R_\theta  \mE\left[ \ltwo{w-w^\prime} \ltwo{\theta_t - \theta^{\lambda*}_t}  |\mf_{t-\tau_t} \right] \nonumber\\
	&\quad + \left( \frac{2C_\phi C_\nu r_{\max}}{1-\gamma} + \frac{L_\phi r_{\max}}{1-\gamma} + 2C_\phi L_z \right) \mE\left[ \ltwo{w-w^\prime} \ltwo{\theta_t - \theta^{\lambda*}_t}  |\mf_{t-\tau_t} \right]\nonumber\\
	&\overset{(ii)}{\leq} C_{25}\beta_{t-\tau_t}\tau_t \mE\left[ \ltwo{\theta_t - \theta^{\lambda*}_t}^2  |\mf_{t-\tau_t} \right] + C_{26}\beta_{t-\tau_t}\tau_t,\label{incre31}
	\end{flalign}
	where $(i)$ follows from Lemma \ref{bias_supp2} and in $(ii)$ we define 
	$$C_{25}=\left[ 2(C_\phi^2C_\nu + C_\phi L_\phi) + (C_\phi^2C_\nu + C_\phi L_\phi) R_\theta + \left( \frac{2C_\phi C_\nu r_{\max}}{1-\gamma} + \frac{L_\phi r_{\max}}{2(1-\gamma)} + C_\phi L_z \right) \right]R_\theta \max\{1, C^2_\phi\} \frac{C_\alpha}{C_\beta}$$ 
	and $C_{26}=\left[ (C_\phi^2C_\nu + C_\phi L_\phi) R_\theta + \left( \frac{2C_\phi C_\nu r_{\max}}{1-\gamma} + \frac{L_\phi r_{\max}}{2(1-\gamma)} + C_\phi L_z \right) \right]R_\theta \max\{1, C^2_\phi\} \frac{C_\alpha}{C_\beta}$. Combining \cref{incre8}, \cref{incre9} and \cref{incre10} yields
	\begin{flalign}
	&\mE[\langle g_t(\theta_t)-\overline{g}_t(\theta_t), \theta_t-\theta^{\lambda*}_t \rangle|\mf_{t-\tau_t}]\nonumber\\
	&\leq\Bigg\{\frac{C_6C_\alpha}{C_\beta}+\left[\big(3(C_\phi^2+\lambda)+C_5\big)\right](C_\phi^2+\lambda) + \frac{3}{4}\left[2(C^2_\phi+\lambda)R_\theta +\frac{3r_{\max}C_\phi}{1-\gamma} \right] \nonumber\\
	&\qquad+\frac{C_\alpha}{2C_\beta}(C_6R_\theta+C_7)\Bigg\}\beta_{t-{\tau_t}}\tau^2_t\ltwo{\theta_{t-\tau_t}-\theta^{\lambda*}_{t-\tau_t}}^2 \nonumber\\
	&\quad + \left(\frac{C_3(C_\phi^2+\lambda)C_\alpha}{C_\beta} + C_{23} + C_{25}\right) \beta_{t-\tau_t}\tau^2_t \mE\left[\ltwo{\theta_t-\theta^{\lambda*}_t}^2|\mf_{t-\tau_t}\right] \nonumber\\
	&\quad + \frac{C_3C_\alpha}{C_\beta}\left( (2R_\theta+1)(C^2_\phi+\lambda) + \frac{3r_{\max}C_\phi}{1-\gamma} \right)\beta_{t-\tau_t}\tau^2_t\nonumber\\
	&\quad + \Bigg[\frac{3}{2}(C^2_\phi+\lambda)^2 R_\theta +\frac{9r_{\max}C_\phi}{4(1-\gamma)}(C_\phi^2+\lambda) + 2C_5(C^2_\phi+\lambda)R_\theta +\frac{3r_{\max}C_\phi C_5}{1-\gamma} + \frac{C_\alpha}{2C_\beta}(C_6R_\theta+C_7)  \nonumber\\
	&\quad \quad \quad+ C_5(C_\phi^2+\lambda) + C_{24} + C_{26} \Bigg]\beta_{t-{\tau_t}}\tau^2_t  + 2(C_\phi^2+\lambda)\mE\left[ \ltwo{\theta_t-\theta_{t-\tau_t}}^2 |\mf_{t-\tau_t}\right] \nonumber\\
	&\overset{(i)}{=}C_9\beta_{t-{\tau_t}}\tau^2_t\ltwo{\theta_{t-\tau_t}-\theta^{\lambda*}_{t-\tau_t}}^2 + C_{10}\beta_{t-\tau_t}\tau^2_t \mE\left[\ltwo{\theta_t-\theta^{\lambda*}_t}^2|\mf_{t-\tau_t}\right]   \nonumber\\
	&\quad+ 2(C_\phi^2+\lambda)\mE\left[ \ltwo{\theta_t-\theta_{t-\tau_t}}^2 |\mf_{t-\tau_t}\right] + C_{11}\beta_{t-{\tau_t}}\tau^2_t. \label{incre21}
	\end{flalign}
	where in $(i)$ we define
	\begin{flalign*}
		C_9&=\frac{C_6C_\alpha}{C_\beta}+\left[\big(3(C_\phi^2+\lambda)+C_5\big)\right](C_\phi^2+\lambda) + \frac{3}{4}\left[2(C^2_\phi+\lambda)R_\theta +\frac{3r_{\max}C_\phi}{1-\gamma} \right] + \frac{C_\alpha}{2C_\beta}(C_6R_\theta+C_7),\\
		C_{10}&=\frac{C_3(C_\phi^2+\lambda)C_\alpha}{C_\beta}
	\end{flalign*} 
	and 
	\begin{flalign*}
		C_{11}&=\frac{C_3C_\alpha}{C_\beta}\left( (2R_\theta+1)(C^2_\phi+\lambda) + \frac{3r_{\max}C_\phi}{1-\gamma} \right) + \frac{3}{2}(C^2_\phi+\lambda)^2 R_\theta +\frac{9r_{\max}C_\phi}{4(1-\gamma)}(C_\phi^2+\lambda) \nonumber\\ 
		&\quad + 2C_5(C^2_\phi+\lambda)R_\theta +\frac{3r_{\max}C_\phi C_5}{1-\gamma} + \frac{C_\alpha}{2C_\beta}(C_6R_\theta+C_7) + C_5(C_\phi^2+\lambda) + C_{24} + C_{26}.
	\end{flalign*}
	Note that
	\begin{flalign}
	&\ltwo{\theta_{t-\tau_t}-\theta^{\lambda*}_{t-\tau_t}}^2\nonumber\\
	&=\ltwo{ \theta_{t}-\theta^{\lambda*}_{t} + \theta^{\lambda*}_{t}-\theta^{\lambda*}_{t-\tau_t} + \theta_{t-\tau_t}-\theta_{t}}^2\nonumber\\
	&\leq 3\ltwo{\theta_{t}-\theta^{\lambda*}_{t}}^2 + 3\ltwo{\theta^{\lambda*}_{t}-\theta^{\lambda*}_{t-\tau_t}}^2 + 3\ltwo{\theta_{t-\tau_t}-\theta_{t}}^2\nonumber\\
	&\overset{(i)}{\leq} 3\ltwo{\theta_{t}-\theta^{\lambda*}_{t}}^2 + 3L^2_\theta\Big(\sum_{i=t-\tau_t}^{t-1}\alpha_i\Big)^2 + 3\ltwo{\theta_{t-\tau_t}-\theta_{t}}^2\nonumber\\
	&\leq 3\ltwo{\theta_{t}-\theta^{\lambda*}_{t}}^2 + 3L^2_\theta R^2_\theta \max\{1,C^2_\phi\}^2 \alpha^2_{t-\tau_t}\tau^2_t +  3\ltwo{\theta_{t-\tau_t}-\theta_{t}}^2\nonumber\\
	&\leq 3\ltwo{\theta_{t}-\theta^{\lambda*}_{t}}^2 + 3L^2_\theta R^2_\theta \max\{1,C^2_\phi\}^2 \alpha^2_{t-\tau_t}\tau^2_t +  \frac{3}{4}(C_\phi^2+\lambda)\beta_{t-{\tau_t}}\tau^2_t\ltwo{\theta_{t-\tau_t}-\theta^{\lambda*}_{t-\tau_t}}^2 \nonumber\\
	 &\quad + \left[C_5+\frac{3}{4}(C_\phi^2+\lambda)\right]\beta_{t-{\tau_t}}\tau^2_t\nonumber\\
	&\leq 3\ltwo{\theta_{t}-\theta^{\lambda*}_{t}}^2 + \frac{3}{16}\ltwo{\theta_{t-\tau_t}-\theta^{\lambda*}_{t-\tau_t}}^2 + \left[ C_5 + \frac{3}{4}(C^2_\phi + \lambda) + \frac{3C^2_\alpha L^2_\theta R^2_\theta \max\{1,C^2_\phi\}^2}{C_\beta}  \right]\beta_{t-{\tau_t}}\tau^2_t, \label{incre22}
	\end{flalign}
	where $(i)$ follows from Lemma \ref{lemma: fixtracking}. Following from \cref{incre22}, we obtain
	\begin{flalign}
	\ltwo{\theta_{t-\tau_t}-\theta^{\lambda*}_{t-\tau_t}}^2\leq \frac{48}{13}\ltwo{\theta_{t}-\theta^{\lambda*}_{t}}^2 + \frac{16}{13}\left[C_5+\frac{3}{4}(C_\phi^2+\lambda) + \frac{3C^2_\alpha L^2_\theta R^2_\theta \max\{1,C^2_\phi\}^2}{C_\beta}  \right]\beta_{t-{\tau_t}}\tau^2_t.\label{incre23}
	\end{flalign}
	Substituting \cref{incre23} into \cref{incre21} yields
	\begin{flalign}
	&\mE[\langle g_t(\theta_t)-\overline{g}_t(\theta_t), \theta_t-\theta^{\lambda*}_t \rangle|\mf_{t-\tau_t}]\nonumber\\
	&\leq C_9\beta_{t-{\tau_t}}\tau^2_t \left\{ \frac{48}{13}\ltwo{\theta_{t}-\theta^{\lambda*}_{t}}^2 + \frac{16}{13}\left[C_5+\frac{3}{4}(C_\phi^2+\lambda) + \frac{3C^2_\alpha L^2_\theta R^2_\theta \max\{1,C^2_\phi\}^2}{C_\beta}  \right]\beta_{t-{\tau_t}}\tau^2_t \right\}  \nonumber\\
	&\quad + C_{10}\beta_{t-\tau_t}\tau^2_t \mE\left[\ltwo{\theta_t-\theta^{\lambda*}_t}^2|\mf_{t-\tau_t}\right] + 2(C_\phi^2+\lambda)\mE\left[ \ltwo{\theta_t-\theta_{t-\tau_t}}^2 |\mf_{t-\tau_t}\right] + C_{11}\beta_{t-{\tau_t}}\tau^2_t\nonumber\\
	&\leq C_9\beta_{t-{\tau_t}}\tau^2_t \left\{ \frac{48}{13}\mE\left[\ltwo{\theta_t-\theta^{\lambda*}_t}^2|\mf_{t-\tau_t}\right] + \frac{16}{13}\left[C_5+\frac{3}{4}(C_\phi^2+\lambda) + \frac{3C^2_\alpha L^2_\theta R^2_\theta \max\{1,C^2_\phi\}^2}{C_\beta} \right]\beta_{t-{\tau_t}}\tau^2_t \right\}  \nonumber\\
	&\quad + 2(C_\phi^2+\lambda)\left[ \frac{3}{4}(C_\phi^2+\lambda)\beta_{t-{\tau_t}}\tau^2_t\ltwo{\theta_{t-\tau_t}-\theta^{\lambda*}_{t-\tau_t}}^2  + \Big[\frac{3}{4}(C^2_\phi+\lambda)+C_5\Big]\beta_{t-{\tau_t}}\tau^2_t \right] + C_{11}\beta_{t-{\tau_t}}\tau^2_t\nonumber\\
	&\quad+ C_{10}\beta_{t-\tau_t}\tau^2_t \mE\left[\ltwo{\theta_t-\theta^{\lambda*}_t}^2|\mf_{t-\tau_t}\right] \nonumber\\
	&=\left( \frac{48C_9}{13} + C_{10}\right)\beta_{t-{\tau_t}}\tau^2_t\mE\left[\ltwo{\theta_t-\theta^{\lambda*}_t}^2|\mf_{t-\tau_t}\right] + \frac{3}{2}(C_\phi^2+\lambda)^2\beta_{t-{\tau_t}}\tau^2_t\ltwo{\theta_{t-\tau_t}-\theta^{\lambda*}_{t-\tau_t}}^2\nonumber\\
	&\quad + \Bigg\{\frac{4C_9}{13(C^2_\phi+\lambda)}\left[C_5+\frac{3}{4}(C_\phi^2+\lambda) + \frac{3C^2_\alpha L^2_\theta R^2_\theta \max\{1,C^2_\phi\}^2}{C_\beta} \right] \nonumber\\
	&\quad\qquad + 2(C_\phi^2+\lambda)\Big[\frac{3}{4}(C^2_\phi+\lambda)+C_5\Big] + C_{11}\Bigg\}\beta_{t-{\tau_t}}\tau^2_t\nonumber\\
	&\leq \left( \frac{48C_9}{13} + C_{10}\right)\beta_{t-{\tau_t}}\tau^2_t\mE\left[\ltwo{\theta_t-\theta^{\lambda*}_t}^2|\mf_{t-\tau_t}\right] + \frac{3}{2}(C_\phi^2+\lambda)^2\beta_{t-{\tau_t}}\tau^2_t \Bigg[\frac{48}{13}\mE\left[\ltwo{\theta_t-\theta^{\lambda*}_t}^2|\mf_{t-\tau_t}\right] \nonumber\\
	&\quad+ \frac{16}{13}\left[C_5+\frac{3}{4}(C_\phi^2+\lambda)\right]\beta_{t-{\tau_t}}\tau^2_t \Bigg]+ \Bigg\{\frac{4C_9}{13(C^2_\phi+\lambda)}\left[C_5+\frac{3}{4}(C_\phi^2+\lambda) + \frac{3C^2_\alpha L^2_\theta R^2_\theta \max\{1,C^2_\phi\}^2}{C_\beta} \right] \nonumber\\
	&\quad + 2(C_\phi^2+\lambda)\Big[\frac{3}{4}(C^2_\phi+\lambda)+C_5\Big] + C_{11}\Bigg\}\beta_{t-{\tau_t}}\tau^2_t\nonumber\\
	&=\left[ \frac{48C_9}{13} + C_{10} + \frac{72}{13}(C_\phi^2+\lambda)^2\right]\beta_{t-{\tau_t}}\tau^2_t\mE\left[\ltwo{\theta_t-\theta^{\lambda*}_t}^2|\mf_{t-\tau_t}\right] \nonumber\\
	&\quad + \Bigg\{\frac{4C_9}{13(C^2_\phi+\lambda)}\left[C_5+\frac{3}{4}(C_\phi^2+\lambda) \frac{3C^2_\alpha L^2_\theta R^2_\theta \max\{1,C^2_\phi\}^2}{C_\beta} \right] + 2(C_\phi^2+\lambda)\Big[\frac{3}{4}(C^2_\phi+\lambda)+C_5\Big] \nonumber\\
	&\quad + \frac{6}{13}(C_\phi^2+\lambda)\left[C_5+\frac{3}{4}(C_\phi^2+\lambda)\right] + C_{11}\Bigg\}\beta_{t-{\tau_t}}\tau^2_t\nonumber\\
	&= C_{12}\beta_{t-{\tau_t}}\tau^2_t\mE\left[\ltwo{\theta_t-\theta^{\lambda*}_t}^2|\mf_{t-\tau_t}\right] + C_{13}\beta_{t-{\tau_t}}\tau^2_t.\nonumber
	\end{flalign}
\end{proof}

\begin{proof}[\textbf{Proof of Lemma \ref{lemma: accumulate1}}]
	Due to the definition, we have
	\begin{flalign}
	\sum_{i=\hat{t}}^{t-1} e^{-\lambda_P \sum_{k=i+1}^{t-1}\beta_k} \beta^2_i\tau^2_i &\leq \tau^2_t \sum_{i=\hat{t}}^{t-1} e^{-\lambda_P \sum_{k=i+1}^{t-1}\beta_k} \beta^2_i\leq \tau^2_t \max_{i\in [\hat{t},t]}\{ e^{-\frac{\lambda_P}{2} \sum_{k=i+1}^{t-1}\beta_k} \beta_i \} \sum_{i=\hat{t}}^{t-1} e^{-\frac{\lambda_P}{2} \sum_{k=i+1}^{t-1}\beta_k} \beta_i\nonumber\\
	&\leq \frac{ 2\tau^2_t e^{ \frac{\lambda_P C_\beta}{2} } }{ \lambda_\theta } \max_{i\in [\hat{t},t]}\{ e^{-\frac{\lambda_P}{2} \sum_{k=i+1}^{t-1}\beta_k} \beta_i \}. \label{incre28}
	\end{flalign}
On the right hand side of \cref{incre28}, we define $y_i=e^{-\frac{\lambda_P}{2} \sum_{k=i+1}^{t-1}\beta_k} \beta_i$. Then it can be shown that the sequence $\{ y_i \}$ is non-decreasing for $i\geq i_\beta =  (\frac{2\nu}{C_\beta \lambda_P})^{\frac{1}{1-\nu}}$. If $\hat{t}\geq i_\beta$, then we have 
	\begin{flalign*}
	\max_{i\in [\hat{t},t]}\{ e^{-\frac{\lambda_P}{2} \sum_{k=i+1}^{t-1}\beta_k} \beta_i \} = y_t = \beta_t.
	\end{flalign*}
	If $\hat{t}\leq i_\beta$, then we have
	\begin{flalign}
	\max_{i\in [\hat{t},t]}\{ e^{-\frac{\lambda_P}{2} \sum_{k=i+1}^{t-1}\beta_k} \beta_i \} &\leq \max_{i\in [\hat{t},i_\beta-1]}\{ e^{-\frac{\lambda_P}{2} \sum_{k=i+1}^{t-1}\beta_k} \beta_i \} + \max_{i\in [i_\beta,t]}\{ e^{-\frac{\lambda_P}{2} \sum_{k=i+1}^{t-1}\beta_k} \beta_i \}\nonumber\\
	&\leq e^{-\frac{\lambda_P}{2} \sum_{k=\hat{t}}^{t-1}\beta_k}  \max_{i\in [\hat{t},i_\beta-1]}\{ e^{\frac{\lambda_P}{2} \sum_{k=i+1}^{t-1}\beta_k} \beta_i \} + \beta_{t-1}.\nonumber
	\end{flalign}
	Thus
	\begin{flalign}
	\sum_{i=\hat{t}}^{t-1} e^{-\lambda_P \sum_{k=i+1}^{t-1}\beta_k} \beta^2_i\tau^2_i &\leq \frac{ 2\tau^2_t e^{ \frac{\lambda_P C_\beta}{2} } }{ \lambda_P } \left[ e^{-\frac{\lambda_P}{2} \sum_{k=\hat{t}}^{t-1}\beta_k} \max_{i\in [\hat{t},i_\beta-1]}\{ e^{\frac{\lambda_P}{2} \sum_{k=i+1}^{t-1}\beta_k} \beta_i \} + \beta_{t-1} \right] \nonumber\\
	& \leq \frac{ 2\tau^2_t e^{ \frac{\lambda_P C_\beta}{2} } }{ \lambda_\theta } \left[ e^{-\frac{\lambda_P C_\beta }{2(1-\nu)} [t^{1-\nu} - (\hat{t}+1)^{1-\nu}]} \max_{i\in [\hat{t},i_\beta-1]}\{ e^{\frac{\lambda_P}{2} \sum_{k=i+1}^{t-1}\beta_k} \beta_i \} + \beta_{t-1} \right]\nonumber\\
	&= C_{18}\tau^2_t e^{-\frac{\lambda_P C_\beta }{2(1-\nu)} [t^{1-\nu} - (\hat{t}+1)^{1-\nu}]} + C_{19} \tau^2_t \beta_{t-1}.
	\end{flalign}
\end{proof}

\begin{proof}[\textbf{Proof of Lemma \ref{lemma: accumulate2}}]
	Due to the definition, we have
	\begin{flalign}
	&\sum_{i=\hat{t}}^{t-1} e^{-\lambda_P \sum_{k=i+1}^{t-1}\beta_k} \alpha_i \nonumber\\
	&\leq \frac{C_\alpha}{C_\beta}\sum_{i=\hat{t}}^{t-1} e^{-\lambda_P \sum_{k=i+1}^{t-1}\beta_k} \beta_i \frac{1}{ (1+i)^{\sigma-\nu}} \leq \frac{C_\alpha}{C_\beta} \max_{i\in [\hat{t},t]}\Big\{ e^{-\frac{\lambda_P}{2} \sum_{k=i+1}^{t-1}\beta_k} \frac{1}{ (1+i)^{\sigma-\nu}} \Big\} \sum_{i=\hat{t}}^{t-1} e^{-\frac{\lambda_P}{2} \sum_{k=i+1}^{t-1}\beta_k} \beta_i\nonumber\\
	&\leq \frac{ 2C_\alpha e^{ \frac{\lambda_P C_\beta}{2} } }{ C_\beta \lambda_P } \max_{i\in [\hat{t},t]}\Big\{ e^{-\frac{\lambda_P}{2} \sum_{k=i+1}^{t-1}\beta_k}  \frac{1}{(1+i)^{\sigma-\nu}} \Big\}. \label{incre29}
	\end{flalign}
On the right-hand side of \cref{incre29}, we define $y_i=e^{-\frac{\lambda_P}{2} \sum_{k=i+1}^{t-1}\beta_k} \frac{1}{ (1+i)^{\sigma-\nu}}$. Then it can be shown that the sequence $\{ y_i \}$ is non-decreasing if $i\geq i_{\alpha} =  (\frac{2(\sigma-\nu)}{C_\beta \lambda_P})^{\frac{1}{1-(\sigma-\nu)}}$. If $\hat{t}\geq i_\alpha$, then we have 
	\begin{flalign*}
	\max_{i\in [\hat{t},t]}\{ e^{-\frac{\lambda_P}{2} \sum_{k=i+1}^{t-1}\beta_k} \beta_i \} = y_t = \frac{1}{(1+t)^{\sigma-\nu}}.
	\end{flalign*}
	If $\hat{t}\leq i_\alpha$, then we have
	\begin{flalign}
	&\max_{i\in [\hat{t},t]}\Big\{ e^{-\frac{\lambda_P}{2} \sum_{k=i+1}^{t-1}\beta_k} \frac{1}{(1+i)^{\sigma-\nu}} \Big\} \nonumber\\
	&\leq \max_{i\in [\hat{t},i_\alpha-1]}\{ e^{-\frac{\lambda_P}{2} \sum_{k=i+1}^{t-1}\beta_k} \frac{1}{(1+i)^{\sigma-\nu}} \} + \max_{i\in [i_\alpha,t]}\{ e^{-\frac{\lambda_P}{2} \sum_{k=i+1}^{t-1}\beta_k} \frac{1}{(1+i)^{\sigma-\nu}} \}\nonumber\\
	&\leq e^{-\frac{\lambda_P}{2} \sum_{k=\hat{t}}^{t-1}\beta_k} \max_{i\in [\hat{t},i_\alpha-1]}\Big\{ e^{\frac{\lambda_P}{2} \sum_{k=i+1}^{t-1}\beta_k} \frac{1}{(1+i)^{\sigma-\nu}} \Big\} + \frac{1}{t^{\sigma-\nu}}.\nonumber
	\end{flalign}
	Thus
	\begin{flalign}
	&\sum_{i=\hat{t}}^{t-1} e^{-\lambda_P \sum_{k=i+1}^{t-1}\beta_k} \alpha_i \nonumber\\
	&\leq \frac{ 2 C_\alpha e^{ \frac{\lambda_P C_\beta}{2} } }{ C_\beta \lambda_P } \left[ e^{-\frac{\lambda_P}{2} \sum_{k=0}^{t-1}\beta_k} \max_{i\in [\hat{t},i_\alpha-1]}\Big\{ e^{\frac{\lambda_P}{2} \sum_{k=i+1}^{t-1}\beta_k} \frac{1}{(1+i)^{\sigma-\nu}} \Big\} + \frac{1}{t^{\sigma-\nu}} \right] \nonumber\\
	& \leq \frac{ 2C_\alpha e^{ \frac{\lambda_P C_\beta}{2} } }{ C_\beta\lambda_P } \left[ e^{-\frac{\lambda_P C_\beta }{2(1-\nu)} [(t+1)^{1-\nu} - (\hat{t}+1)^{1-\nu}]} \max_{i\in [\hat{t},i_\beta-1]}\Big\{ e^{\frac{\lambda_\theta}{2} \sum_{k=i+1}^{t-1}\beta_k} \frac{1}{(1+i)^{\sigma-\nu}}  \Big\} + \frac{1}{t^{\sigma-\nu}}  \right]\nonumber\\
	&= C_{20}e^{-\frac{\lambda_P C_\beta }{2(1-\nu)} [(t+1)^{1-\nu} - (\hat{t}+1)^{1-\nu}]} + \frac{C_{21}}{t^{\sigma-\nu}}.\nonumber
	\end{flalign}
\end{proof}

\bibliographystyle{apalike}
\bibliography{ref}

\end{document}